\documentclass[sn-apa,iicol]{sn-jnl}

\usepackage{multirow}
\usepackage{subcaption}
\usepackage{amsmath}
\usepackage{float}
\usepackage[flushleft]{threeparttable}
\usepackage[symbol]{footmisc}
\usepackage{gensymb}
\usepackage{xcolor}

\jyear{2022}%

\theoremstyle{thmstyleone}%
\theoremstyle{thmstyletwo}%

\theoremstyle{thmstylethree}%

\begin{document}

\title[Spatial Monitoring and Insect Behavioural Analysis Using Computer Vision for Precision Pollination]{Spatial Monitoring and Insect Behavioural Analysis Using Computer Vision for Precision Pollination}



\author*[1]{\fnm{Malika Nisal} \sur{Ratnayake}}\email{malika.ratnayake@monash.edu}

\author[1]{\fnm{Don Chathurika} \sur{Amarathunga}}\email{don.amarathunga@monash.edu}

\author[1]{\fnm{Asaduz} \sur{Zaman}}\email{asaduzzaman@monash.edu}

\author[2,3]{\fnm{Adrian G.} \sur{Dyer}}\email{adrian.dyer@rmit.edu.au, adrian.dyer@monash.edu}

\author[1]{\fnm{Alan} \sur{Dorin}}\email{alan.dorin@monash.edu}


\affil[1]{\orgdiv{Computational and Collective Intelligence Group, Dept. of Data Science and AI, Faculty of Information Technology}, \orgname{Monash University}, \orgaddress{\street{Wellington Road}, \city{Melbourne}, \postcode{3800}, \state{Victoria}, \country{Australia}}}

\affil[2]{\orgdiv{School of Media and Communication}, \orgname{RMIT University}, \orgaddress{\street{La Trobe Street}, \city{Melbourne}, \postcode{3000}, \state{Victoria}, \country{Australia}}}

\affil[3]{\orgdiv{Department of Physiology}, \orgname{Monash University}, \orgaddress{\street{Wellington Road}, \city{Melbourne}, \postcode{3800}, \state{Victoria}, \country{Australia}}}



\abstract{Insects are the most important global pollinator of crops and play a key role in maintaining the sustainability of natural ecosystems. Insect pollination monitoring and management are therefore essential for improving crop production and food security. Computer vision facilitated pollinator monitoring can intensify data collection over what is feasible using manual approaches. The new data it generates may provide a detailed understanding of insect distributions and facilitate fine-grained analysis sufficient to predict their pollination efficacy and underpin precision pollination. Current computer vision facilitated insect tracking in complex outdoor environments is restricted in spatial coverage and often constrained to a single insect species. This limits its relevance to agriculture. Therefore, in this article we introduce a novel system to facilitate markerless data capture for insect counting, insect motion tracking, behaviour analysis and pollination prediction across large agricultural areas. Our system is comprised of edge computing multi-point video recording, offline automated multi-species insect counting, tracking and behavioural analysis. We implement and test our system on a commercial berry farm to demonstrate its capabilities. Our system successfully tracked four insect varieties, at nine monitoring stations within polytunnels, obtaining an F-score above 0.8 for each variety. The system enabled calculation of key metrics to assess the relative pollination impact of each insect variety. With this technological advancement, detailed, ongoing data collection for precision pollination becomes achievable. This is important to inform growers and apiarists managing crop pollination, as it allows data-driven decisions to be made to improve food production and food security.
}




\maketitle

\section{Introduction}\label{sec1}

Pollinators play a key role in world food production and ecosystem management. Three out of four flowering plants \citep{fao2019} and 35\% of agricultural land \citep{faobee} require some degree of animal pollination. This includes over 87 high-value food crops consumed by humans \citep{aizen2009}. The annual market value of pollinator contributions to global food production is estimated to be in the range of $235-577$ billion USD \citep{potts2016}.

Recently, climate change and other anthropogenic pressures have been implicated in declines in some pollinator populations \citep{ecosystemadam, schweiger2010multiple}, threatening global food security. In many instances, pollinator population size is directly correlated with crop yield \citep{rollin2019impacts}, although the efficiency of different pollinator populations varies between crops \citep{macinnis2019}. Hence, improved understanding and management of pollinator communities is important to boost crop yield \citep{garibaldi2017towards}, and for the long-term viability of many farming projects \citep{garibaldi2020a}. This need strongly motivates the research presented here to describe the design and implementation of computer vision facilitated spatial monitoring and insect behavioural analysis for \emph{precision pollination}.

Insect monitoring and sampling can help us to understand different insect species' roles in crop and other flowering plant pollination. Traditional methods of insect monitoring are straightforward to conduct but are time-consuming and labour intensive. The use of human labour for traditional sampling may unintentionally bias results \citep{dennis2005,simons1999}, increase processing lead times, reduce reproducibility, and inhibit or interfere with active pollination monitoring conducted simultaneously in different areas of a site. Furthermore, conventional sampling methods lack functional precision – the capacity to model pollinator movements, motion paths and spatial distributions. This restricts their value as a means to understand how insect behaviour effects pollination. Automated and detailed pollination monitoring techniques with high functional precision are needed that allow continuous assessment of pollination levels. Mechanised efforts to count insects have been attempted and improved over the last century, although it is only with improved technology and Artificial Intelligence that individual recognition in complex environments has started to emerge as a realistic proposition \citep{odemer2022approaches}. In turn, this will facilitate the efficient management of pollinator resources as agriculture increasingly embraces data-driven, AI-enhanced technology \citep{10.1371/journal.pone.0251572, 10.1038/s41598-021-82537-1, breeze2020}. 

Improvement in sensor technology has enabled the use of inexpensive Internet of Things (IoT) devices, such as cameras and miniature insect-mounted sensors, for pollination monitoring. Insect-mounted sensors allow movement tracking of tagged insects over large areas \citep{10.1038/s41598-021-82537-1}. However, the technique is unsuitable for agriculture since tagging is laborious, it may increase insect stress or alter behaviour \citep{batsleer2020}, and it is simply impractical on a large enough scale to be relevant in this context. Camera-based pollination monitoring can overcome these drawbacks by tracking untagged insects using computer vision and deep learning \citep{10.1371/journal.pone.0251572,Ratnayake_2021_CVPR}.

In this research, we introduce a novel computer vision system to facilitate pollination monitoring for large-scale agriculture. Our system is comprised of edge computing multi-point remote capture of unmarked insect video footage, automated offline multi-species motion tracking, as well as insect counting and behavioural analysis. We implemented and tested our methods on a commercial berry farm to (i) track individual movements of multiple varieties of unmarked insect, (ii) count insects, (iii) monitor their flower visitation behaviour, and (iv) analyse contributions of different species to pollination. Along with this article we publish the monitoring software, a dataset of over 2000 insect tracks of four insect classes, and an annotated dataset of images from the four classes. We believe that these will serve as a benchmark for future research in \emph{precision pollination}, a new and important area of precision agriculture.

The remainder of the paper is organised as follows. In Section \ref{sec:related_work} we present a brief overview of related work concerning computer vision for insect tracking in the wild. Section \ref{sec:methods} presents our new methods and their implementation. In section \ref{sec:results} we describe experiments to evaluate the performance of our approach and present the results of a pollination analysis to demonstrate our methods' application. In Section \ref{sec:discussion} we discuss the strengths and limitations of our approach and suggest future work. Section \ref{sec:conclusion} concludes the paper.

\section{Related Work}
\label{sec:related_work}

Recently there has been an increase in the use of computer vision and deep learning in agriculture \citep{kamilaris2018deep, odemer2022approaches}. This has been prominent in land cover classification \citep{lu2017cultivated}, fruit counting \citep{afonso2020tomato}, yield estimation \citep{koirala2019deep}, weed detection \citep{su2021data}, beneficial and insect pest monitoring \citep{amarathunga2021methods}, and insect tracking and behavioural analysis \citep{hoye2021}. Applications of insect tracking and behavioural analysis algorithms are usually confined to controlled environments such as laboratories \citep{Branson2009, Perez-Escudero2014, walter2021trex, Haalck2020}, and semi-controlled environments such as at beehive entrances \citep{campbell2008, magnier2019a, yang2018}. In these situations, image backgrounds and illumination under which insects are tracked vary only a little, simplifying automated detection and tracking tasks. Pollination monitoring of crops however, may require tracking unmarked insects outdoors in uncontrolled environments subjected to vegetation movement caused by the wind, frequent illumination shifts, and movements of tracked and non-target animals. These environmental changes, combined with the complexity of insect movement under such variable conditions, increases the difficulty of the tracking problem. Recent studies attempted to address these issues through in-situ insect monitoring algorithms \citep{bjerge2021real, bjerge2021automated}, but were limited in the spatiotemporal resolution required for efficient pollination monitoring. 

To overcome the difficulties listed above, we previously presented a Hybrid Detection and Tracking (HyDaT) algorithm \citep{10.1371/journal.pone.0239504} and a Polytrack algorithm \citep{Ratnayake_2021_CVPR} to track multiple unmarked insects in uncontrolled conditions. \textcolor{black}{The HyDaT algorithm uses a hybrid detection model consisting of a deep learning-based object detection model (YOLOv2 \citep{redmon2016yolo9000}) and a foreground/background segmentation-based detection model (K-nearest neighbours \citep{zivkovic2006efficient}) to track individual insects. In HyDaT, the deep learning object detection is used to detect insects at their first appearance in the frame. The foreground/background segmentation is used to detect insects' position in subsequent frames, provided that there are no multiple detections in the foreground. If the environment is too dynamic and the foreground/background segmentation cannot accurately identify the position of the insect, the deep learning model is used for the detection. This enables tracking unmarked and free-flying insects amidst the changes in the environment. The Polytrack algorithm \citep{Ratnayake_2021_CVPR} extended methods in HyDaT to track multiple insects simultaneously. In addition, Polytrack includes a low resolution mode to improve its video processing speed.}

Although previous algorithms enable tracking unmarked and free-flying insects amidst the changes in the environment, they are limited to one species and one study location at a time. To gain a sophisticated understanding of agricultural pollination, these constraints are limiting since analysis of the behaviour of multiple insect species that contribute simultaneously, in multiple locations, to overall pollination levels or deficiencies is important \citep{garibaldi2020a, rader2016}. Currently there is no computer vision facilitated system, or any other practical system, capable of achieving this goal. In addition, no previous method can identify and classify insect pollination behaviour across large-scale industrial agricultural areas at a level of detail that permits sub-site-specific interventions to increase farm yield via improved pollination.


\section{Methods and Implementation}
\label{sec:methods}

\begin{figure*}[h!]
\centering
  \includegraphics[width=\linewidth]{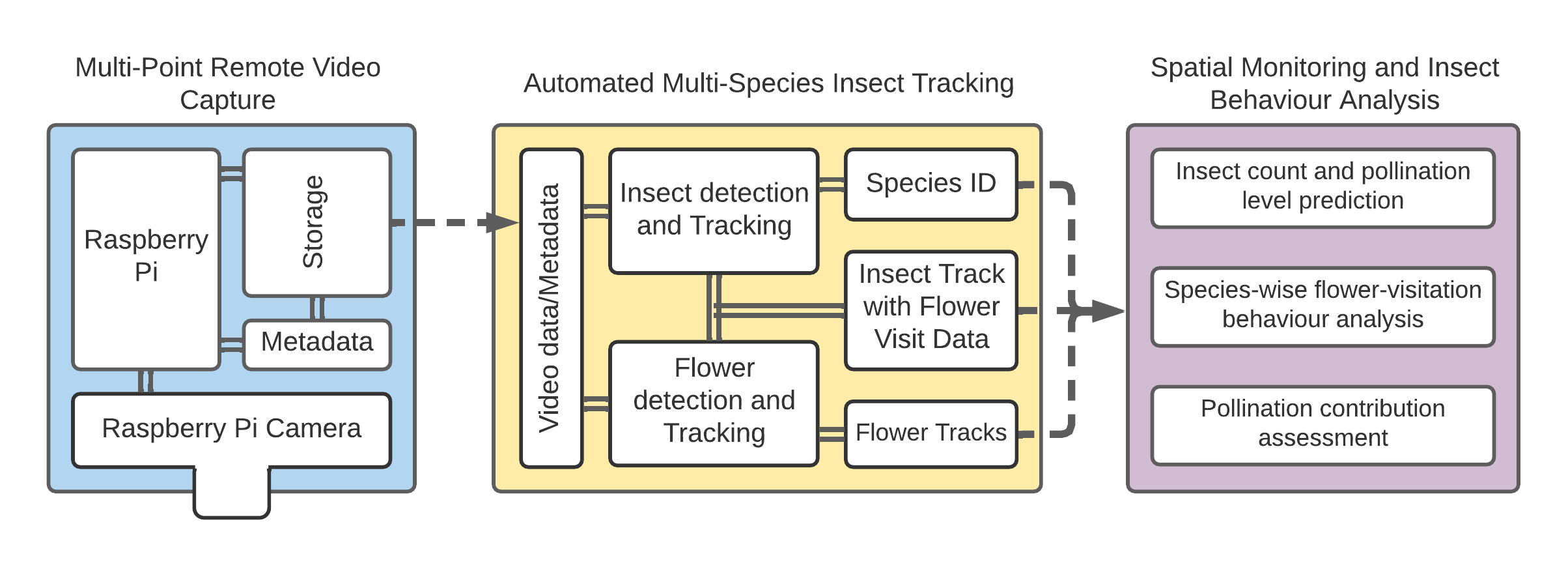}%
  \caption{\textbf{Overview of the proposed methodology} \label{fig:system_overview}}
  
\end{figure*}

In this section, we explain the methods and implementation of our insect and pollination monitoring system. An overview of the proposed methodology is shown in Fig. \ref{fig:system_overview}.

\subsection{Multi-point remote video capture}

Video footage of freely foraging, unmarked insects required for insect tracking and behavioural analysis was collected using edge computing-based remote camera trap devices built on the Raspberry Pi single board computer. We used a Raspberry Pi 4 and Raspberry Pi camera v2 (Sony IMX219 8-megapixel sensor) because it is widely available, customisable, there's a wide range of plug-in sensors, and it is sufficiently low-cost for replication across a large area \citep{jolles2021broad}. Videos are recorded at $1920 \times 1080$ resolution at $30 fps$, which is the maximum possible frame-rate for $1920 \times 1080$ resolution on our devices. The system is powered using a $20000 mAh$ battery bank. However, we do not process videos to track pollinators \textit{in situ} since the Raspberry Pi is currently incapable of processing high quality videos in real-time, and our key goals required detection of insects. Reducing the video resolution or the capture frame-rate to compensate for the lack of speed of the device is not currently feasible within the limitations imposed by pollinator insect speed and size. Video recording units were distributed across nine data collection points in an experimental site (section \ref{implementation} below) and were programmed to continuously record sets of footage clips of 10 minutes duration. The caption of each video clip contained metadata on camera location, recording date and recording time. (Refer to code availability for the software used in the video recording unit.)

\subsection{Automated multi-species insect tracking}
\label{sec:tracking}

We processed the videos captured remotely using an offline automated video processing algorithm. Since food crops are usually grown in uncontrolled or semi-controlled environments subject to changes in illumination and foliage movement caused by wind and/or insect and human activity, robust tracking of insects and flowers is essential for accurate pollination and insect behavioural analysis. \textcolor{black}{Here, we build on methods presented in HyDaT \citep{10.1371/journal.pone.0239504} and Polytrack \citep{Ratnayake_2021_CVPR} algorithms to develop an automated algorithm to track multiple insect varieties simultaneously and detail their interactions with flowers. Our algorithm uses a hybrid detection model (adopted from HyDaT \citep{10.1371/journal.pone.0239504}) consisting of a YOLOv4 \citep{bochkovskiy} deep learning-based object detection model and a K-nearest neighbours \citep{zivkovic2006efficient} foreground/background segmentation model to detect and identify insects in videos. Detected insect positions are formed into a coherent trajectory using the methods proposed in Polytrack \citep{Ratnayake_2021_CVPR}. This includes a low-resolution processing mode that rapidly processes videos when no insects are being tracked. In addition, we introduce two novel algorithms to track flowers and identify insect-flower interactions that enable insect behaviour analysis.} In the following sections we present the technical details of our methods.

At the start of processing each video sequence, our algorithm extracts the time and location at which the video was captured from the sequence's embedded metadata. Next, the video is processed to track movement of insects and their interactions with flowers. Pilot research revealed that the position of each respective flower being recorded varies throughout a day due to wind and farm management activities, and flowers may physically move termed heliotropism in some cases to track sunlight \citep{kevan1975sun, van2019thermal}. Therefore, it is essential to track flower position within the frame to reliably identify insect-flower interactions. The positions of all visible flowers are \textcolor{black}{detected and} recorded at the start of a video sequence \textcolor{black}{using the deep learning-based object detector in the hybrid detection model. The deep learning model was preferred for the flower detection over a segmentation model as it can be extended to detect and identify different types of flowers in the frame. Flower positions are updated in predefined user-specified intervals. In the current implementation an update interval of 100 seconds is used}. A ``predict and detect'' approach is used to track flower movement. The predicted next position of each flower is initially identical to its current position, since the magnitude of flower movement within a short interval (e.g., $\approx 100 seconds$) is assumed to be small. We then used the Hungarian algorithm \citep{kuhn1955hungarian} to associate the predicted position of each flower to a flower detection in order to form a continuous flower movement track. If a flower being tracked is undetected in a given frame, the last detected position is carried forward. If a detected flower cannot be assigned to any predictions it is considered to be a new flower. At the end of a video sequence, the final positions of flowers and their respective tracks of interacting insects are saved for later pollination analysis and visualisation.

When an insect \textcolor{black}{first enters the video frame, the deep learning-based object detector of the hybrid detection model detects its position and identifies its species}. In addition, it saves a snapshot of the insect for (optional human) visual verification. After detection and identification of an insect, \textcolor{black}{it is tracked through subsequent frames using the methods presented in the Polytrack algorithm \citep{Ratnayake_2021_CVPR}}. In each frame after the first detection of an insect, its position is compared with the position of recorded flowers to identify flower visits. If an insect is detected inside the radius of a flower for more than 5 consecutive frames (at 30 fps this ensures it is not flying over the flower at typical foraging flight speeds \citep{spaethe2001visual}), the spatial overlap is stored as a flower visit. The radius of a flower is computed to include its dorsal area and an external boundary threshold. This threshold is incorporated as some insects station themselves outside of a flower while accessing nectar or pollen. Repeat visits to a flower that occur after an intermediate visit to another flower are recorded as flower re-visits. \textcolor{black}{When an insect exits the video frame, the corresponding track is analysed to identify whether it originated from a false positive detection made by the deep learning model. If a track has not visited flowers and the length is less than a predefined threshold value (10 pixels $\approx$ minimum radius of a flower), it is considered a false positive.} After the verification, a file with data on camera location, time of capture and insect trajectories with flower visitation information is saved for behavioural analysis. The software and recommended tracking parameter values are available with the source code.

\subsection{Insect behaviour analysis}

We analysed insect flower visiting behaviour using the extracted movement trajectories to infer likely pollination events. This is appropriate since flowers have evolved structures that enable visiting insects to conduct pollen dispersal and transfer between floral reproductive organs for fertilisation of ovules by pollen \citep{real2012pollination}. \textcolor{black}{Metrics} used to analyse flower visitation behaviour and pollination are presented below.

Let $S=\{s^1, s^2, ..., s^{\lvert S \rvert}\}$ and $F$ be the set of insects belonging to different species (or varieties at any taxonomic level) and the set of flowers in the experimental environment respectively. Here, $s^i = \{s^i_1,s^i_2, ..., s^i_{\lvert s^i\rvert} \}$ denotes the subset of insects in $S$ that belong to the $i^{th}$ species type, and $s^i_j$ is the $j^{th}$ insect in $s^i$. \textcolor{black}{Here, if an insect exits a video frame (i.e., it flies out of the camera view or under vegetation) and later reappears, it will be counted as a new insect.} $\lvert .\rvert$ is the cardinality of a given set -- e.g., $\lvert S \rvert$ is the number of species types, $\lvert s^i \lvert$ is the number of insects belonging to the $i^{th}$ species.   

\begin{itemize}

\item The number of flowers visited by an insect species $s^i$ is defined as $FV(s^i)$, where $n_{fs^i_j}$ is the number of times insect $s^i_j$ of species $s^i$ visited flower $f\in F$.
    
    \begin{equation}
        FV(s^i) = \sum_{j = 1}^{\lvert s^i\rvert}\sum_{f \in F}n_{fs^i_j} 
    \end{equation}
    
\item Total number of visits to a flower $f$ from species $s^i$ is defined as $VF(f,s^i)$.
    
    \begin{equation}
        VF(f,s^i) = \sum_{j = 1}^{\lvert s^i \rvert}n_{fs^i_j}
    \end{equation}
    
\item Total number of visits to a flower $f$ is defined as $V(f)$.
    \begin{equation}
        V(f) = \sum_{i=1}^{\lvert S \rvert}\sum_{j = 1}^{\lvert s^i\rvert}n_{fs^i_j} 
    \end{equation}
    
\item Number of flowers fertilised with visits from species $s^i$ is defined as $N_{pol}(s^i)$, where $\hat{V}$ is the number of visits required \textcolor{black}{to fully fertilise a flower}.
    \begin{equation}
        N_{pol}(s^i) = \sum_{f\in F} [VF(f,s^i) \geq \hat{V}]
    \end{equation}

\item Total number of fertilised flowers in a location defined as $N_{pol}$. 
    \begin{equation}
        N_{pol} = \sum_{i = 1}^{\lvert S \rvert}\sum_{f\in F} [VF(f,s^i) \geq \hat{V}]
    \end{equation}

\end{itemize}

\subsection{Implementation}
\label{implementation}

\begin{figure*}[h!]
\centering

  \includegraphics[width=0.97\linewidth]{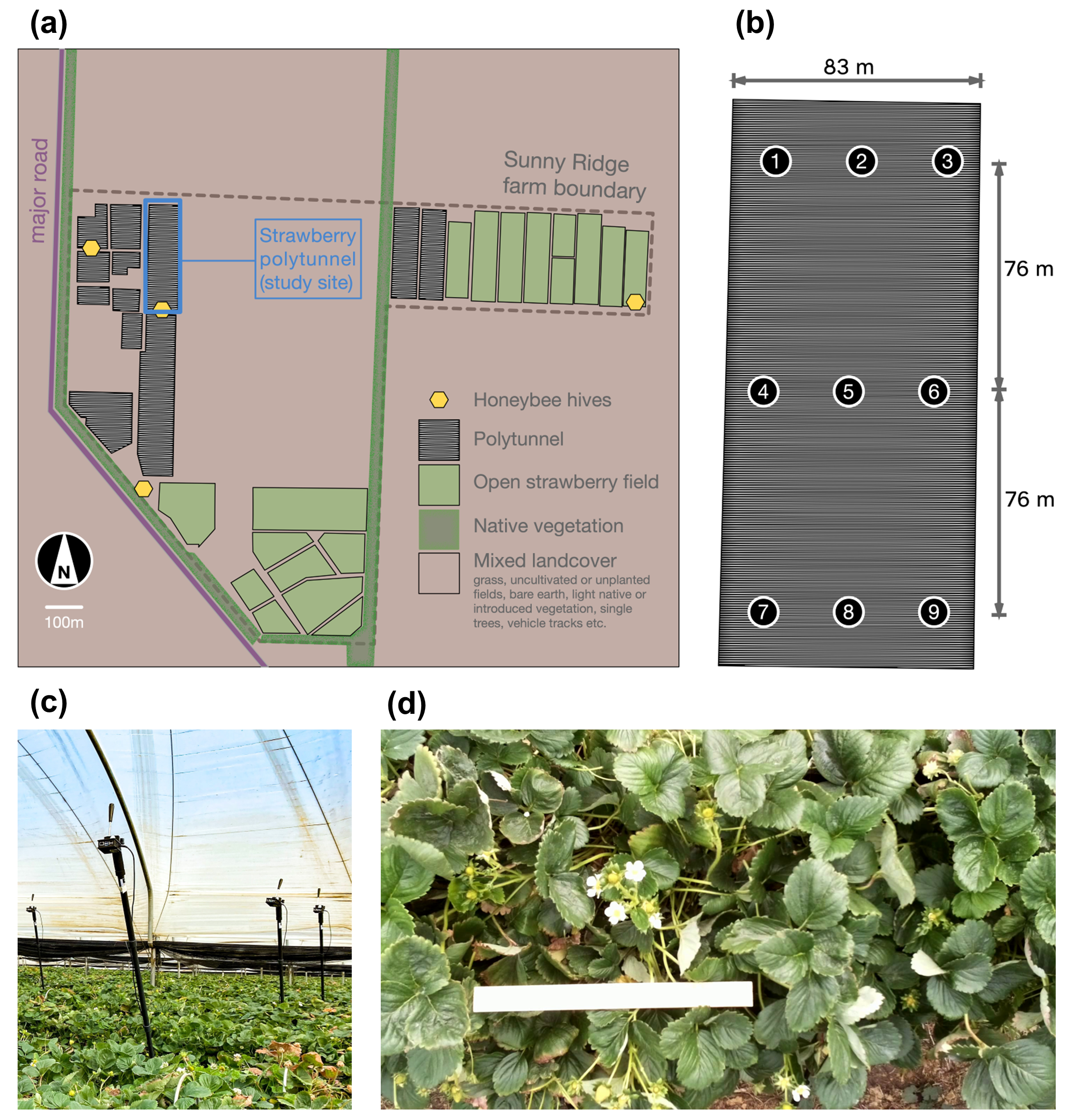}
  \caption{\textbf{Implementation of the pollination monitoring system}. (a) A map of the Sunny Ridge berry farm (implementation site near the city of Melbourne, Victoria, Australia.). Locations of managed honeybee hives are indicated with yellow \textcolor{black}{hexagons}. (b) Nine data collection points in strawberry polytunnels. (c) Edge computing-based remote video capture units placed over strawberry vegetation. (d) A sample image to indicate the field of view captured by a monitoring unit. (The white ruler measures 31 cm end-to-end). \label{fig:setup_overview}}
  
\end{figure*}

We implemented the proposed spatial monitoring and insect behavioural analysis system on the commercial Sunny Ridge farm in Boneo, Victoria, Australia ($ lat. \  38.420942 \degree \ S$, $long. \ 144.890422 \degree \ E$) (Fig. \ref{fig:setup_overview}a). Sunny Ridge grows strawberries in polytunnels covered with translucent LDPE diffusing plastic and in open fields. We installed remote video recording units over nine data collection points in strawberry polytunnels (Fig. \ref{fig:setup_overview} b) \textcolor{black}{ and manually adjusted camera lenses to focus on strawberry flowers}. These data collection points were selected to cover the edges and central regions of the polytunnels because previous studies indicated that edge effects might impact insect movement, foraging behaviour and numbers within polytunnels \citep{hall2020, 10.1371/journal.pone.0251572}. Videos were recorded for a period of 6 days (\textcolor{black}{between} $8^{th}$ - $17^{th}$ March 2021) from $11:00 am$ to $4:00 pm$ ($\approx 5$ hours) to coincide with the key pollination period. The video frames covered an area of $ \sim 700mm \times \sim 400mm$ which is the width of a planted strawberry row at the site (Fig. \ref{fig:setup_overview}d). 

\begin{figure}[h!]
     \centering
     \begin{subfigure}[b]{0.45\linewidth}
         \centering
         \includegraphics[width = \linewidth]{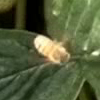}
         \caption{}
         \label{fig:honeybee}
     \end{subfigure}
     \begin{subfigure}[b]{0.45\linewidth}
         \centering
         \includegraphics[width = \linewidth]{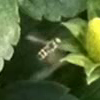}
         \caption{}
         \label{fig:hoverfly}
     \end{subfigure}
     \begin{subfigure}[b]{0.45\linewidth}
         \centering
         \includegraphics[width = \linewidth]{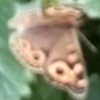}
         \caption{}
         \label{fig:moth}
     \end{subfigure}
     \begin{subfigure}[b]{0.45\linewidth}
         \centering
         \includegraphics[width = \linewidth]{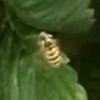}
         \caption{}
         \label{fig:wasp}
     \end{subfigure}
        \caption{\textbf{Insect pollinator types foraging on the farm.} Images of key insect types (a) Apis mellifera sp. (Hymenoptera),  (b) Syrphidae (Diptera), (c) Lepidoptera, and (d) Vespidae (Hymenoptera) captured using the low-cost edge computing-based remote video recording devices. }
        \label{fig:insects}
\end{figure}

The strawberry farm uses honeybees as managed pollinators but farm management staff had also observed other insects visiting crop flowers. We monitored the behaviour of four key insect types, honeybees (\textit{Apis mellifera}), Syrphidae (hover flies), Lepidoptera (moths and butterflies), and Vespidae (wasps) that actively forage on the farm (Fig. \ref{fig:insects}). Moths and butterflies were treated as a single insect pollinator class (Lepidoptera) for pollination analysis because of their relatively low numbers. 

\subsubsection{Training the deep learning model}

The automated video processing system employs a deep learning model YOLOv4 to detect insects and flowers. We created a custom dataset of 3073 images divided into four classes: (i) honeybees/Vespidae (2231/371 instances), (ii) Syrphidae (204 instances), (iii) Lepidoptera (93 instances), and (iv) strawberry flowers (14050 instances). Honeybees and Vespidae were included in a single Hymenopteran class due to their physical similarities and the difficulty of automatically distinguishing between them using the low-quality video footage extracted from the basic cameras (discussed further below). The prepared dataset was \textcolor{black}{manually} annotated with bounding boxes using the Computer Vision Annotation Tool \citep{cvat}. \textcolor{black}{When annotating small insects such as Syrphidae, videos associated with annotation images were carefully referenced to minimise the possibility of false negative annotations.} The YOLOv4 model was then trained on this dataset using TensorFlow \citep{abadi2016tensorflow} with a learning rate of 0.001. The pretrained YOLOv4 model and its evaluation data are available with the software code. 

\subsubsection{Processing videos}

We processed \textcolor{black}{all recorded videos} to extract insect tracks and insect-flower visiting behaviour using the methods described in Section \ref{sec:tracking}. Videos were processed on the MASSIVE high performance computing infrastructure \citep{goscinski2014multi} with Intel Xeon Gold 6150 (2.70 GHz) CPU, 55 GB RAM, NVIDIA Tesla P4 GPU and CentOS Linux (7).

\subsubsection{Insect trajectory dataset preparation}

We post-processed insect tracks extracted from the videos to correct insect type identifications. Insect type identification was performed on multiple still frames of each insect assigned to a motion track. A further step was appended to this process to manually classify Hymenoptera into two separate classes, honeybees and Vespidae. As reported above, these insects were initially treated as a single class in training the deep learning model due to the difficulty of clearly resolving morphological differences between them in flight at low video resolution and 30 fps. \textcolor{black}{If the insect type could not be confidently identified through still images, the insect was classified based on its movement behaviour after observing the videos (e.g. if the insect visited flowers, it was identified as a honeybee as opposed to a Vespidae since relevant Vespids are considered in study conditions to be predatory insects \citep{10.1016/j.fooweb.2020.e00144}). Trajectories that originated through detections that do not correspond to insects were identified as false positives and removed during this process.}

\section{Results}
\label{sec:results}

\subsection{Experimental evaluation}

We evaluated the performance of our system for extracting the trajectory and flower visitation behaviour of four insect types (Fig. \ref{fig:insects}). Experiments were conducted using a test dataset of $180,000$ frames/$100$ minutes at 30 frames per second (comprised of 10 sequential videos of 10 minutes each). These videos were randomly selected from the set of recordings unused in deep learning model training and captured from different polytunnel locations (Test video dataset is accessible from Data Availability).

\textcolor{black}{We measured the detection accuracy of our algorithm by calculating precision (Equation \ref{eqn:precision}), recall (Equation \ref{eqn:recall}), and $F_{score}$ (Equation \ref{eqn:Fmeasure}) \textcolor{black}{metrics} \citep{deoliveirabarreiros2021} for tracked insects and flowers}.  

\begin{equation}
    Precision = \frac{True Positive}{True Positive + False Positive}
    \label{eqn:precision}
\end{equation}

\begin{equation}
    Recall = \frac{True Positive}{True Positive + False Negative}
    \label{eqn:recall}
\end{equation}

\begin{equation}
    F_{score} = \frac{2 \times (Recall \times Precision)}{Recall + Precision}
    \label{eqn:Fmeasure}
\end{equation}

where, \textcolor{black}{$True Positive$ is the total number of correctly detected insect positions in a track. A detection was considered correct if the algorithm recorded the position of an insect in an area that was in fact covered by the body of the insect.} $False Negative$ is the total number of undetected insect positions and $False Positive$ is the total number of incorrectly detected insect positions in a track.

\begin{figure*}[h!]
\centering
  \includegraphics[width=\linewidth]{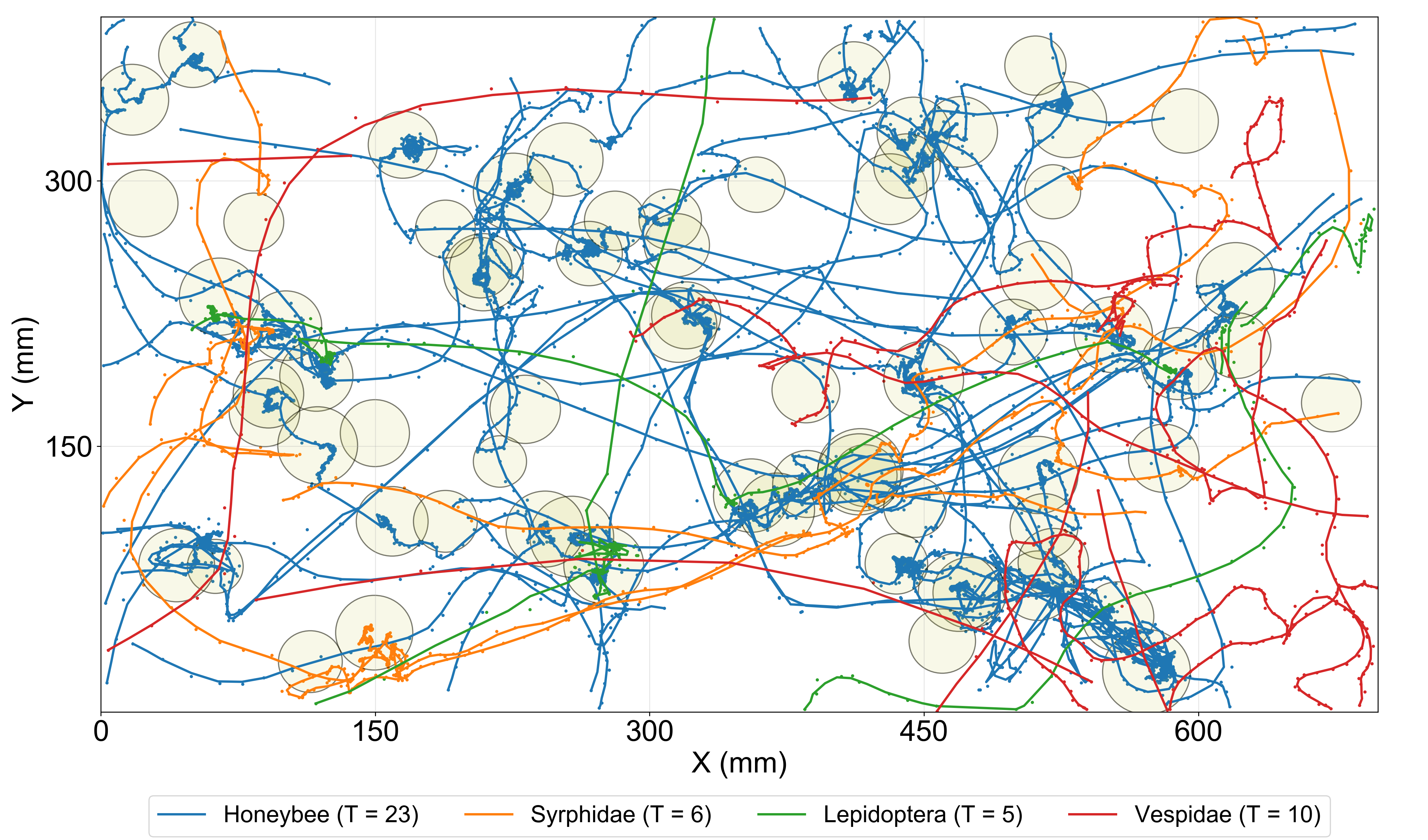}%
  \caption{\textbf{Trajectories of insects and flower positions recorded in test videos.} Track colour indicates insect variety. The number of tracks recorded for each insect type is shown in the legend in brackets beside insect type. Flower locations are circled in yellow.  \label{fig:test_tracks}}
  
\end{figure*}

The tracks and flower visits reported by our system were compared against human observations made from the videos for validation as we found no other existing monitoring system against which to compare our software. Test videos were observed by playing them on VLC media player at $\times 5$ speed to record insects and flowers. When an insect appeared in the frame, the video was analysed frame by frame to record its flower visits. An insect landing on the dorsal side of a flower was counted as a flower visitor. Insects that appeared inside the frame of the video for less than 5 frames were ignored since at 30 fps this time is too brief to be likely to have any biological impact on pollination. If an insect departed the frame and later reappeared, or if it flew under the foliage and later reappeared, it was considered as a ``new'' insect. \textcolor{black}{Experimental results related to insect and flower detection are shown in Table  \ref{table:tracking} and results on flower-visit detection are presented in Table \ref{table:flower_visits}.} \textcolor{black}{Fig}. \ref{fig:test_tracks} shows the trajectories of insects recorded in test videos. \textcolor{black}{A detailed description of experimental results is available in Supplementary Information}.

\begin{table*}[h]
\centering
\caption{\textcolor{black}{\textbf{Results of the evaluations of detections for test video dataset.} ``No. of Obs.'' and ``Visible Frames'' shows the number of insects/flowers and number of frames in which the insect/flower were fully visible as observed through human observations. ``Trackletts Generated'' shows the number of trackletts generated by the algorithm for each variety. ``Track Evaluation'' categorises the trackletts generated to TP=True Positive, FN = False Negative, FP = False Positive, and IS = Identity Swaps. Multiple tracks generated by a single insect are considered as Identity Swaps. ``Evaluation Metrics'' present the average precision, recall and F-score metrics for tracked insects. A detailed description of experimental results is available in Supplementary Information.}}
\label{table:tracking}
\small
\renewcommand{\arraystretch}{1.2}
\resizebox{\textwidth}{!}{
\begin{threeparttable}
\begin{tabular}{|c|c|c|c|c|c|c|c|c|c|c|} 
\hline
\multirow{2}{*}{\begin{tabular}[c]{@{}c@{}}\textbf{Insect/ }\\\textbf{ Flower}\end{tabular}} & \multirow{2}{*}{\begin{tabular}[c]{@{}c@{}}\textbf{No. of}\\\textbf{Obs.}\end{tabular}} & \multirow{2}{*}{\begin{tabular}[c]{@{}c@{}}\textbf{Visible}\\\textbf{Frames}\end{tabular}} & \multirow{2}{*}{\begin{tabular}[c]{@{}c@{}}\textbf{Trackletts}\\\textbf{Generated }\end{tabular}} & \multicolumn{4}{c|}{\textbf{\textcolor{black}{Track Evaluation}}} & \multicolumn{3}{c|}{\textbf{\textbf{Evaluation \textcolor{black}{Metrics}}}} \\ 
\cline{5-11}
 &  &  &  & \textbf{TP} & \textbf{FN} & \textcolor{black}{\textbf{FP}} & \textbf{\textcolor{black}{IS}} & \textbf{Precision} & \textbf{Recall} & \textbf{F-score} \\ 
\hline
Honeybee & 20 & 16846 & 23 & 20 & 0 & \textcolor{black}{0} & \textcolor{black}{3} & 0.99  & 0.92  & 0.95\\
Syrphidae & 6 & 3436 & 6 & 5 & 1 & \textcolor{black}{0} & \textcolor{black}{1} & 1.00  & 0.71 & 0.81 \\
Lepidoptera & 4 & 3158 & 5 & 3 & 1 & \textcolor{black}{0} & \textcolor{black}{2} & 0.99  & 0.71 & 0.81\\
Vespidae & 10 & 589 & 10 & 10 & 0 & \textcolor{black}{0} & \textcolor{black}{0} & 1.00 & 0.73 & 0.83 \\
\textcolor{black}{Flower$^{\ast}$}& 72 & 179306 & 68 & 68 & 4 & \textcolor{black}{0} & \textcolor{black}{0} & \textcolor{black}{1.00} & \textcolor{black}{1.00} & \textcolor{black}{1.00} \\
\hline
\end{tabular}
\begin{tablenotes}
    \small
    \item \textcolor{black}{$^{\ast}$ Flower positions were detected and recorded at 100 second (= 3000 frame) intervals.}
\end{tablenotes}

\end{threeparttable}}
\end{table*}

\begin{table}[h!]
\centering
\caption{\textcolor{black}{\textbf{Results of the experimental evaluations for flower visit detections for the test video dataset.} ``Observed Visits'' shows the total number of insect visits to flowers counted through human observations. ``Visit Detection Evaluation'' shows the evaluation of flower visits automatically identified through the software for tracked insects. TP = True positive, FP = False positive, FN = False-negative.}}
\label{table:flower_visits}

\renewcommand{\arraystretch}{1.2}
\resizebox{\linewidth}{!}{
\begin{threeparttable}
\begin{tabular}{|c|c|c|c|c|} 
\hline
\multirow{2}{*}{\begin{tabular}[c]{@{}c@{}}\textbf{Insect}\\\textbf{Type}\end{tabular}} & \multirow{2}{*}{\begin{tabular}[c]{@{}c@{}}\textbf{Observed}\\\textbf{Visits}\end{tabular}} & \multicolumn{3}{c|}{\textbf{Visit Detection Evaluation}} \\ 
\cline{3-5}
 &  & \textbf{ TP } & \textbf{ FP } & \textbf{ FN } \\ 
\hline
Honeybee & 67 & 65 & 0 & $2^{\ast}$ \\
Syrphidae & 5 & 4 & 1 & 1 \\
Lepidoptera & 6 & 6 & 1 & 0 \\
Vespidae & 0 & 0 & 0 & 0 \\
\hline
\end{tabular}
\begin{tablenotes}
            \small
            \item[$\ast$] Resulted from undetected flower(s).
        \end{tablenotes}

\end{threeparttable} }
\end{table}

In our test videos, the proposed algorithm tracked honeybees with a precision of 0.99, a recall of 0.92 and an F-score of 0.95. The insect behavioural analysis component of the algorithm accurately detected $97\%$ of honeybee-flower interactions, and $3\%$ of flower interactions were not recorded due to undetected flowers. Test videos comprised six appearances of Syrphidae and the algorithm accurately detected five of them resulting in a detection rate of $83\%$. The algorithm tracked Syrphidae with high precision (1.00), but the recall rate of 0.71 and F-score of 0.81 were lower than that of honeybees. These lower values were due to the frames where the insect was undetected (see Discussion). Tracking \textcolor{black}{metrics} related to Lepidoptera were similar to that of Syrphidae, where the algorithm detected and tracked $75\%$ of Lepidopterans with precision, recall and F-score values of 0.99, 0.71 and 0.81 respectively. It also recorded all Lepidopteran flower interactions. The algorithm detected and tracked all Vespidae present in test videos with a precision rate of $1.00$. However, the recall rate and the F-score were $0.73$ and $0.83$, respectively. This was because the video frame rate was \textcolor{black}{too} low to track some high-speed Vespidae movements. \textcolor{black}{The proposed algorithm recorded identity swaps (multiple tracks generated by the same insect) for honeybees, Syrphidae and Lepidoptera. The study results did not contain false positive tracks for any insect type, as the algorithm accurately identified and discarded tracks that originated from false positive insect detections. The values of the detection evaluation metrics for flowers were high as there was little or no movement of flowers apparent within test videos.}

\subsection{Insect behavioural analysis for precision pollination} 

\begin{figure*}[h!]
     \centering
         
         \includegraphics[width = \linewidth]{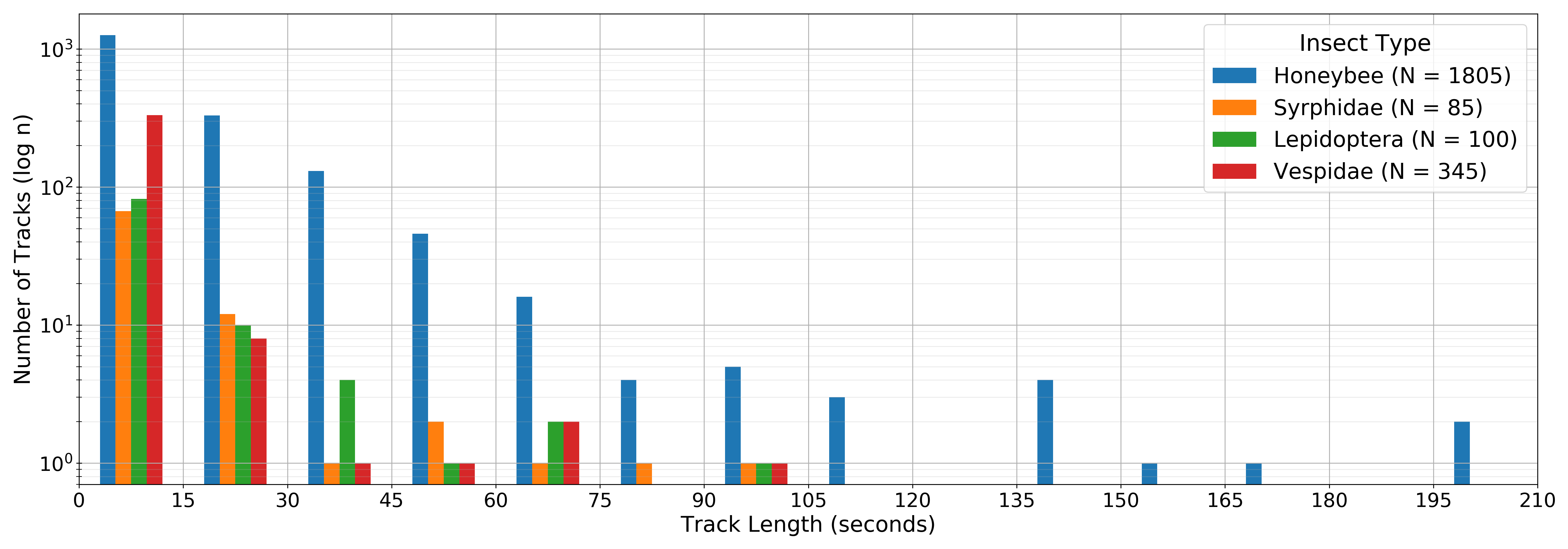}
         
    \caption{\textbf{The distribution of recorded track lengths (in seconds) for the four insect types.} ``N'' in the legend shows the total number of tracks recorded for each insect type.}
        \label{fig:track_lengths}
\end{figure*}

\begin{figure*}[h!]
     \centering
         
         \includegraphics[width = \linewidth]{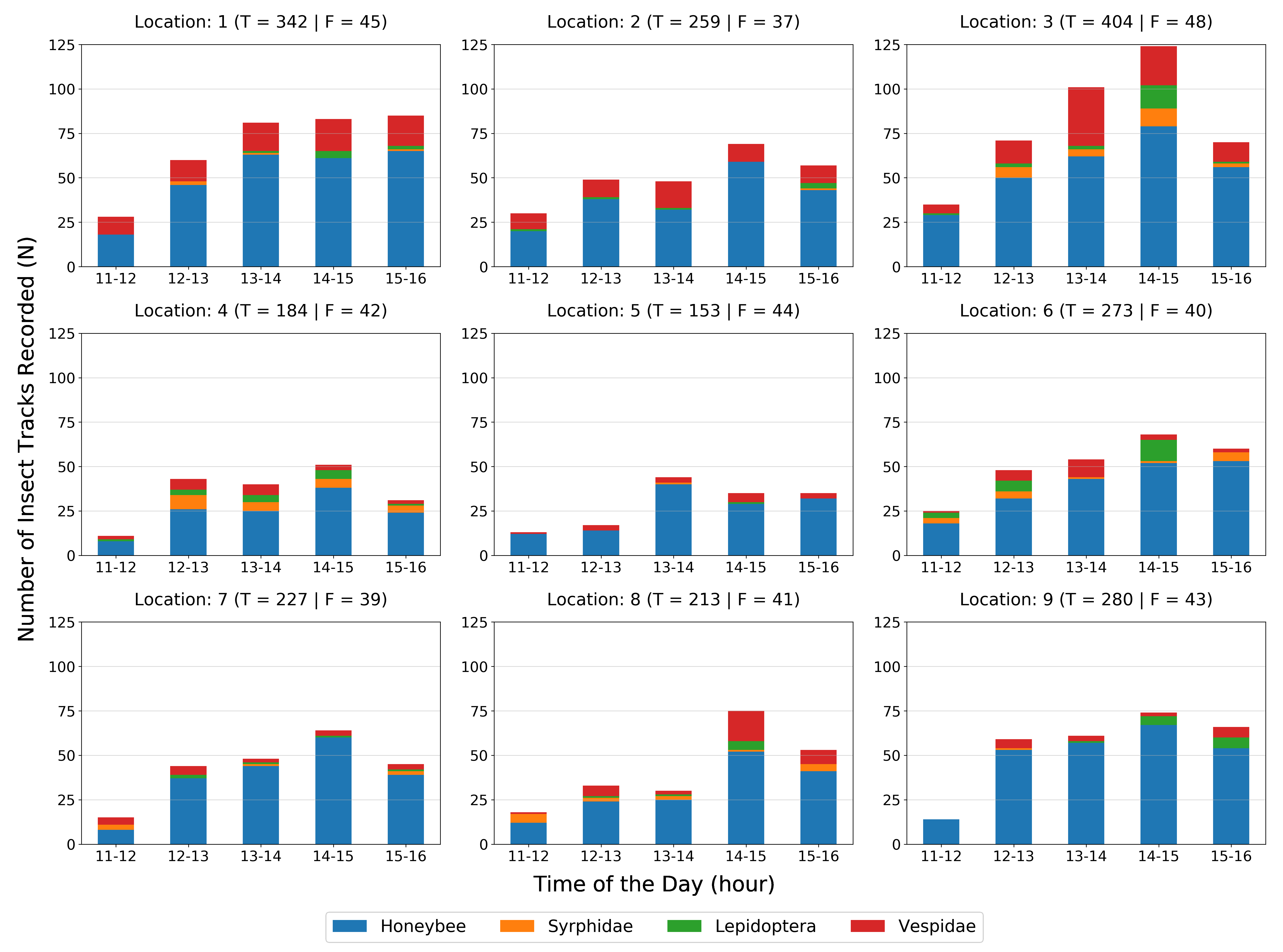}
        \caption{\textbf{Temporal variation in insect counts.} Figure show the frequency of the insects recorded in each hour of the day at each data collection point. ``T'' and ``F'' in the title blocks show the total number of tracks and flowers recorded at each location.}
        \label{fig:temporal_variation}
\end{figure*}

We applied our methods to analyse pollination in a commercial berry farm to demonstrate its practical relevance for \emph{precision pollination}. The dataset for pollination analysis consisted of \textcolor{black}{1805 honeybees, 85 Syrphidae, 100 Lepidoptera and 345 Vespids}. The complete trajectory dataset of insects and flowers is accessible from Data Availability. \textcolor{black}{The distribution of the trajectory lengths is shown in Fig \ref{fig:track_lengths}. An analysis of the temporal variations in insect counts across the data collection points is shown in Fig. \ref{fig:temporal_variation}. } 

\begin{figure*}[h!]
\centering

  \includegraphics[width=0.89\linewidth]{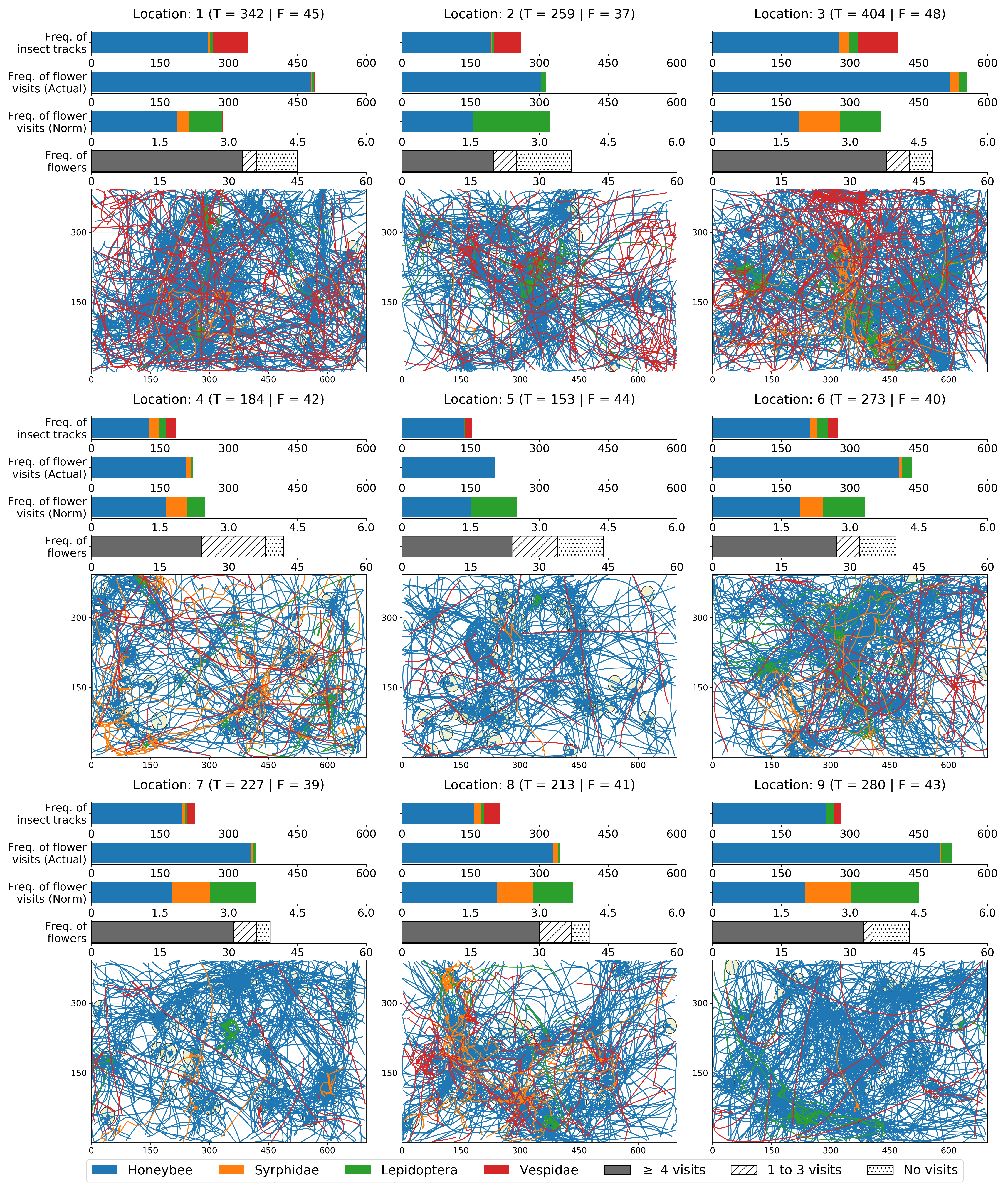}%
  
  \caption{\textbf{Results of the spatial monitoring and insect behavioural analysis for precision pollination.} Bar charts above the plots indicate the number of tracks, total number of flower visits \textcolor{black}{(actual), total number of flower visits normalised by the recorded number of tracks for each insect type at each location}, and number of flowers recorded at each location. Bar colour for tracks and flower visits indicates the proportion of tracks recorded for each insect type. Strawberry flowers typically require four visits for full fertilisation \citep{garibaldi2020a, chagnon1989}. The dark grey portion of the flowers' bar graph shows the number of flowers with over four insect visits. ``T'' and ``F'' in the title blocks are the total number of tracks and flowers recorded at each location. Trajectory plots show all insect tracks recorded at each location throughout the data collection period. Track colours represent different insect varieties. Flower locations are indicated by yellow circles. }
  \label{fig:summary_plots}
\end{figure*}

Spatial monitoring and insect behavioural analysis can help growers quantify pollination across different farm areas. We compared pollination levels across farm strawberry polytunnels using insect counts and the number of insect-flower interactions recorded at each location. Research suggests that a strawberry flower requires a minimum of four insect visits to be fully fertilised \citep{garibaldi2020a, chagnon1989}. Therefore, the number of insect visits to a flower can be used to predict its pollination level. We used the collected spatial monitoring data to identify flowers that received at least four insect visits during the biologically relevant data collection period [5 hours] over which our system operated. Analysis results are shown in Fig. \ref{fig:summary_plots}.

\begin{figure*}[h!]
\centering
  \includegraphics[width=\linewidth]{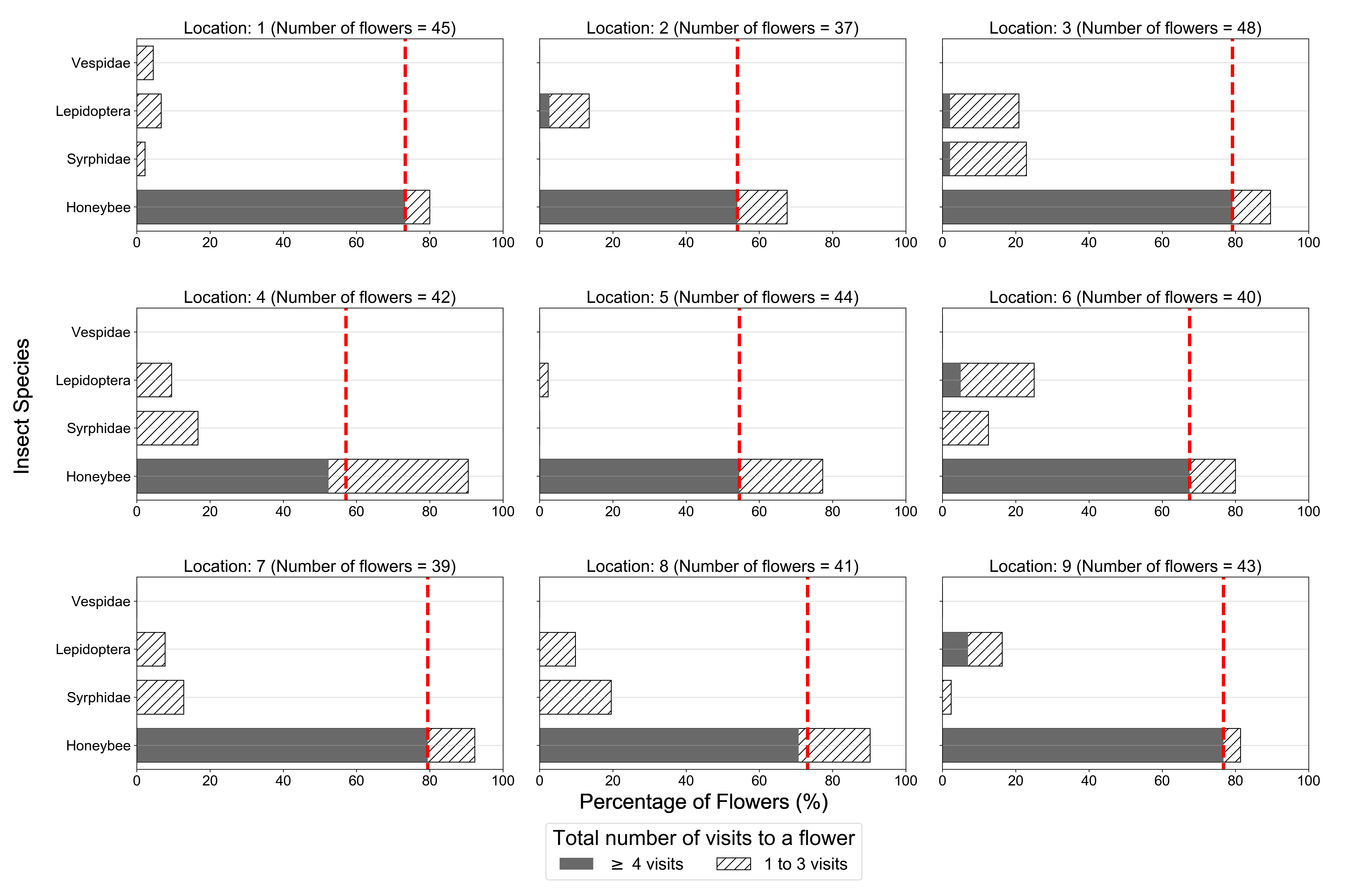}%
  \caption{\textbf{Contribution of different insect varieties towards strawberry pollination.} Bar chart shows percentage of flowers visited by each insect type. The dark grey portion shows the percentage of flowers with over four (number of visits required for strawberry flower fertilisation \citep{garibaldi2020a, chagnon1989}) from each insect type. The red dashed line in the plots show the total percentage of flowers with more than four visits in a location.  \label{fig:species_contribution}}
  
\end{figure*}

Flower-visitation behaviour reflects insects' crop pollination contributions. We quantified this on the strawberry flowers by calculating the percentage of flowers that received visits from each insect type. We further analysed insect-flower visits to evaluate the pollination efficacy of insect types by calculating the proportion of flowers that received the minimum of four insect visits required for fertilisation. Results of this analysis are shown in Fig. \ref{fig:species_contribution}.

At all data collection points, we recorded a higher number of honeybees than other insects (Fig. \ref{fig:summary_plots}). These insects contributed the most towards achieving the flower-visitation targets required for fertilisation (Fig. \ref{fig:species_contribution}). The next highest recorded insect were the Vespids (341 tracks) (Fig. \ref{fig:summary_plots}). However, Vespids were rarely observed to be visiting flowers – at location 1 we did identify Vespidae flower visits; see Fig. \ref{fig:species_contribution}. This suggests that Vespids do not contribute much to strawberry pollination. Indeed Vespids may be a predator of other insects \citep{10.1016/j.fooweb.2020.e00144} and can act to inhibit pollination. We recorded relatively low Lepidopteran and Syrphidae counts in most areas of the farm (Fig. \ref{fig:summary_plots}). The contribution of these species towards achieving flower-visitor targets required for pollination was observed to be much lower than that of honeybees (Fig. \ref{fig:species_contribution}). This effect is evident by the low relative frequency with which these insects made successive visits to flowers to meet the four required for optimal fertilisation (Fig. \ref{fig:species_contribution}). For example, the highest frequency of a non-honeybee pollinator to meet four visits was Lepidoptera at location 9 where less than 15\% of flowers achieve this level of pollination; whilst at all locations honeybees significantly exceeded this level of pollination performance (Fig. \ref{fig:species_contribution}). When pollination across all locations is considered, over 68\% of the recorded strawberry flowers received the minimum of four insect visits required for fertilisation, and 67\% of flowers attained this threshold through honeybee visits alone. This data thus reconfirms which insects seem, at least as far as the number of visits is concerned, to contribute the most towards pollination at the site.

\section{Discussion and Future Work}
\label{sec:discussion}

Insect pollination monitoring can improve our understanding of the behaviour of insects on crops. It can therefore potentially boost crop yield on farms were it not currently heavily constrained by the labour required for manual data collection. In this study, a novel multi-point computer vision-based system is presented to facilitate digital spatial monitoring and insect behavioural analysis on large scale farms. Our system operates in real-world commercial agricultural environments (Fig. \ref{fig:setup_overview}) to capture videos of insects, identify them (Fig. \ref{fig:insects}), and count the number of different varieties over large areas (Fig. \ref{fig:summary_plots}) \textcolor{black}{across time (Fig. \ref{fig:temporal_variation})}. Analysis of the insect behavioural data allows comparison of the contributions of different insect varieties to crop pollination (Fig. \ref{fig:summary_plots} and \ref{fig:species_contribution}). Here, we discuss the implications of our research for precision pollination.

\subsection{Computer vision for insect tracking and behavioural analysis}

Our methods remove the major constraints imposed by the limitations of human observers for horticultural pollination monitoring and the collection of high-resolution spatiotemporal data (Fig. \ref{fig:summary_plots}) on insect behaviour. The approach therefore also paves the way for computer vision and edge computing devices to identify insect species for other entomological and ethological applications.

The use of relatively inexpensive Raspberry Pi edge computing devices (Fig. \ref{fig:setup_overview}) for remote recording provides a high degree of scalability and customisability \citep{aslanpour2021serverless, o2019edge} for insect monitoring. However, the limited capabilities of these devices \textcolor{black}{requires manual focusing of cameras,} confines the size of recorded study areas (Fig. \ref{fig:setup_overview}d) and offers only low frame rates and low quality video. This reduced the system's ability to detect small Syrphidae, and resulted in issues with the detection and tracking of fast-moving Vespids (Table \ref{table:tracking}).  In addition, the current implementation continuously recorded videos on the Raspberry Pi even when there was no insect in the camera frame. This wastes the limited storage and power capacities available on edge computing devices. We aim to address this drawback in future work by implementing an \emph{in-situ} algorithm on the edge-computing device for real-time event processing. It is likely that with the rapid improvement of camera technology, video quality and resolution will overcome current limitations and enhance the accuracy and efficiency of our methods.

\textcolor{black}{We used a fixed camera setup covering a confined area to record videos (Fig. \ref{fig:setup_overview}d). This results in a subsampling of insect flower visitation and behavioural data. We propose that future research should address this limitation by developing methods to extend study areas using multiple or moving cameras.} We applied our new methods to monitor insect pollination behaviour in strawberry crops. Strawberry flowers bloom within a narrow vertical spatial range and are usually visible from above (Fig. \ref{fig:setup_overview}d). By contrast, other crops, such as tomatoes or raspberry, grow within complex three-dimensional structures of vines or canes, making overhead camera tracking of insects problematic. Monitoring their behaviour in such three-dimensional crops \textcolor{black}{will require multi-view video capture and three-dimensional tracking, which is currently a highly complex and unsolved challenge}.

Insect detection is an essential precursor to tracking and monitoring. Our algorithm accurately detected honeybees and Vespidae but performed relatively poorly on Syrphidae (Table \ref{table:tracking}). This is because of the relatively small pixel area covered by the insect with our setup (Syrphidae covers $\approx 40 \pm 10$ pixels compared to $\approx 1001 \pm 475$ pixels for a honeybee) (Fig. \ref{fig:insects}). Future improvements in cameras and object detection technologies \citep{stojnic2021method} will help here. 

\textcolor{black}{We used a hybrid detection model consisting of a deep learning-based and a segmentation-based detection model to detect insects in videos. Using a segmentation-based detection model in tandem reduced the demand for the deep learning model. This helped to 
achieve F-scores of 0.8 for each variety (Table \ref{table:tracking}) even when trained with a limited and unbalanced dataset \citep{10.1371/journal.pone.0239504, Ratnayake_2021_CVPR}.} Our algorithm uses deep learning to detect and classify insects as they enter the video frame. The results of experimental evaluation showed limitations in Lepidopteran detection and visually similar insect detection (i.e. honeybees, Syrphidae and Vespidae (Fig. \ref{fig:insects} and Table \ref{table:tracking})). Detection of Lepidopterans was challenging because they sometimes appear similar in shape to foliage and shadows in the environment. Also, both Lepidopterans and Syrphidae rested stationary on flowers for extended periods, prompting the algorithm to classify them as part of the background. Detection and classification of visually similar insects requires a deep learning model trained with large annotated datasets. \textcolor{black}{Although there is a considerable increase in the number of open datasets for animal classification \citep{van2018inaturalist}, there is an absence of suitable open annotated datasets for insect detection in entomology \citep{hoye2021}}. Hence, for the current study, we built a dataset from scratch. However, our dataset was unbalanced, since the number of instances in each class was influenced by the relative abundance of insects recorded at the site \citep{7727770}. We propose that future research should use characteristics of insect behaviour, such as spatial signatures of insect movement, to improve species classification tasks \citep{kirkeby2021advances}. This will help overcome limitations associated with camera quality and deep learning datasets. The insect trajectory and video data we publish with this article offers a starting point for such solutions.

\textcolor{black}{We used the extracted insect trajectory data to monitor insect flower visitation behaviour and infer pollination levels. In our analysis, if an insect departed the frame and later reappeared, or if it flew under the foliage and later reappeared, a ``new'' trajectory was generated. Although this does not affect the flower-visitor counts, our approach could bias trajectory counts for species with different behaviours or flight path characteristics (e.g., flying under foliage and being occluded, cornering sharply rather than turning smoothly). Our multi-camera system will enable future research on these topics especially when combined with emerging solutions for individual insect identification.} 

\textcolor{black}{A classic question in any data sampling is the tradeoff between focused detail and global context. Our multi-point system enables a view of what specific insect pollinators are doing at flowers and also a holistic appraisal of how pollination is enabled across an entire agricultural field. This solution can be implemented in a variety of ways by choosing different camera lens focal lengths and thus fields of view, different numbers of cameras, and any field area, depending upon the resolution required to answer a particular research question.}

\subsection{Spatial monitoring for precision pollination}  

Spatial monitoring and insect behavioural analysis can help growers understand the distribution of pollinators across a farm and their impact on pollination. We quantified pollination by counting insect numbers and insect-flower interactions (Fig. \ref{fig:summary_plots}). Farm areas with many flowers and insects will likely yield the most crop if there are a suitable number of insect-flower interactions. Strawberry flowers require at least four insect visits for full fertilisation \citep{garibaldi2020a, chagnon1989}. However, it is important to note that crop yield and visitation rates have been observed to have a non-linear relationship \citep{garibaldi2020a}, where higher flower visitation rates can result in lower crop yield \citep{rollin2019impacts, garibaldi2020a}. Therefore, it is beneficial to maintain insect flower visits at an optimum value that depends on the crop type, pollinator species, and environmental conditions \citep{garibaldi2020a}.

 Although different behaviours and morphologies make some insect species more effective pollinators of some flowers than others, we compared the contribution of different insect varieties to strawberry pollination using the number of insect flower visits as a proxy (Fig. \ref{fig:species_contribution}). The analysis suggests that strawberries can obtain sufficient pollination solely from honeybees (Fig. \ref{fig:species_contribution}), even without the presence of other insects. \textcolor{black}{Whilst non-honeybee insect species do not reach the threshold of four visits for high effectiveness (Fig. \ref{fig:species_contribution}), it is possible these insects may still contribute to pollination (Fig. \ref{fig:summary_plots} and \ref{fig:species_contribution}). Indeed, the absolute volume of insects present may impact how thresholds are achieved. Employing the computer vision solutions we share here in different environments where insect abundance naturally varies will likely inform researchers about what insects are most beneficial in specific contexts. In addition,} an agricultural system driven by a single pollinator type may not be desirable. Pollinator diversity and associated high flower visitor richness have been shown to affect pollination and crop yield \citep{garibaldi2016mutually}. Often the high abundance of a single pollinator species cannot be used as a substitute for species richness \citep{garibaldi2016mutually, fijen2018insect} as variations in behaviour and foraging inherent to different insect species may be important. 

Compared to manual pollination monitoring, our methods provide high-resolution \textcolor{black}{spatio-temporal} behavioural data classified by insect type. Our spatial monitoring results (Fig. \ref{fig:summary_plots}) can assist farm managers to identify farm areas that require immediate attention in order to maximise fruit set. \textcolor{black}{The temporal analysis of variations in insect counts (Fig. \ref{fig:temporal_variation}) can be used as a guide to understand which duration or sampling frequency is necessary for a site to understand pollinator behaviour.} Furthermore, the behavioural pollination contribution analysis (Fig. \ref{fig:species_contribution}) can provide tools and data to identify efficient pollinator species for a particular crop, enabling data-driven pollination management.
 
 Pollination monitoring helps understand the impact of climate change and other anthropogenic activities on insect populations \citep{settele2016climate}. Recently, climate change and other anthropogenic pressures, including intensive agriculture, have caused a decline in some pollinator populations \citep{ecosystemadam, schweiger2010multiple, hallmann2017more, outhwaite2022agriculture} threatening global food security and terrestrial ecosystem health. The most impacted pollinator populations are native and wild insects that must compete for food with managed pollinators while coping with disease, pollution and habitat loss \citep{wood2020managed}. Digital pollination monitoring systems like that described here, provide much-needed data for understanding the impacts of climate change on insect biodiversity and can ultimately provide a sound basis for conservation.

\section{Conclusions}
\label{sec:conclusion}

In this paper, we presented a computer vision facilitated system for spatial monitoring and insect behavioural analysis to underpin agricultural precision pollination. Our system comprised of edge computing-based remote video capture, offline, automated, unmarked multi-species insect tracking, and insect behavioural analysis. The system tracked four insect types with F-scores above 0.8 when implemented on a commercial strawberry farm. Analysis of the spatial distribution of flower-visiting behaviour of different insect varieties across the farm, allowed for the inference of flower fertilisation, and the comparison of insects' pollination contribution. We determined that $67\%$ of flowers met or exceeded the specified criteria for reliable pollination through honeybee visits. However, alternative pollinators were less effective at our study site. This advancement of computer vision, spatial monitoring and insect behavioural analysis, provides pollinator data to growers much more rapidly, broadly and deeply than manual observation. Such rich sources of insect-flower interaction data potentially enable precision pollination and pollinator management for large-scale commercial agriculture.



\backmatter

\bmhead{\textcolor{black}{Supplementary information}}

\textcolor{black}{Additional and detailed experimental results on presented methods are available in the Supplementary Information file ``Supplementary\_Information.pdf''.}



\bmhead{Acknowledgments}

The authors would like to thank Sunny Ridge Australia for the opportunity to conduct research at their farm. 



\section*{Declarations}


\begin{itemize}
\item Funding: Authors were supported by the Australian Research Council Discovery Projects grant DP160100161 and Monash-Bosch AgTech Launchpad primer grant. This study was funded by AgriFutures grant PRJ-012993. Amarathunga is supported by ARC Research Hub IH180100002.
\item Competing interests: The authors have no competing interests to declare that are relevant to the content of this article.
\item Ethics approval: Not applicable
\item Consent to participate: Not applicable
\item Consent for publication: Not applicable
\item Availability of data and materials: The datasets generated during and/or analysed during the current study are available in the repository \url{https://doi.org/10.26180/21533760}
\item Code availability: Code is available through \url{https://github.com/malikaratnayake/Polytrack2.0}
\item Authors' contributions: Conceptualization: Malika Nisal Ratnayake, Adrian G. Dyer, Alan Dorin; Data curation: Malika Nisal Ratnayake; Formal analysis: Malika Nisal Ratnayake; Funding acquisition: Adrian G. Dyer, Alan Dorin; Investigation: Malika Nisal Ratnayake, Don Chathurika Amarathunga, Asaduz Zaman; Methodology: Malika Nisal Ratnayake, Adrian G. Dyer, Alan Dorin; Project administration: Adrian G. Dyer, Alan Dorin; Resources: Adrian G. Dyer, Alan Dorin; Software: Malika Nisal Ratnayake; Supervision: Adrian G. Dyer, Alan Dorin; Validation: Malika Nisal Ratnayake, Don Chathurika Amarathunga; Writing – original draft: Malika Nisal Ratnayake; Writing – review \& editing: Malika Nisal Ratnayake, Don Chathurika Amarathunga, Asaduz Zaman, Adrian G. Dyer, Alan Dorin.
\end{itemize}

\noindent

\bigskip

\bibliography{sn-bibliography.bib}

\begin{thebibliography}{}
\providecommand{\doi}[1]{\url{https://doi.org/#1}}
\bibcommenthead

\bibitem [\protect \citeauthoryear {%
Abadi%
\ \protect \BOthers {.}}{%
Abadi%
\ \protect \BOthers {.}}{%
{\protect \APACyear {2016}}%
}]{%
abadi2016tensorflow}
\APACinsertmetastar {%
abadi2016tensorflow}%
\begin{APACrefauthors}%
Abadi, M.%
, Barham, P.%
, Chen, J.%
, Chen, Z.%
, Davis, A.%
, Dean, J.%
\BDBL {}Zheng, X.%
\end{APACrefauthors}%
\unskip\
\newblock
\APACrefYearMonthDay{2016}{}{}.
\newblock
{\BBOQ}\APACrefatitle {{TensorFlow: A system for large-scale machine learning}}
  {{TensorFlow: A system for large-scale machine learning}}.{\BBCQ}
\newblock
 \APACrefbtitle {Proceedings of the 12th USENIX Symposium on Operating Systems
  Design and Implementation, OSDI 2016} {Proceedings of the 12th usenix
  symposium on operating systems design and implementation, osdi 2016}\ (\BPGS\
  265---283).
\PrintBackRefs{\CurrentBib}

\bibitem [\protect \citeauthoryear {%
Abdel-Raziq%
, Palmer%
, Koenig%
, Molnar%
\BCBL {}\ \BBA {} Petersen%
}{%
Abdel-Raziq%
\ \protect \BOthers {.}}{%
{\protect \APACyear {2021}}%
}]{%
10.1038/s41598-021-82537-1}
\APACinsertmetastar {%
10.1038/s41598-021-82537-1}%
\begin{APACrefauthors}%
Abdel-Raziq, H.M.%
, Palmer, D.M.%
, Koenig, P.A.%
, Molnar, A.C.%
\BCBL {} Petersen, K.H.%
\end{APACrefauthors}%
\unskip\
\newblock
\APACrefYearMonthDay{2021}{}{}.
\newblock
{\BBOQ}\APACrefatitle {System design for inferring colony-level pollination
  activity through miniature bee-mounted sensors} {System design for inferring
  colony-level pollination activity through miniature bee-mounted
  sensors}.{\BBCQ}
\newblock
\APACjournalVolNumPages{Scientific reports}{11}{1}{1--12}.
\newblock

\newblock

\PrintBackRefs{\CurrentBib}

\bibitem [\protect \citeauthoryear {%
Afonso%
\ \protect \BOthers {.}}{%
Afonso%
\ \protect \BOthers {.}}{%
{\protect \APACyear {2020}}%
}]{%
afonso2020tomato}
\APACinsertmetastar {%
afonso2020tomato}%
\begin{APACrefauthors}%
Afonso, M.%
, Fonteijn, H.%
, Fiorentin, F.S.%
, Lensink, D.%
, Mooij, M.%
, Faber, N.%
\BDBL {}Wehrens, R.%
\end{APACrefauthors}%
\unskip\
\newblock
\APACrefYearMonthDay{2020}{}{}.
\newblock
{\BBOQ}\APACrefatitle {Tomato fruit detection and counting in greenhouses using
  deep learning} {Tomato fruit detection and counting in greenhouses using deep
  learning}.{\BBCQ}
\newblock
\APACjournalVolNumPages{Frontiers in plant science}{11}{}{1759}.
\newblock

\newblock

\PrintBackRefs{\CurrentBib}

\bibitem [\protect \citeauthoryear {%
Aizen%
, Garibaldi%
, Cunningham%
\BCBL {}\ \BBA {} Klein%
}{%
Aizen%
\ \protect \BOthers {.}}{%
{\protect \APACyear {2009}}%
}]{%
aizen2009}
\APACinsertmetastar {%
aizen2009}%
\begin{APACrefauthors}%
Aizen, M.A.%
, Garibaldi, L.A.%
, Cunningham, S.A.%
\BCBL {} Klein, A.M.%
\end{APACrefauthors}%
\unskip\
\newblock
\APACrefYearMonthDay{2009}{}{}.
\newblock
{\BBOQ}\APACrefatitle {How much does agriculture depend on pollinators? Lessons
  from long-term trends in crop production} {How much does agriculture depend
  on pollinators? lessons from long-term trends in crop production}.{\BBCQ}
\newblock
\APACjournalVolNumPages{Annals of botany}{103}{9}{1579--1588}.
\newblock

\newblock

\PrintBackRefs{\CurrentBib}

\bibitem [\protect \citeauthoryear {%
Amarathunga%
, Grundy%
, Parry%
\BCBL {}\ \BBA {} Dorin%
}{%
Amarathunga%
\ \protect \BOthers {.}}{%
{\protect \APACyear {2021}}%
}]{%
amarathunga2021methods}
\APACinsertmetastar {%
amarathunga2021methods}%
\begin{APACrefauthors}%
Amarathunga, D.C.K.%
, Grundy, J.%
, Parry, H.%
\BCBL {} Dorin, A.%
\end{APACrefauthors}%
\unskip\
\newblock
\APACrefYearMonthDay{2021}{}{}.
\newblock
{\BBOQ}\APACrefatitle {Methods of Insect Image Capture and Classification: A
  Systematic Literature Review} {Methods of insect image capture and
  classification: A systematic literature review}.{\BBCQ}
\newblock
\APACjournalVolNumPages{Smart Agricultural Technology}{}{}{100023}.
\newblock

\newblock

\PrintBackRefs{\CurrentBib}

\bibitem [\protect \citeauthoryear {%
Aslanpour%
\ \protect \BOthers {.}}{%
Aslanpour%
\ \protect \BOthers {.}}{%
{\protect \APACyear {2021}}%
}]{%
aslanpour2021serverless}
\APACinsertmetastar {%
aslanpour2021serverless}%
\begin{APACrefauthors}%
Aslanpour, M.S.%
, Toosi, A.N.%
, Cicconetti, C.%
, Javadi, B.%
, Sbarski, P.%
, Taibi, D.%
\BDBL {}Dustdar, S.%
\end{APACrefauthors}%
\unskip\
\newblock
\APACrefYearMonthDay{2021}{}{}.
\newblock
{\BBOQ}\APACrefatitle {Serverless edge computing: vision and challenges}
  {Serverless edge computing: vision and challenges}.{\BBCQ}
\newblock
 \APACrefbtitle {2021 Australasian Computer Science Week Multiconference} {2021
  australasian computer science week multiconference}\ (\BPGS\ 1--10).
\PrintBackRefs{\CurrentBib}

\bibitem [\protect \citeauthoryear {%
Barreiros%
, Dantas%
, Silva%
, Ribeiro%
\BCBL {}\ \BBA {} Barros%
}{%
Barreiros%
\ \protect \BOthers {.}}{%
{\protect \APACyear {2021}}%
}]{%
deoliveirabarreiros2021}
\APACinsertmetastar {%
deoliveirabarreiros2021}%
\begin{APACrefauthors}%
Barreiros, M.d.O.%
, Dantas, D.d.O.%
, Silva, L.C.d.O.%
, Ribeiro, S.%
\BCBL {} Barros, A.K.%
\end{APACrefauthors}%
\unskip\
\newblock
\APACrefYearMonthDay{2021}{}{}.
\newblock
{\BBOQ}\APACrefatitle {Zebrafish tracking using YOLOv2 and Kalman filter}
  {Zebrafish tracking using yolov2 and kalman filter}.{\BBCQ}
\newblock
\APACjournalVolNumPages{Scientific reports}{11}{1}{1--14}.
\newblock

\newblock

\PrintBackRefs{\CurrentBib}

\bibitem [\protect \citeauthoryear {%
Batsleer%
\ \protect \BOthers {.}}{%
Batsleer%
\ \protect \BOthers {.}}{%
{\protect \APACyear {2020}}%
}]{%
batsleer2020}
\APACinsertmetastar {%
batsleer2020}%
\begin{APACrefauthors}%
Batsleer, F.%
, Bonte, D.%
, Dekeukeleire, D.%
, Goossens, S.%
, Poelmans, W.%
, Van~der Cruyssen, E.%
\BDBL {}Vandegehuchte, M.L.%
\end{APACrefauthors}%
\unskip\
\newblock
\APACrefYearMonthDay{2020}{}{}.
\newblock
{\BBOQ}\APACrefatitle {The neglected impact of tracking devices on terrestrial
  arthropods} {The neglected impact of tracking devices on terrestrial
  arthropods}.{\BBCQ}
\newblock
\APACjournalVolNumPages{Methods in Ecology and Evolution}{11}{3}{350--361}.
\newblock

\newblock

\PrintBackRefs{\CurrentBib}

\bibitem [\protect \citeauthoryear {%
Bjerge%
, Mann%
\BCBL {}\ \BBA {} H{\o}ye%
}{%
Bjerge%
, Mann%
\BCBL {}\ \BBA {} H{\o}ye%
}{%
{\protect \APACyear {2021}}%
}]{%
bjerge2021real}
\APACinsertmetastar {%
bjerge2021real}%
\begin{APACrefauthors}%
Bjerge, K.%
, Mann, H.M.%
\BCBL {} H{\o}ye, T.T.%
\end{APACrefauthors}%
\unskip\
\newblock
\APACrefYearMonthDay{2021}{}{}.
\newblock
{\BBOQ}\APACrefatitle {Real-time insect tracking and monitoring with computer
  vision and deep learning} {Real-time insect tracking and monitoring with
  computer vision and deep learning}.{\BBCQ}
\newblock
\APACjournalVolNumPages{Remote Sensing in Ecology and Conservation}{}{}{}.
\newblock

\newblock

\PrintBackRefs{\CurrentBib}

\bibitem [\protect \citeauthoryear {%
Bjerge%
, Nielsen%
, Sepstrup%
, Helsing-Nielsen%
\BCBL {}\ \BBA {} H{\o}ye%
}{%
Bjerge%
, Nielsen%
\BCBL {}\ \protect \BOthers {.}}{%
{\protect \APACyear {2021}}%
}]{%
bjerge2021automated}
\APACinsertmetastar {%
bjerge2021automated}%
\begin{APACrefauthors}%
Bjerge, K.%
, Nielsen, J.B.%
, Sepstrup, M.V.%
, Helsing-Nielsen, F.%
\BCBL {} H{\o}ye, T.T.%
\end{APACrefauthors}%
\unskip\
\newblock
\APACrefYearMonthDay{2021}{}{}.
\newblock
{\BBOQ}\APACrefatitle {An automated light trap to monitor moths (Lepidoptera)
  using computer vision-based tracking and deep learning} {An automated light
  trap to monitor moths (lepidoptera) using computer vision-based tracking and
  deep learning}.{\BBCQ}
\newblock
\APACjournalVolNumPages{Sensors}{21}{2}{343}.
\newblock

\newblock

\PrintBackRefs{\CurrentBib}

\bibitem [\protect \citeauthoryear {%
Bochkovskiy%
, Wang%
\BCBL {}\ \BBA {} Liao%
}{%
Bochkovskiy%
\ \protect \BOthers {.}}{%
{\protect \APACyear {2020}}%
}]{%
bochkovskiy}
\APACinsertmetastar {%
bochkovskiy}%
\begin{APACrefauthors}%
Bochkovskiy, A.%
, Wang, C\BHBI Y.%
\BCBL {} Liao, H\BHBI Y.M.%
\end{APACrefauthors}%
\unskip\
\newblock
\APACrefYearMonthDay{2020}{}{}.
\newblock
{\BBOQ}\APACrefatitle {Yolov4: Optimal speed and accuracy of object detection}
  {Yolov4: Optimal speed and accuracy of object detection}.{\BBCQ}
\newblock
\APACjournalVolNumPages{arXiv preprint arXiv:2004.10934}{}{}{}.
\newblock

\newblock

\PrintBackRefs{\CurrentBib}

\bibitem [\protect \citeauthoryear {%
Branson%
, Robie%
, Bender%
, Perona%
\BCBL {}\ \BBA {} Dickinson%
}{%
Branson%
\ \protect \BOthers {.}}{%
{\protect \APACyear {2009}}%
}]{%
Branson2009}
\APACinsertmetastar {%
Branson2009}%
\begin{APACrefauthors}%
Branson, K.%
, Robie, A.A.%
, Bender, J.%
, Perona, P.%
\BCBL {} Dickinson, M.H.%
\end{APACrefauthors}%
\unskip\
\newblock
\APACrefYearMonthDay{2009}{}{}.
\newblock
{\BBOQ}\APACrefatitle {High-throughput ethomics in large groups of Drosophila}
  {High-throughput ethomics in large groups of drosophila}.{\BBCQ}
\newblock
\APACjournalVolNumPages{Nature methods}{6}{6}{451--457}.
\newblock

\newblock

\PrintBackRefs{\CurrentBib}

\bibitem [\protect \citeauthoryear {%
Breeze%
\ \protect \BOthers {.}}{%
Breeze%
\ \protect \BOthers {.}}{%
{\protect \APACyear {2021}}%
}]{%
breeze2020}
\APACinsertmetastar {%
breeze2020}%
\begin{APACrefauthors}%
Breeze, T.D.%
, Bailey, A.P.%
, Balcombe, K.G.%
, Brereton, T.%
, Comont, R.%
, Edwards, M.%
\BDBL {}others%
\end{APACrefauthors}%
\unskip\
\newblock
\APACrefYearMonthDay{2021}{}{}.
\newblock
{\BBOQ}\APACrefatitle {Pollinator monitoring more than pays for itself}
  {Pollinator monitoring more than pays for itself}.{\BBCQ}
\newblock
\APACjournalVolNumPages{Journal of Applied Ecology}{58}{1}{44--57}.
\newblock

\newblock

\PrintBackRefs{\CurrentBib}

\bibitem [\protect \citeauthoryear {%
Campbell%
, Mummert%
\BCBL {}\ \BBA {} Sukthankar%
}{%
Campbell%
\ \protect \BOthers {.}}{%
{\protect \APACyear {2008}}%
}]{%
campbell2008}
\APACinsertmetastar {%
campbell2008}%
\begin{APACrefauthors}%
Campbell, J.%
, Mummert, L.%
\BCBL {} Sukthankar, R.%
\end{APACrefauthors}%
\unskip\
\newblock
\APACrefYearMonthDay{2008}{}{}.
\newblock
{\BBOQ}\APACrefatitle {{Video monitoring of honey bee colonies at the hive
  entrance}} {{Video monitoring of honey bee colonies at the hive
  entrance}}.{\BBCQ}
\newblock
\APACjournalVolNumPages{Visual observation \& analysis of animal \& insect
  behavior, ICPR}{8}{}{1---4}.
\newblock

\newblock

\PrintBackRefs{\CurrentBib}

\bibitem [\protect \citeauthoryear {%
Chagnon%
, Gingras%
\BCBL {}\ \BBA {} De~Oliveira%
}{%
Chagnon%
\ \protect \BOthers {.}}{%
{\protect \APACyear {1989}}%
}]{%
chagnon1989}
\APACinsertmetastar {%
chagnon1989}%
\begin{APACrefauthors}%
Chagnon, M.%
, Gingras, J.%
\BCBL {} De~Oliveira, D.%
\end{APACrefauthors}%
\unskip\
\newblock
\APACrefYearMonthDay{1989}{}{}.
\newblock
{\BBOQ}\APACrefatitle {Effect of honey bee (Hymenoptera: Apidae) visits on the
  pollination rate of strawberries} {Effect of honey bee (hymenoptera: Apidae)
  visits on the pollination rate of strawberries}.{\BBCQ}
\newblock
\APACjournalVolNumPages{Journal of Economic Entomology}{82}{5}{1350--1353}.
\newblock

\newblock

\PrintBackRefs{\CurrentBib}

\bibitem [\protect \citeauthoryear {%
Dennis%
\ \protect \BOthers {.}}{%
Dennis%
\ \protect \BOthers {.}}{%
{\protect \APACyear {2006}}%
}]{%
dennis2005}
\APACinsertmetastar {%
dennis2005}%
\begin{APACrefauthors}%
Dennis, R.%
, Shreeve, T.%
, Isaac, N.%
, Roy, D.%
, Hardy, P.%
, Fox, R.%
\BCBL {} Asher, J.%
\end{APACrefauthors}%
\unskip\
\newblock
\APACrefYearMonthDay{2006}{}{}.
\newblock
{\BBOQ}\APACrefatitle {The effects of visual apparency on bias in butterfly
  recording and monitoring} {The effects of visual apparency on bias in
  butterfly recording and monitoring}.{\BBCQ}
\newblock
\APACjournalVolNumPages{Biological conservation}{128}{4}{486--492}.
\newblock

\newblock

\PrintBackRefs{\CurrentBib}

\bibitem [\protect \citeauthoryear {%
FAO%
}{%
FAO%
}{%
{\protect \APACyear {2018}}%
}]{%
faobee}
\APACinsertmetastar {%
faobee}%
\begin{APACrefauthors}%
FAO%
\end{APACrefauthors}%
\unskip\
\newblock
\APACrefYearMonthDay{2018}{}{}.
\newblock
{\BBOQ}\APACrefatitle {{Why bees matter; the importance of bees and other
  pollinators for food and agriculture}} {{Why bees matter; the importance of
  bees and other pollinators for food and agriculture}}.{\BBCQ}
\newblock

\newblock

\newblock

\PrintBackRefs{\CurrentBib}

\bibitem [\protect \citeauthoryear {%
Fijen%
\ \protect \BOthers {.}}{%
Fijen%
\ \protect \BOthers {.}}{%
{\protect \APACyear {2018}}%
}]{%
fijen2018insect}
\APACinsertmetastar {%
fijen2018insect}%
\begin{APACrefauthors}%
Fijen, T.P.%
, Scheper, J.A.%
, Boom, T.M.%
, Janssen, N.%
, Raemakers, I.%
\BCBL {} Kleijn, D.%
\end{APACrefauthors}%
\unskip\
\newblock
\APACrefYearMonthDay{2018}{}{}.
\newblock
{\BBOQ}\APACrefatitle {Insect pollination is at least as important for
  marketable crop yield as plant quality in a seed crop} {Insect pollination is
  at least as important for marketable crop yield as plant quality in a seed
  crop}.{\BBCQ}
\newblock
\APACjournalVolNumPages{Ecology letters}{21}{11}{1704--1713}.
\newblock

\newblock

\PrintBackRefs{\CurrentBib}

\bibitem [\protect \citeauthoryear {%
{Food \& Agriculture Organization of the United Nation}%
}{%
{Food \& Agriculture Organization of the United Nation}%
}{%
{\protect \APACyear {2019}}%
}]{%
fao2019}
\APACinsertmetastar {%
fao2019}%
\begin{APACrefauthors}%
{Food \& Agriculture Organization of the United Nation}%
\end{APACrefauthors}%
\unskip\
\newblock
\APACrefYearMonthDay{2019}{}{}.
\newblock
{\BBOQ}\APACrefatitle {Global Action on Pollination Services for Sustainable
  Agriculture} {Global action on pollination services for sustainable
  agriculture}.{\BBCQ}
\newblock

\newblock

\newblock

\PrintBackRefs{\CurrentBib}

\bibitem [\protect \citeauthoryear {%
Garibaldi%
\ \protect \BOthers {.}}{%
Garibaldi%
\ \protect \BOthers {.}}{%
{\protect \APACyear {2016}}%
}]{%
garibaldi2016mutually}
\APACinsertmetastar {%
garibaldi2016mutually}%
\begin{APACrefauthors}%
Garibaldi, L.A.%
, Carvalheiro, L.G.%
, Vaissi{\`e}re, B.E.%
, Gemmill-Herren, B.%
, Hip{\'o}lito, J.%
, Freitas, B.M.%
\BDBL {}others%
\end{APACrefauthors}%
\unskip\
\newblock
\APACrefYearMonthDay{2016}{}{}.
\newblock
{\BBOQ}\APACrefatitle {Mutually beneficial pollinator diversity and crop yield
  outcomes in small and large farms} {Mutually beneficial pollinator diversity
  and crop yield outcomes in small and large farms}.{\BBCQ}
\newblock
\APACjournalVolNumPages{Science}{351}{6271}{388--391}.
\newblock

\newblock

\PrintBackRefs{\CurrentBib}

\bibitem [\protect \citeauthoryear {%
Garibaldi%
, Requier%
, Rollin%
\BCBL {}\ \BBA {} Andersson%
}{%
Garibaldi%
\ \protect \BOthers {.}}{%
{\protect \APACyear {2017}}%
}]{%
garibaldi2017towards}
\APACinsertmetastar {%
garibaldi2017towards}%
\begin{APACrefauthors}%
Garibaldi, L.A.%
, Requier, F.%
, Rollin, O.%
\BCBL {} Andersson, G.K.S.%
\end{APACrefauthors}%
\unskip\
\newblock
\APACrefYearMonthDay{2017}{}{}.
\newblock
{\BBOQ}\APACrefatitle {{Towards an integrated species and habitat management of
  crop pollination}} {{Towards an integrated species and habitat management of
  crop pollination}}.{\BBCQ}
\newblock
\APACjournalVolNumPages{Current opinion in insect science}{21}{}{105---114}.
\newblock

\newblock

\PrintBackRefs{\CurrentBib}

\bibitem [\protect \citeauthoryear {%
Garibaldi%
, S{\'a}ez%
, Aizen%
, Fijen%
\BCBL {}\ \BBA {} Bartomeus%
}{%
Garibaldi%
\ \protect \BOthers {.}}{%
{\protect \APACyear {2020}}%
}]{%
garibaldi2020a}
\APACinsertmetastar {%
garibaldi2020a}%
\begin{APACrefauthors}%
Garibaldi, L.A.%
, S{\'a}ez, A.%
, Aizen, M.A.%
, Fijen, T.%
\BCBL {} Bartomeus, I.%
\end{APACrefauthors}%
\unskip\
\newblock
\APACrefYearMonthDay{2020}{}{}.
\newblock
{\BBOQ}\APACrefatitle {Crop pollination management needs flower-visitor
  monitoring and target values} {Crop pollination management needs
  flower-visitor monitoring and target values}.{\BBCQ}
\newblock
\APACjournalVolNumPages{Journal of Applied Ecology}{57}{4}{664--670}.
\newblock

\newblock

\PrintBackRefs{\CurrentBib}

\bibitem [\protect \citeauthoryear {%
Goscinski%
\ \protect \BOthers {.}}{%
Goscinski%
\ \protect \BOthers {.}}{%
{\protect \APACyear {2014}}%
}]{%
goscinski2014multi}
\APACinsertmetastar {%
goscinski2014multi}%
\begin{APACrefauthors}%
Goscinski, W.J.%
, McIntosh, P.%
, Felzmann, U.C.%
, Maksimenko, A.%
, Hall, C.J.%
, Gureyev, T.%
\BDBL {}others%
\end{APACrefauthors}%
\unskip\
\newblock
\APACrefYearMonthDay{2014}{}{}.
\newblock
{\BBOQ}\APACrefatitle {The multi-modal Australian ScienceS Imaging and
  Visualization Environment (MASSIVE) high performance computing
  infrastructure: applications in neuroscience and neuroinformatics research}
  {The multi-modal australian sciences imaging and visualization environment
  (massive) high performance computing infrastructure: applications in
  neuroscience and neuroinformatics research}.{\BBCQ}
\newblock
\APACjournalVolNumPages{Frontiers in Neuroinformatics}{8}{}{30}.
\newblock

\newblock

\PrintBackRefs{\CurrentBib}

\bibitem [\protect \citeauthoryear {%
Haalck%
, Mangan%
, Webb%
\BCBL {}\ \BBA {} Risse%
}{%
Haalck%
\ \protect \BOthers {.}}{%
{\protect \APACyear {2020}}%
}]{%
Haalck2020}
\APACinsertmetastar {%
Haalck2020}%
\begin{APACrefauthors}%
Haalck, L.%
, Mangan, M.%
, Webb, B.%
\BCBL {} Risse, B.%
\end{APACrefauthors}%
\unskip\
\newblock
\APACrefYearMonthDay{2020}{}{}.
\newblock
{\BBOQ}\APACrefatitle {Towards image-based animal tracking in natural
  environments using a freely moving camera} {Towards image-based animal
  tracking in natural environments using a freely moving camera}.{\BBCQ}
\newblock
\APACjournalVolNumPages{Journal of neuroscience methods}{330}{}{108455}.
\newblock

\newblock

\PrintBackRefs{\CurrentBib}

\bibitem [\protect \citeauthoryear {%
Hall%
, Jones%
, Rocchetti%
, Wright%
\BCBL {}\ \BBA {} Rader%
}{%
Hall%
\ \protect \BOthers {.}}{%
{\protect \APACyear {2020}}%
}]{%
hall2020}
\APACinsertmetastar {%
hall2020}%
\begin{APACrefauthors}%
Hall, M.A.%
, Jones, J.%
, Rocchetti, M.%
, Wright, D.%
\BCBL {} Rader, R.%
\end{APACrefauthors}%
\unskip\
\newblock
\APACrefYearMonthDay{2020}{}{}.
\newblock
{\BBOQ}\APACrefatitle {Bee visitation and fruit quality in berries under
  protected cropping vary along the length of polytunnels} {Bee visitation and
  fruit quality in berries under protected cropping vary along the length of
  polytunnels}.{\BBCQ}
\newblock
\APACjournalVolNumPages{Journal of Economic Entomology}{113}{3}{1337--1346}.
\newblock

\newblock

\PrintBackRefs{\CurrentBib}

\bibitem [\protect \citeauthoryear {%
Hallmann%
\ \protect \BOthers {.}}{%
Hallmann%
\ \protect \BOthers {.}}{%
{\protect \APACyear {2017}}%
}]{%
hallmann2017more}
\APACinsertmetastar {%
hallmann2017more}%
\begin{APACrefauthors}%
Hallmann, C.A.%
, Sorg, M.%
, Jongejans, E.%
, Siepel, H.%
, Hofland, N.%
, Schwan, H.%
\BDBL {}others%
\end{APACrefauthors}%
\unskip\
\newblock
\APACrefYearMonthDay{2017}{}{}.
\newblock
{\BBOQ}\APACrefatitle {More than 75 percent decline over 27 years in total
  flying insect biomass in protected areas} {More than 75 percent decline over
  27 years in total flying insect biomass in protected areas}.{\BBCQ}
\newblock
\APACjournalVolNumPages{PloS one}{12}{10}{e0185809}.
\newblock

\newblock

\PrintBackRefs{\CurrentBib}

\bibitem [\protect \citeauthoryear {%
Howard%
, Nisal~Ratnayake%
, Dyer%
, Garcia%
\BCBL {}\ \BBA {} Dorin%
}{%
Howard%
\ \protect \BOthers {.}}{%
{\protect \APACyear {2021}}%
}]{%
10.1371/journal.pone.0251572}
\APACinsertmetastar {%
10.1371/journal.pone.0251572}%
\begin{APACrefauthors}%
Howard, S.R.%
, Nisal~Ratnayake, M.%
, Dyer, A.G.%
, Garcia, J.E.%
\BCBL {} Dorin, A.%
\end{APACrefauthors}%
\unskip\
\newblock
\APACrefYearMonthDay{2021}{}{}.
\newblock
{\BBOQ}\APACrefatitle {Towards precision apiculture: Traditional and
  technological insect monitoring methods in strawberry and raspberry crop
  polytunnels tell different pollination stories} {Towards precision
  apiculture: Traditional and technological insect monitoring methods in
  strawberry and raspberry crop polytunnels tell different pollination
  stories}.{\BBCQ}
\newblock
\APACjournalVolNumPages{Plos one}{16}{5}{e0251572}.
\newblock

\newblock

\PrintBackRefs{\CurrentBib}

\bibitem [\protect \citeauthoryear {%
H{\o}ye%
\ \protect \BOthers {.}}{%
H{\o}ye%
\ \protect \BOthers {.}}{%
{\protect \APACyear {2021}}%
}]{%
hoye2021}
\APACinsertmetastar {%
hoye2021}%
\begin{APACrefauthors}%
H{\o}ye, T.T.%
, {\"A}rje, J.%
, Bjerge, K.%
, Hansen, O.L.%
, Iosifidis, A.%
, Leese, F.%
\BDBL {}Raitoharju, J.%
\end{APACrefauthors}%
\unskip\
\newblock
\APACrefYearMonthDay{2021}{}{}.
\newblock
{\BBOQ}\APACrefatitle {Deep learning and computer vision will transform
  entomology} {Deep learning and computer vision will transform
  entomology}.{\BBCQ}
\newblock
\APACjournalVolNumPages{Proceedings of the National Academy of
  Sciences}{118}{2}{}.
\newblock

\newblock

\PrintBackRefs{\CurrentBib}

\bibitem [\protect \citeauthoryear {%
Jolles%
}{%
Jolles%
}{%
{\protect \APACyear {2021}}%
}]{%
jolles2021broad}
\APACinsertmetastar {%
jolles2021broad}%
\begin{APACrefauthors}%
Jolles, J.W.%
\end{APACrefauthors}%
\unskip\
\newblock
\APACrefYearMonthDay{2021}{}{}.
\newblock
{\BBOQ}\APACrefatitle {Broad-scale applications of the Raspberry Pi: A review
  and guide for biologists} {Broad-scale applications of the raspberry pi: A
  review and guide for biologists}.{\BBCQ}
\newblock
\APACjournalVolNumPages{Methods in Ecology and Evolution}{12}{9}{1562--1579}.
\newblock

\newblock

\PrintBackRefs{\CurrentBib}

\bibitem [\protect \citeauthoryear {%
Kamilaris%
\ \BBA {} Prenafeta-Bold{\'u}%
}{%
Kamilaris%
\ \BBA {} Prenafeta-Bold{\'u}%
}{%
{\protect \APACyear {2018}}%
}]{%
kamilaris2018deep}
\APACinsertmetastar {%
kamilaris2018deep}%
\begin{APACrefauthors}%
Kamilaris, A.%
\BCBT {}\ \BBA {} Prenafeta-Bold{\'u}, F.X.%
\end{APACrefauthors}%
\unskip\
\newblock
\APACrefYearMonthDay{2018}{}{}.
\newblock
{\BBOQ}\APACrefatitle {Deep learning in agriculture: A survey} {Deep learning
  in agriculture: A survey}.{\BBCQ}
\newblock
\APACjournalVolNumPages{Computers and electronics in
  agriculture}{147}{}{70--90}.
\newblock

\newblock

\PrintBackRefs{\CurrentBib}

\bibitem [\protect \citeauthoryear {%
Kevan%
}{%
Kevan%
}{%
{\protect \APACyear {1975}}%
}]{%
kevan1975sun}
\APACinsertmetastar {%
kevan1975sun}%
\begin{APACrefauthors}%
Kevan, P.G.%
\end{APACrefauthors}%
\unskip\
\newblock
\APACrefYearMonthDay{1975}{}{}.
\newblock
{\BBOQ}\APACrefatitle {Sun-tracking solar furnaces in high arctic flowers:
  significance for pollination and insects} {Sun-tracking solar furnaces in
  high arctic flowers: significance for pollination and insects}.{\BBCQ}
\newblock
\APACjournalVolNumPages{Science}{189}{4204}{723--726}.
\newblock

\newblock

\PrintBackRefs{\CurrentBib}

\bibitem [\protect \citeauthoryear {%
Kirkeby%
\ \protect \BOthers {.}}{%
Kirkeby%
\ \protect \BOthers {.}}{%
{\protect \APACyear {2021}}%
}]{%
kirkeby2021advances}
\APACinsertmetastar {%
kirkeby2021advances}%
\begin{APACrefauthors}%
Kirkeby, C.%
, Rydhmer, K.%
, Cook, S.M.%
, Strand, A.%
, Torrance, M.T.%
, Swain, J.L.%
\BDBL {}others%
\end{APACrefauthors}%
\unskip\
\newblock
\APACrefYearMonthDay{2021}{}{}.
\newblock
{\BBOQ}\APACrefatitle {Advances in automatic identification of flying insects
  using optical sensors and machine learning} {Advances in automatic
  identification of flying insects using optical sensors and machine
  learning}.{\BBCQ}
\newblock
\APACjournalVolNumPages{Scientific reports}{11}{1}{1--8}.
\newblock

\newblock

\PrintBackRefs{\CurrentBib}

\bibitem [\protect \citeauthoryear {%
Koirala%
, Walsh%
, Wang%
\BCBL {}\ \BBA {} McCarthy%
}{%
Koirala%
\ \protect \BOthers {.}}{%
{\protect \APACyear {2019}}%
}]{%
koirala2019deep}
\APACinsertmetastar {%
koirala2019deep}%
\begin{APACrefauthors}%
Koirala, A.%
, Walsh, K.B.%
, Wang, Z.%
\BCBL {} McCarthy, C.%
\end{APACrefauthors}%
\unskip\
\newblock
\APACrefYearMonthDay{2019}{}{}.
\newblock
{\BBOQ}\APACrefatitle {Deep learning--Method overview and review of use for
  fruit detection and yield estimation} {Deep learning--method overview and
  review of use for fruit detection and yield estimation}.{\BBCQ}
\newblock
\APACjournalVolNumPages{Computers and electronics in
  agriculture}{162}{}{219--234}.
\newblock

\newblock

\PrintBackRefs{\CurrentBib}

\bibitem [\protect \citeauthoryear {%
Kuhn%
}{%
Kuhn%
}{%
{\protect \APACyear {1955}}%
}]{%
kuhn1955hungarian}
\APACinsertmetastar {%
kuhn1955hungarian}%
\begin{APACrefauthors}%
Kuhn, H.W.%
\end{APACrefauthors}%
\unskip\
\newblock
\APACrefYearMonthDay{1955}{}{}.
\newblock
{\BBOQ}\APACrefatitle {The Hungarian method for the assignment problem} {The
  hungarian method for the assignment problem}.{\BBCQ}
\newblock
\APACjournalVolNumPages{Naval research logistics quarterly}{2}{1-2}{83--97}.
\newblock

\newblock

\PrintBackRefs{\CurrentBib}

\bibitem [\protect \citeauthoryear {%
Lu%
\ \protect \BOthers {.}}{%
Lu%
\ \protect \BOthers {.}}{%
{\protect \APACyear {2017}}%
}]{%
lu2017cultivated}
\APACinsertmetastar {%
lu2017cultivated}%
\begin{APACrefauthors}%
Lu, H.%
, Fu, X.%
, Liu, C.%
, Li, L\BHBI g.%
, He, Y\BHBI x.%
\BCBL {} Li, N\BHBI w.%
\end{APACrefauthors}%
\unskip\
\newblock
\APACrefYearMonthDay{2017}{}{}.
\newblock
{\BBOQ}\APACrefatitle {Cultivated land information extraction in UAV imagery
  based on deep convolutional neural network and transfer learning} {Cultivated
  land information extraction in uav imagery based on deep convolutional neural
  network and transfer learning}.{\BBCQ}
\newblock
\APACjournalVolNumPages{Journal of Mountain Science}{14}{4}{731--741}.
\newblock

\newblock

\PrintBackRefs{\CurrentBib}

\bibitem [\protect \citeauthoryear {%
MacInnis%
\ \BBA {} Forrest%
}{%
MacInnis%
\ \BBA {} Forrest%
}{%
{\protect \APACyear {2019}}%
}]{%
macinnis2019}
\APACinsertmetastar {%
macinnis2019}%
\begin{APACrefauthors}%
MacInnis, G.%
\BCBT {}\ \BBA {} Forrest, J.R.%
\end{APACrefauthors}%
\unskip\
\newblock
\APACrefYearMonthDay{2019}{}{}.
\newblock
{\BBOQ}\APACrefatitle {Pollination by wild bees yields larger strawberries than
  pollination by honey bees} {Pollination by wild bees yields larger
  strawberries than pollination by honey bees}.{\BBCQ}
\newblock
\APACjournalVolNumPages{Journal of Applied Ecology}{56}{4}{824--832}.
\newblock

\newblock

\PrintBackRefs{\CurrentBib}

\bibitem [\protect \citeauthoryear {%
Magnier%
\ \protect \BOthers {.}}{%
Magnier%
\ \protect \BOthers {.}}{%
{\protect \APACyear {2019}}%
}]{%
magnier2019a}
\APACinsertmetastar {%
magnier2019a}%
\begin{APACrefauthors}%
Magnier, B.%
, Gabbay, E.%
, Bougamale, F.%
, Moradi, B.%
, Pfister, F.%
\BCBL {} Slangen, P.%
\end{APACrefauthors}%
\unskip\
\newblock
\APACrefYearMonthDay{2019}{}{}.
\newblock
{\BBOQ}\APACrefatitle {Multiple honey bees tracking and trajectory modeling}
  {Multiple honey bees tracking and trajectory modeling}.{\BBCQ}
\newblock
 \APACrefbtitle {Multimodal Sensing: Technologies and Applications} {Multimodal
  sensing: Technologies and applications}\ (\BVOL\ 11059, \BPG~110590Z).
\PrintBackRefs{\CurrentBib}

\bibitem [\protect \citeauthoryear {%
Odemer%
}{%
Odemer%
}{%
{\protect \APACyear {2022}}%
}]{%
odemer2022approaches}
\APACinsertmetastar {%
odemer2022approaches}%
\begin{APACrefauthors}%
Odemer, R.%
\end{APACrefauthors}%
\unskip\
\newblock
\APACrefYearMonthDay{2022}{}{}.
\newblock
{\BBOQ}\APACrefatitle {Approaches, challenges and recent advances in automated
  bee counting devices: A review} {Approaches, challenges and recent advances
  in automated bee counting devices: A review}.{\BBCQ}
\newblock
\APACjournalVolNumPages{Annals of Applied Biology}{180}{1}{73--89}.
\newblock

\newblock

\PrintBackRefs{\CurrentBib}

\bibitem [\protect \citeauthoryear {%
O'Grady%
, Langton%
\BCBL {}\ \BBA {} O'Hare%
}{%
O'Grady%
\ \protect \BOthers {.}}{%
{\protect \APACyear {2019}}%
}]{%
o2019edge}
\APACinsertmetastar {%
o2019edge}%
\begin{APACrefauthors}%
O'Grady, M.%
, Langton, D.%
\BCBL {} O'Hare, G.%
\end{APACrefauthors}%
\unskip\
\newblock
\APACrefYearMonthDay{2019}{}{}.
\newblock
{\BBOQ}\APACrefatitle {Edge computing: A tractable model for smart
  agriculture?} {Edge computing: A tractable model for smart
  agriculture?}{\BBCQ}
\newblock
\APACjournalVolNumPages{Artificial Intelligence in Agriculture}{3}{}{42--51}.
\newblock

\newblock

\PrintBackRefs{\CurrentBib}

\bibitem [\protect \citeauthoryear {%
Outhwaite%
, McCann%
\BCBL {}\ \BBA {} Newbold%
}{%
Outhwaite%
\ \protect \BOthers {.}}{%
{\protect \APACyear {2022}}%
}]{%
outhwaite2022agriculture}
\APACinsertmetastar {%
outhwaite2022agriculture}%
\begin{APACrefauthors}%
Outhwaite, C.%
, McCann, P.%
\BCBL {} Newbold, T.%
\end{APACrefauthors}%
\unskip\
\newblock
\APACrefYearMonthDay{2022}{}{}.
\newblock
{\BBOQ}\APACrefatitle {Agriculture and climate change reshape insect
  biodiversity worldwide} {Agriculture and climate change reshape insect
  biodiversity worldwide}.{\BBCQ}
\newblock
\APACjournalVolNumPages{Nature}{}{}{}.
\newblock

\newblock

\PrintBackRefs{\CurrentBib}

\bibitem [\protect \citeauthoryear {%
P{\'e}rez-Escudero%
, Vicente-Page%
, Hinz%
, Arganda%
\BCBL {}\ \BBA {} De~Polavieja%
}{%
P{\'e}rez-Escudero%
\ \protect \BOthers {.}}{%
{\protect \APACyear {2014}}%
}]{%
Perez-Escudero2014}
\APACinsertmetastar {%
Perez-Escudero2014}%
\begin{APACrefauthors}%
P{\'e}rez-Escudero, A.%
, Vicente-Page, J.%
, Hinz, R.C.%
, Arganda, S.%
\BCBL {} De~Polavieja, G.G.%
\end{APACrefauthors}%
\unskip\
\newblock
\APACrefYearMonthDay{2014}{}{}.
\newblock
{\BBOQ}\APACrefatitle {idTracker: tracking individuals in a group by automatic
  identification of unmarked animals} {idtracker: tracking individuals in a
  group by automatic identification of unmarked animals}.{\BBCQ}
\newblock
\APACjournalVolNumPages{Nature methods}{11}{7}{743--748}.
\newblock

\newblock

\PrintBackRefs{\CurrentBib}

\bibitem [\protect \citeauthoryear {%
Potts%
\ \protect \BOthers {.}}{%
Potts%
\ \protect \BOthers {.}}{%
{\protect \APACyear {2016}}%
}]{%
potts2016}
\APACinsertmetastar {%
potts2016}%
\begin{APACrefauthors}%
Potts, S.G.%
, Imperatriz-Fonseca, V.%
, Ngo, H.T.%
, Aizen, M.A.%
, Biesmeijer, J.C.%
, Breeze, T.D.%
\BDBL {}others%
\end{APACrefauthors}%
\unskip\
\newblock
\APACrefYearMonthDay{2016}{}{}.
\newblock
{\BBOQ}\APACrefatitle {Safeguarding pollinators and their values to human
  well-being} {Safeguarding pollinators and their values to human
  well-being}.{\BBCQ}
\newblock
\APACjournalVolNumPages{Nature}{540}{7632}{220--229}.
\newblock

\newblock

\PrintBackRefs{\CurrentBib}

\bibitem [\protect \citeauthoryear {%
Rader%
\ \protect \BOthers {.}}{%
Rader%
\ \protect \BOthers {.}}{%
{\protect \APACyear {2016}}%
}]{%
rader2016}
\APACinsertmetastar {%
rader2016}%
\begin{APACrefauthors}%
Rader, R.%
, Bartomeus, I.%
, Garibaldi, L.A.%
, Garratt, M.P.%
, Howlett, B.G.%
, Winfree, R.%
\BDBL {}others%
\end{APACrefauthors}%
\unskip\
\newblock
\APACrefYearMonthDay{2016}{}{}.
\newblock
{\BBOQ}\APACrefatitle {Non-bee insects are important contributors to global
  crop pollination} {Non-bee insects are important contributors to global crop
  pollination}.{\BBCQ}
\newblock
\APACjournalVolNumPages{Proceedings of the National Academy of
  Sciences}{113}{1}{146--151}.
\newblock

\newblock

\PrintBackRefs{\CurrentBib}

\bibitem [\protect \citeauthoryear {%
Ratnayake%
, Dyer%
\BCBL {}\ \BBA {} Dorin%
}{%
Ratnayake%
\ \protect \BOthers {.}}{%
{\protect \APACyear {2021}}%
{\protect \APACexlab {{\protect \BCnt {1}}}}}]{%
Ratnayake_2021_CVPR}
\APACinsertmetastar {%
Ratnayake_2021_CVPR}%
\begin{APACrefauthors}%
Ratnayake, M.N.%
, Dyer, A.G.%
\BCBL {} Dorin, A.%
\end{APACrefauthors}%
\unskip\
\newblock
\APACrefYearMonthDay{2021{\protect \BCnt {1}}}{}{}.
\newblock
{\BBOQ}\APACrefatitle {Towards Computer Vision and Deep Learning Facilitated
  Pollination Monitoring for Agriculture} {Towards computer vision and deep
  learning facilitated pollination monitoring for agriculture}.{\BBCQ}
\newblock
 \APACrefbtitle {Proceedings of the IEEE/CVF Conference on Computer Vision and
  Pattern Recognition} {Proceedings of the ieee/cvf conference on computer
  vision and pattern recognition}\ (\BPGS\ 2921--2930).
\PrintBackRefs{\CurrentBib}

\bibitem [\protect \citeauthoryear {%
Ratnayake%
, Dyer%
\BCBL {}\ \BBA {} Dorin%
}{%
Ratnayake%
\ \protect \BOthers {.}}{%
{\protect \APACyear {2021}}%
{\protect \APACexlab {{\protect \BCnt {2}}}}}]{%
10.1371/journal.pone.0239504}
\APACinsertmetastar {%
10.1371/journal.pone.0239504}%
\begin{APACrefauthors}%
Ratnayake, M.N.%
, Dyer, A.G.%
\BCBL {} Dorin, A.%
\end{APACrefauthors}%
\unskip\
\newblock
\APACrefYearMonthDay{2021{\protect \BCnt {2}}}{}{}.
\newblock
{\BBOQ}\APACrefatitle {Tracking individual honeybees among wildflower clusters
  with computer vision-facilitated pollinator monitoring} {Tracking individual
  honeybees among wildflower clusters with computer vision-facilitated
  pollinator monitoring}.{\BBCQ}
\newblock
\APACjournalVolNumPages{Plos one}{16}{2}{e0239504}.
\newblock

\newblock

\PrintBackRefs{\CurrentBib}

\bibitem [\protect \citeauthoryear {%
Real%
}{%
Real%
}{%
{\protect \APACyear {2012}}%
}]{%
real2012pollination}
\APACinsertmetastar {%
real2012pollination}%
\begin{APACrefauthors}%
Real, L.%
\end{APACrefauthors}%
\unskip\
\newblock
\APACrefYear{2012}.
\newblock
\APACrefbtitle {Pollination biology} {Pollination biology}.
\newblock
\APACaddressPublisher{}{Elsevier}.
\PrintBackRefs{\CurrentBib}

\bibitem [\protect \citeauthoryear {%
Redmon%
\ \BBA {} Farhadi%
}{%
Redmon%
\ \BBA {} Farhadi%
}{%
{\protect \APACyear {2017}}%
}]{%
redmon2016yolo9000}
\APACinsertmetastar {%
redmon2016yolo9000}%
\begin{APACrefauthors}%
Redmon, J.%
\BCBT {}\ \BBA {} Farhadi, A.%
\end{APACrefauthors}%
\unskip\
\newblock
\APACrefYearMonthDay{2017}{}{}.
\newblock
{\BBOQ}\APACrefatitle {YOLO9000: better, faster, stronger} {Yolo9000: better,
  faster, stronger}.{\BBCQ}
\newblock
 \APACrefbtitle {Proceedings of the IEEE conference on computer vision and
  pattern recognition} {Proceedings of the ieee conference on computer vision
  and pattern recognition}\ (\BPGS\ 7263--7271).
\PrintBackRefs{\CurrentBib}

\bibitem [\protect \citeauthoryear {%
Rollin%
\ \BBA {} Garibaldi%
}{%
Rollin%
\ \BBA {} Garibaldi%
}{%
{\protect \APACyear {2019}}%
}]{%
rollin2019impacts}
\APACinsertmetastar {%
rollin2019impacts}%
\begin{APACrefauthors}%
Rollin, O.%
\BCBT {}\ \BBA {} Garibaldi, L.A.%
\end{APACrefauthors}%
\unskip\
\newblock
\APACrefYearMonthDay{2019}{}{}.
\newblock
{\BBOQ}\APACrefatitle {{Impacts of honeybee density on crop yield: A
  meta-analysis}} {{Impacts of honeybee density on crop yield: A
  meta-analysis}}.{\BBCQ}
\newblock
\APACjournalVolNumPages{Journal of Applied Ecology}{56}{5}{1152---1163}.
\newblock

\newblock

\PrintBackRefs{\CurrentBib}

\bibitem [\protect \citeauthoryear {%
Schweiger%
\ \protect \BOthers {.}}{%
Schweiger%
\ \protect \BOthers {.}}{%
{\protect \APACyear {2010}}%
}]{%
schweiger2010multiple}
\APACinsertmetastar {%
schweiger2010multiple}%
\begin{APACrefauthors}%
Schweiger, O.%
, Biesmeijer, J.C.%
, Bommarco, R.%
, Hickler, T.%
, Hulme, P.E.%
, Klotz, S.%
\BDBL {}others%
\end{APACrefauthors}%
\unskip\
\newblock
\APACrefYearMonthDay{2010}{}{}.
\newblock
{\BBOQ}\APACrefatitle {Multiple stressors on biotic interactions: how climate
  change and alien species interact to affect pollination} {Multiple stressors
  on biotic interactions: how climate change and alien species interact to
  affect pollination}.{\BBCQ}
\newblock
\APACjournalVolNumPages{Biological Reviews}{85}{4}{777--795}.
\newblock

\newblock

\PrintBackRefs{\CurrentBib}

\bibitem [\protect \citeauthoryear {%
Sekachev%
, Manovich%
\BCBL {}\ \BBA {} Zhavoronkov%
}{%
Sekachev%
\ \protect \BOthers {.}}{%
{\protect \APACyear {2019}}%
}]{%
cvat}
\APACinsertmetastar {%
cvat}%
\begin{APACrefauthors}%
Sekachev, B.%
, Manovich, N.%
\BCBL {} Zhavoronkov, A.%
\end{APACrefauthors}%
\unskip\
\newblock
\APACrefYearMonthDay{2019}{}{}.
\newblock
\APACrefbtitle {Computer Vision Annotation Tool.} {Computer vision annotation
  tool.}
\newblock
\APACaddressPublisher{}{Zenodo}.
\newblock
\APACrefnote{GitHub: https://github.com/opencv/cvat}
\newblock
\begin{APACrefDOI} \doi{10.5281/zenodo.3497106} \end{APACrefDOI}
\PrintBackRefs{\CurrentBib}

\bibitem [\protect \citeauthoryear {%
Settele%
, Bishop%
\BCBL {}\ \BBA {} Potts%
}{%
Settele%
\ \protect \BOthers {.}}{%
{\protect \APACyear {2016}}%
}]{%
settele2016climate}
\APACinsertmetastar {%
settele2016climate}%
\begin{APACrefauthors}%
Settele, J.%
, Bishop, J.%
\BCBL {} Potts, S.G.%
\end{APACrefauthors}%
\unskip\
\newblock
\APACrefYearMonthDay{2016}{}{}.
\newblock
{\BBOQ}\APACrefatitle {Climate change impacts on pollination} {Climate change
  impacts on pollination}.{\BBCQ}
\newblock
\APACjournalVolNumPages{Nature Plants}{2}{7}{1--3}.
\newblock

\newblock

\PrintBackRefs{\CurrentBib}

\bibitem [\protect \citeauthoryear {%
Simons%
\ \BBA {} Chabris%
}{%
Simons%
\ \BBA {} Chabris%
}{%
{\protect \APACyear {1999}}%
}]{%
simons1999}
\APACinsertmetastar {%
simons1999}%
\begin{APACrefauthors}%
Simons, D.J.%
\BCBT {}\ \BBA {} Chabris, C.F.%
\end{APACrefauthors}%
\unskip\
\newblock
\APACrefYearMonthDay{1999}{}{}.
\newblock
{\BBOQ}\APACrefatitle {Gorillas in our midst: Sustained inattentional blindness
  for dynamic events} {Gorillas in our midst: Sustained inattentional blindness
  for dynamic events}.{\BBCQ}
\newblock
\APACjournalVolNumPages{perception}{28}{9}{1059--1074}.
\newblock

\newblock

\PrintBackRefs{\CurrentBib}

\bibitem [\protect \citeauthoryear {%
Spaethe%
, Tautz%
\BCBL {}\ \BBA {} Chittka%
}{%
Spaethe%
\ \protect \BOthers {.}}{%
{\protect \APACyear {2001}}%
}]{%
spaethe2001visual}
\APACinsertmetastar {%
spaethe2001visual}%
\begin{APACrefauthors}%
Spaethe, J.%
, Tautz, J.%
\BCBL {} Chittka, L.%
\end{APACrefauthors}%
\unskip\
\newblock
\APACrefYearMonthDay{2001}{}{}.
\newblock
{\BBOQ}\APACrefatitle {Visual constraints in foraging bumblebees: flower size
  and color affect search time and flight behavior} {Visual constraints in
  foraging bumblebees: flower size and color affect search time and flight
  behavior}.{\BBCQ}
\newblock
\APACjournalVolNumPages{Proceedings of the National Academy of
  Sciences}{98}{7}{3898--3903}.
\newblock

\newblock

\PrintBackRefs{\CurrentBib}

\bibitem [\protect \citeauthoryear {%
Spencer%
, Barton%
, Ripple%
\BCBL {}\ \BBA {} Newsome%
}{%
Spencer%
\ \protect \BOthers {.}}{%
{\protect \APACyear {2020}}%
}]{%
10.1016/j.fooweb.2020.e00144}
\APACinsertmetastar {%
10.1016/j.fooweb.2020.e00144}%
\begin{APACrefauthors}%
Spencer, E.E.%
, Barton, P.S.%
, Ripple, W.J.%
\BCBL {} Newsome, T.M.%
\end{APACrefauthors}%
\unskip\
\newblock
\APACrefYearMonthDay{2020}{}{}.
\newblock
{\BBOQ}\APACrefatitle {Invasive European wasps alter scavenging dynamics around
  carrion} {Invasive european wasps alter scavenging dynamics around
  carrion}.{\BBCQ}
\newblock
\APACjournalVolNumPages{Food Webs}{24}{}{e00144}.
\newblock

\newblock

\PrintBackRefs{\CurrentBib}

\bibitem [\protect \citeauthoryear {%
Stojni{\'c}%
\ \protect \BOthers {.}}{%
Stojni{\'c}%
\ \protect \BOthers {.}}{%
{\protect \APACyear {2021}}%
}]{%
stojnic2021method}
\APACinsertmetastar {%
stojnic2021method}%
\begin{APACrefauthors}%
Stojni{\'c}, V.%
, Risojevi{\'c}, V.%
, Mu{\v{s}}tra, M.%
, Jovanovi{\'c}, V.%
, Filipi, J.%
, Kezi{\'c}, N.%
\BCBL {} Babi{\'c}, Z.%
\end{APACrefauthors}%
\unskip\
\newblock
\APACrefYearMonthDay{2021}{}{}.
\newblock
{\BBOQ}\APACrefatitle {A method for detection of small moving objects in UAV
  videos} {A method for detection of small moving objects in uav
  videos}.{\BBCQ}
\newblock
\APACjournalVolNumPages{Remote Sensing}{13}{4}{653}.
\newblock

\newblock

\PrintBackRefs{\CurrentBib}

\bibitem [\protect \citeauthoryear {%
Su%
, Kong%
, Qiao%
\BCBL {}\ \BBA {} Sukkarieh%
}{%
Su%
\ \protect \BOthers {.}}{%
{\protect \APACyear {2021}}%
}]{%
su2021data}
\APACinsertmetastar {%
su2021data}%
\begin{APACrefauthors}%
Su, D.%
, Kong, H.%
, Qiao, Y.%
\BCBL {} Sukkarieh, S.%
\end{APACrefauthors}%
\unskip\
\newblock
\APACrefYearMonthDay{2021}{}{}.
\newblock
{\BBOQ}\APACrefatitle {Data augmentation for deep learning based semantic
  segmentation and crop-weed classification in agricultural robotics} {Data
  augmentation for deep learning based semantic segmentation and crop-weed
  classification in agricultural robotics}.{\BBCQ}
\newblock
\APACjournalVolNumPages{Computers and Electronics in
  Agriculture}{190}{}{106418}.
\newblock

\newblock

\PrintBackRefs{\CurrentBib}

\bibitem [\protect \citeauthoryear {%
Vanbergen%
\ \BBA {} Initiative%
}{%
Vanbergen%
\ \BBA {} Initiative%
}{%
{\protect \APACyear {2013}}%
}]{%
ecosystemadam}
\APACinsertmetastar {%
ecosystemadam}%
\begin{APACrefauthors}%
Vanbergen, A.J.%
\BCBT {}\ \BBA {} Initiative, t.I.P.%
\end{APACrefauthors}%
\unskip\
\newblock
\APACrefYearMonthDay{2013}{}{}.
\newblock
{\BBOQ}\APACrefatitle {Threats to an ecosystem service: pressures on
  pollinators} {Threats to an ecosystem service: pressures on
  pollinators}.{\BBCQ}
\newblock
\APACjournalVolNumPages{Frontiers in Ecology and the
  Environment}{11}{5}{251--259}.
\newblock

\newblock

\PrintBackRefs{\CurrentBib}

\bibitem [\protect \citeauthoryear {%
van~der Kooi%
, Kevan%
\BCBL {}\ \BBA {} Koski%
}{%
van~der Kooi%
\ \protect \BOthers {.}}{%
{\protect \APACyear {2019}}%
}]{%
van2019thermal}
\APACinsertmetastar {%
van2019thermal}%
\begin{APACrefauthors}%
van~der Kooi, C.J.%
, Kevan, P.G.%
\BCBL {} Koski, M.H.%
\end{APACrefauthors}%
\unskip\
\newblock
\APACrefYearMonthDay{2019}{}{}.
\newblock
{\BBOQ}\APACrefatitle {The thermal ecology of flowers} {The thermal ecology of
  flowers}.{\BBCQ}
\newblock
\APACjournalVolNumPages{Annals of Botany}{124}{3}{343--353}.
\newblock

\newblock

\PrintBackRefs{\CurrentBib}

\bibitem [\protect \citeauthoryear {%
Van~Horn%
\ \protect \BOthers {.}}{%
Van~Horn%
\ \protect \BOthers {.}}{%
{\protect \APACyear {2018}}%
}]{%
van2018inaturalist}
\APACinsertmetastar {%
van2018inaturalist}%
\begin{APACrefauthors}%
Van~Horn, G.%
, Mac~Aodha, O.%
, Song, Y.%
, Cui, Y.%
, Sun, C.%
, Shepard, A.%
\BDBL {}Belongie, S.%
\end{APACrefauthors}%
\unskip\
\newblock
\APACrefYearMonthDay{2018}{}{}.
\newblock
{\BBOQ}\APACrefatitle {The inaturalist species classification and detection
  dataset} {The inaturalist species classification and detection
  dataset}.{\BBCQ}
\newblock
 \APACrefbtitle {Proceedings of the IEEE conference on computer vision and
  pattern recognition} {Proceedings of the ieee conference on computer vision
  and pattern recognition}\ (\BPGS\ 8769--8778).
\PrintBackRefs{\CurrentBib}

\bibitem [\protect \citeauthoryear {%
Walter%
\ \BBA {} Couzin%
}{%
Walter%
\ \BBA {} Couzin%
}{%
{\protect \APACyear {2021}}%
}]{%
walter2021trex}
\APACinsertmetastar {%
walter2021trex}%
\begin{APACrefauthors}%
Walter, T.%
\BCBT {}\ \BBA {} Couzin, I.D.%
\end{APACrefauthors}%
\unskip\
\newblock
\APACrefYearMonthDay{2021}{}{}.
\newblock
{\BBOQ}\APACrefatitle {TRex, a fast multi-animal tracking system with
  markerless identification, and 2D estimation of posture and visual fields}
  {Trex, a fast multi-animal tracking system with markerless identification,
  and 2d estimation of posture and visual fields}.{\BBCQ}
\newblock
\APACjournalVolNumPages{eLife}{10}{}{e64000}.
\newblock

\newblock

\PrintBackRefs{\CurrentBib}

\bibitem [\protect \citeauthoryear {%
Wang%
\ \protect \BOthers {.}}{%
Wang%
\ \protect \BOthers {.}}{%
{\protect \APACyear {2016}}%
}]{%
7727770}
\APACinsertmetastar {%
7727770}%
\begin{APACrefauthors}%
Wang, S.%
, Liu, W.%
, Wu, J.%
, Cao, L.%
, Meng, Q.%
\BCBL {} Kennedy, P.J.%
\end{APACrefauthors}%
\unskip\
\newblock
\APACrefYearMonthDay{2016}{}{}.
\newblock
{\BBOQ}\APACrefatitle {Training deep neural networks on imbalanced data sets}
  {Training deep neural networks on imbalanced data sets}.{\BBCQ}
\newblock
 \APACrefbtitle {2016 international joint conference on neural networks
  (IJCNN)} {2016 international joint conference on neural networks (ijcnn)}\
  (\BPGS\ 4368--4374).
\PrintBackRefs{\CurrentBib}

\bibitem [\protect \citeauthoryear {%
Wood%
\ \protect \BOthers {.}}{%
Wood%
\ \protect \BOthers {.}}{%
{\protect \APACyear {2020}}%
}]{%
wood2020managed}
\APACinsertmetastar {%
wood2020managed}%
\begin{APACrefauthors}%
Wood, T.J.%
, Michez, D.%
, Paxton, R.J.%
, Drossart, M.%
, Neumann, P.%
, Gerard, M.%
\BDBL {}others%
\end{APACrefauthors}%
\unskip\
\newblock
\APACrefYearMonthDay{2020}{}{}.
\newblock
{\BBOQ}\APACrefatitle {Managed honey bees as a radar for wild bee decline?}
  {Managed honey bees as a radar for wild bee decline?}{\BBCQ}
\newblock
\APACjournalVolNumPages{Apidologie}{51}{6}{1100--1116}.
\newblock

\newblock

\PrintBackRefs{\CurrentBib}

\bibitem [\protect \citeauthoryear {%
Yang%
, Collins%
\BCBL {}\ \BBA {} Beckerleg%
}{%
Yang%
\ \protect \BOthers {.}}{%
{\protect \APACyear {2018}}%
}]{%
yang2018}
\APACinsertmetastar {%
yang2018}%
\begin{APACrefauthors}%
Yang, C.%
, Collins, J.%
\BCBL {} Beckerleg, M.%
\end{APACrefauthors}%
\unskip\
\newblock
\APACrefYearMonthDay{2018}{}{}.
\newblock
{\BBOQ}\APACrefatitle {A model for pollen measurement using video monitoring of
  honey bees} {A model for pollen measurement using video monitoring of honey
  bees}.{\BBCQ}
\newblock
\APACjournalVolNumPages{Sensing and Imaging}{19}{1}{1--29}.
\newblock

\newblock

\PrintBackRefs{\CurrentBib}

\bibitem [\protect \citeauthoryear {%
Zivkovic%
\ \BBA {} Van Der~Heijden%
}{%
Zivkovic%
\ \BBA {} Van Der~Heijden%
}{%
{\protect \APACyear {2006}}%
}]{%
zivkovic2006efficient}
\APACinsertmetastar {%
zivkovic2006efficient}%
\begin{APACrefauthors}%
Zivkovic, Z.%
\BCBT {}\ \BBA {} Van Der~Heijden, F.%
\end{APACrefauthors}%
\unskip\
\newblock
\APACrefYearMonthDay{2006}{}{}.
\newblock
{\BBOQ}\APACrefatitle {Efficient adaptive density estimation per image pixel
  for the task of background subtraction} {Efficient adaptive density
  estimation per image pixel for the task of background subtraction}.{\BBCQ}
\newblock
\APACjournalVolNumPages{Pattern recognition letters}{27}{7}{773--780}.
\newblock

\newblock

\PrintBackRefs{\CurrentBib}

\end{thebibliography}


\end{document}


\maketitle
\tableofcontents 

\clearpage

\addcontentsline{toc}{section}{Supplementary Table: Details of Test Video Dataset}
\begin{landscape}

\begin{table}[H]
\centering
\caption*{\textbf{Supplementary Table: Details of Test Video Dataset} ``Camera Number'' shows the number of the Raspberry Pi camera unit that was used to record the date. ``Record Date'' is the date video was captured. ``Number of frames'' gives the total frames in the video, and ``Frames with insects'' presents the total number of frames with an insect visible in each video. The number of insects visible for more than 5 frames is shown in the insect column. ``No. of Flowers'' shows the total number of fully open visible flowers in a test video.}
\small
\label{table:chp6_test_video_details}
\renewcommand{\arraystretch}{1.2}
\begin{tabular}{|c|c|c|c|c|c|c|c|c|c|} 
\hline
\multirow{2}{*}{\begin{tabular}[c]{@{}c@{}}\textbf{Test }\\\textbf{Video}\end{tabular}} & \multirow{2}{*}{\begin{tabular}[c]{@{}c@{}}\textbf{Camera}\\\textbf{Number}\end{tabular}} & \multirow{2}{*}{\begin{tabular}[c]{@{}c@{}}\textbf{Record}\\\textbf{Date}\end{tabular}} & \multirow{2}{*}{\begin{tabular}[c]{@{}c@{}}\textbf{\textbf{No. of}}\\\textbf{\textbf{Frames}}\end{tabular}} & \multirow{2}{*}{\begin{tabular}[c]{@{}c@{}}\textbf{No. of~frames}\\\textbf{with Insects}\end{tabular}} & \multicolumn{4}{c|}{\textbf{Number of Insects Observed}} & \multirow{2}{*}{\begin{tabular}[c]{@{}c@{}}\textbf{No. of }\\\textbf{ Flowers}\end{tabular}} \\ 
\cline{6-9}
 &  &  &  &  & \textbf{Honeybees} & \textbf{Syrphidae} & \textbf{Lepidoptera} & \textbf{Vespidae} &  \\ 
\hline
T1 & 4 & 9/03/2021 & 18002 & 3181 & 1 & 2 & - & - & 8 \\
T2 & 8 & 10/03/2021 & 18003 & 1352 & 1 & - & - & - & 9 \\
T3 & 7 & 10/03/2021 & 18004 & 5580 & 1 & - & 1 & 1 & 6 \\
T4 & 3 & 17/03/2021 & 17959 & 2044 & 7 & - & - & 1 & 7 \\
T5 & 1 & 17/03/2021 & 18012 & 645 & 2 & - & - & 1 & 7 \\
T6 & 2 & 8/03/2021 & 18004 & 532 & - & - & - & 5 & 5 \\
T7 & 3 & 12/03/2021 & 18000 & 5319 & 3 & - & 1 & - & 7 \\
T8 & 1 & 15/03/2021 & 17967 & 2341 & 5 & - & 1 & 2 & 6 \\
T9 & 4 & 17/03/2021 & 18004 & 623 & - & 1 & 1 & - & 9 \\
T10 & 4 & 17/03/2021 & 17951 & 2412 & - & 3 & - & 8 & 8 \\ 
\hline
\textbf{Total} & \textbf{-} & - & \textbf{179906} & \textbf{24029} & \textbf{20} & \textbf{6} & \textbf{4} & \textbf{10} & \textbf{72} \\
\hline
\end{tabular}
\end{table}
\end{landscape}

\section*{Supplementary Table: Fields of View (FoV) of the Test Video Dataset}
\addcontentsline{toc}{section}{Supplementary Table: Fields of View of the Test Video Dataset}

\begin{table}[H]
\caption*{\textbf{Supplementary Table: Fields of View (FoV) of the Test Video Dataset}}
\label{table:c3_test_fov1}
\centering
\graphicspath{{locations/}}
\begin{tabular}{|c|c|}
\hline
 T1 & T2 \\ \hline
 \includegraphics[width= 0.49\linewidth]{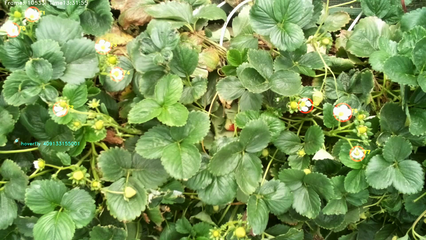}  & \includegraphics[width=0.49\linewidth]{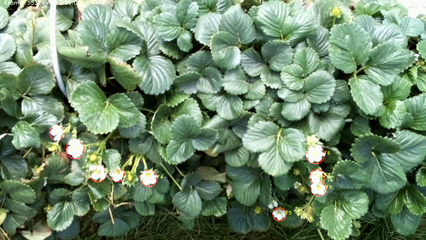} \\ \hline \hline
 T3 & T4 \\ \hline
\includegraphics[width= 0.49\linewidth]{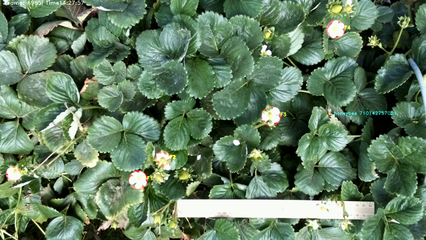}  & \includegraphics[width=0.49\linewidth]{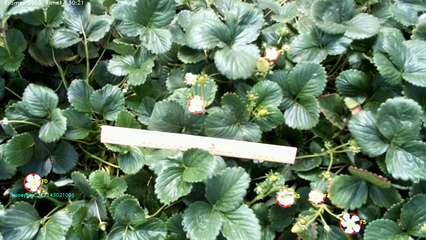} \\ \hline \hline
T5 & T6 \\ \hline
 \includegraphics[width= 0.49\linewidth]{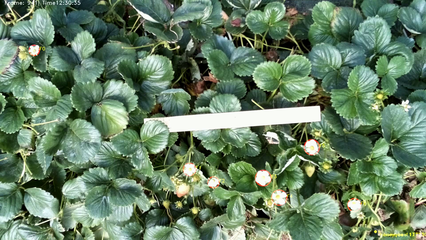}  & \includegraphics[width=0.49\linewidth]{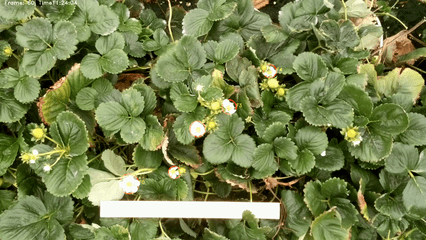} \\ \hline \hline
\end{tabular}
\end{table}

\begin{table}[H]
\caption*{\textbf{Supplementary Table: Fields of Views (FoV) of the Test Video Dataset}}
\label{table:c3_test_fov2}
\centering
\graphicspath{{locations/}}
\begin{tabular}{|c|c|}
\hline
T7 & T8 \\ \hline
 \includegraphics[width= 0.49\linewidth]{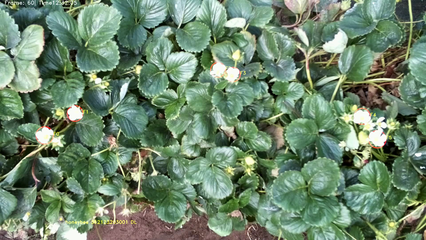}  & \includegraphics[width=0.49\linewidth]{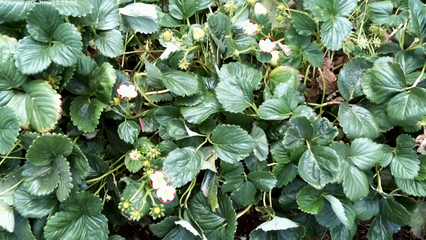} \\ \hline \hline
 T9 & T10 \\ \hline
 \includegraphics[width= 0.49\linewidth]{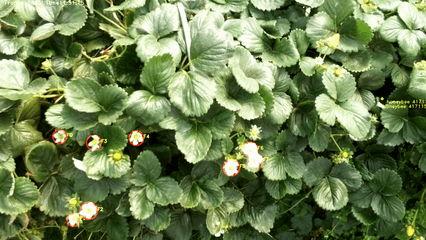}  & \includegraphics[width=0.49\linewidth]{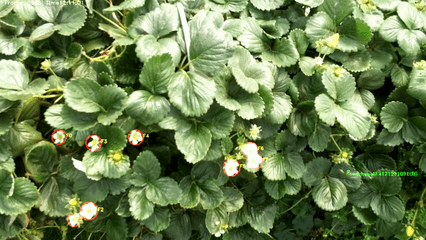} \\ \hline 
\end{tabular}
\end{table}

\section*{Supplementary Figure: Distribution of the Number of Image Region Changes in Test Videos}
\addcontentsline{toc}{section}{Supplementary Figure: Distribution of Image Region Changes in Test Videos}

\begin{figure}[H]
\begin{center}
   \includegraphics[width=\linewidth]{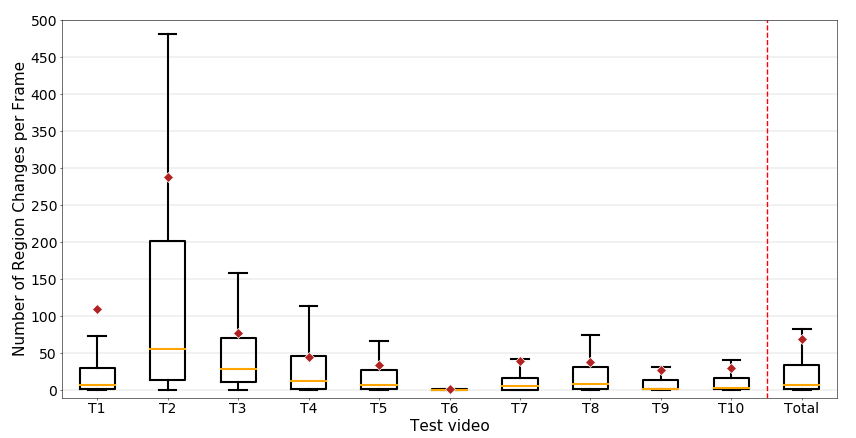}
\end{center}
   \caption*{\textbf{Supplementary Figure: Distribution of the Number of Image Region Changes in Test Videos.} The number of image regions is the number of non-intersecting regions of area greater than that of an insect. The red diamond indicates the mean value and the orange line shows the median.}
\label{fig:c6_test_video_bgchange_ex2}
\end{figure}

\clearpage

\section*{Supplementary Table: YOLOv4 Test and Training Dataset Information}
\addcontentsline{toc}{section}{Supplementary Table: YOLOv4 Test and Training Dataset Information}

\begin{table}[H]
\centering
\caption*{\textbf{Supplementary Table: YOLOv4 Training and Test Dataset Information}. The YOLOv4 dataset was trained for four classes (3 insect classes and 1 flower class). Honeybees and Vespidae were included in a single Hymenopteran class due to their physical similarities and the difficulty of automatically distinguishing between them using the low-quality video footage extracted from the Raspberry Pi cameras.}
\label{table:chp3_datset}
\renewcommand{\arraystretch}{1.2}
\resizebox{\textwidth}{!}{
\begin{tabular}{|c|c|c|c|c|c|} 
\hline
\multirow{2}{*}{\textbf{Dataset}} & \multirow{2}{*}{\textbf{Images}} & \multicolumn{4}{c|}{\textbf{Number of instances in a class}} \\ 
\cline{3-6}
 &  & \begin{tabular}[c]{@{}c@{}}\textbf{Honeybee /}\\\textbf{Vespidae (wasp)}\end{tabular} & \begin{tabular}[c]{@{}c@{}}\textbf{~Syrphidae}\\\textbf{(hoverfly)}\end{tabular} & \begin{tabular}[c]{@{}c@{}}\textbf{~Lepidoptera }\\\textbf{(moth and butterfly)}\end{tabular} & \begin{tabular}[c]{@{}c@{}}\textbf{Strawberry}\\\textbf{~Flower}\end{tabular} \\ 
\hline
Training & 3073 & 2331/371 & 204 & 93 & 14050 \\
Testing & 337 & 258/44 & 20 & 15 & 2909 \\ 
\hline
\textbf{Total} & \textbf{3410} & \textbf{2589/415} & \textbf{224} & \textbf{108} & \textbf{16959} \\
\hline
\end{tabular}}
\end{table}

\section*{Supplementary Data: YOLOv4 Evaluation Results}
\addcontentsline{toc}{section}{Supplementary Data: YOLOv4 Evaluation Results}

\begin{verbatim}
Training_epoches = 7000
detections_count = 5590, unique_truth_count = 3246
class id = 0, name = Honeybee/Vespidae, ap = 65.79%  (TP - 189,   FP = 147)
class id = 1, name = flower, ap = 93.67% (TP = 2558, FP = 163)
class_id = 2, name = hoverfly, ap = 0.15% (TP = 0, FP = 2)
class_id = 3, name = moth, ap = 85.42% (TP - 8, FP = 0)


for conf_thresh = 0.25, precision = 0.90, recall - 0.85, F1-score = 0.87
for conf_thresh = 0.25, TP = 2755, FP - 312, FN = 491, average IOU = 75.50%

IOU threshold = 50%, used Area-Under-Curve for each unique Recall
mean Average Precision (mAP@0.50) = 0.612543, or 61.25%
Total Detection Time: 28 Seconds

\end{verbatim}

\clearpage

\addcontentsline{toc}{section}{Supplementary Table: Insect Trajectory and Flower Visit Detection Evaluation} 

\begin{landscape}
\begin{table}
\centering
\caption*{\textbf{Supplementary Table: Results of the Insect Trajectory and Flower Visit Detection Evaluation for the Test Video Dataset.} ``Trajectory Code'' shows the identification code of the extracted track. ``Manual Observation'' shows the insect track data obtained through manual human observation of test videos. ``Entry frame'' is the frame the insect first entered the video frame. ``Exit frame'' is the last frame the insect is visible inside the frame. If an insect departed the frame and later reappeared, or if it flew under the foliage and later reappeared, it was considered a ``new'' insect. ``Occluded Frames'' shows the number of frames the insect was partially occluded. ``Visible Frames'' shows the number of frames the insect was fully visible. ``Confusion Metrics'' compares trajectories extracted by our algorithm against manual observations. TP = True Positive, FP = False Positive, and FN = False Negative. A detection was considered a True Positive if the algorithm recorded the position of an insect in an area that was in fact covered by the body of the insect.  ``Detection Evaluation Metrics'' presents the Precision, Recall and F-Score metrics for each track based on the confusion metrics. Metrics were only calculated for extracted trajectories. ``Flower Visit Data'' compares the insect flower visits detected by the system against manual human observations.}
\resizebox{\linewidth}{!}{
\renewcommand{\arraystretch}{1.2}
\centering
\begin{tabular}{|c||c||c||c|c|c|c||c|c|c||c|c|c||c|c|c|c|} 
\hline
\multirow{2}{*}{\begin{tabular}[c]{@{}c@{}}\textbf{Test}\\\textbf{Video}\end{tabular}} & \multirow{2}{*}{\begin{tabular}[c]{@{}c@{}}\textbf{Insect}\\\textbf{Type}\end{tabular}} & \multirow{2}{*}{\begin{tabular}[c]{@{}c@{}}\textbf{Trajectory}\\\textbf{Code}\end{tabular}} & \multicolumn{4}{c||}{\textbf{Manual Observations}} & \multicolumn{3}{c||}{\textbf{Confusion Metrics}} & \multicolumn{3}{c||}{\textbf{Detection Evaluation Metrics}} & \multicolumn{4}{c|}{\textbf{Flower Visit Detection}} \\ 
\cline{4-17}
 &  &  & \begin{tabular}[c]{@{}c@{}}\textbf{Entry}\\\textbf{Frame}\end{tabular} & \begin{tabular}[c]{@{}c@{}}\textbf{Exit}\\\textbf{Frame~}\end{tabular} & \begin{tabular}[c]{@{}c@{}}\textbf{Occluded}\\\textbf{Frames}\end{tabular} & \begin{tabular}[c]{@{}c@{}}\textbf{Visible}\\\textbf{Frames}\end{tabular} & \textbf{TP} & \textbf{FP} & \textbf{FN} & \textbf{Precision} & \textbf{Recall} & \textbf{F-Score} & \textbf{Observed} & \textbf{TP} & \textbf{FP} & \textbf{FN} \\ 
\hline\hline
\multirow{5}{*}{T1} & Syrphidae & 40913315501 & 10529 & 10931 & 0 & 402 & 336 & 0 & 66 & 1.00 & 0.84 & 0.91 & 1 & 1 & 0 & 0 \\ 
\cline{2-17}
 & Syrphidae & 40913321301 & 11072 & 11119 & 0 & 47 & 37 & 0 & 10 & 1.00 & 0.79 & 0.88 & 0 & 0 & 1 & 0 \\ 
\cline{2-17}
 & \multirow{3}{*}{Honeybee} & 40913341400 & \multirow{3}{*}{14722} & \multirow{3}{*}{17493} & \multirow{3}{*}{39} & \multirow{3}{*}{2732} & \multirow{3}{*}{2566} & \multirow{3}{*}{116} & \multirow{3}{*}{50} & \multirow{3}{*}{0.96} & \multirow{3}{*}{0.98} & \multirow{3}{*}{0.97} & \multirow{3}{*}{10} & \multirow{3}{*}{10} & \multirow{3}{*}{0} & \multirow{3}{*}{0} \\ 
\cline{3-3}
 &  & 40913342400 &  &  &  &  &  &  &  &  &  &  &  &  &  &  \\ 
\cline{3-3}
 &  & 40913345500 &  &  &  &  &  &  &  &  &  &  &  &  &  &  \\ 
\hline\hline
T2 & Honeybee & 81013473200 & 15861 & 17213 & 0 & 1352 & 1347 & 3 & 2 & 1.00 & 1.00 & 1.00 & 7 & 7 & 0 & 0 \\ 
\hline\hline
\multirow{3}{*}{T3} & Lepidoptera & Not Tracked & 6352 & 6367 & 0 & 15 & \multicolumn{6}{c||}{Trajectory Not Available} & 0 & 0 & 0 & 0 \\ 
\cline{2-17}
 & Vespidae & 71014275703 & 6958 & 6968 & 0 & 10 & 5 & 0 & 5 & 1.00 & 0.50 & 0.67 & 0 & 0 & 0 & 0 \\ 
\cline{2-17}
 & Honeybee & 71014294500 & 9918 & 15473 & 0 & 5555 & 5243 & 7 & 305 & 1.00 & 0.95 & 0.97 & 11 & 9 & 0 & 2 \\ 
\hline\hline
\multirow{9}{*}{T4} & Honeybee & 31714302100 & 5902 & 6126 & 0 & 224 & 212 & 0 & 12 & 1.00 & 0.95 & 0.97 & 1 & 1 & 0 & 0 \\ 
\cline{2-17}
 & Honeybee & 31714304100 & 6505 & 6585 & 0 & 80 & 78 & 0 & 2 & 1.00 & 0.98 & 0.99 & 0 & 0 & 0 & 0 \\ 
\cline{2-17}
 & Honeybee & 31714305200 & 6831 & 7142 & 0 & 311 & 290 & 15 & 6 & 0.95 & 0.98 & 0.97 & 2 & 2 & 0 & 0 \\ 
\cline{2-17}
 & Honeybee & 31714310600 & 7232 & 7586 & 0 & 354 & 354 & 0 & 0 & 1.00 & 1.00 & 1.00 & 2 & 2 & 0 & 0 \\ 
\cline{2-17}
 & Honeybee & 31714311900 & 7596 & 7679 & 0 & 83 & 57 & 0 & 26 & 1.00 & 0.69 & 0.81 & 1 & 1 & 0 & 0 \\ 
\cline{2-17}
 & Vespidae & 31714325703 & 10569 & 10598 & 0 & 29 & 22 & 0 & 7 & 1.00 & 0.76 & 0.86 & 0 & 0 & 0 & 0 \\ 
\cline{2-17}
 & Honeybee & 31714330200 & 10733 & 10749 & 0 & 16 & 11 & 0 & 5 & 1.00 & 0.69 & 0.81 & 0 & 0 & 0 & 0 \\ 
\cline{2-17}
 & \multirow{2}{*}{Honeybee} & 31714331200 & \multirow{2}{*}{11014} & \multirow{2}{*}{11961} & \multirow{2}{*}{0} & \multirow{2}{*}{947} & \multirow{2}{*}{731} & \multirow{2}{*}{58} & \multirow{2}{*}{158} & \multirow{2}{*}{0.93} & \multirow{2}{*}{0.82} & \multirow{2}{*}{0.87} & \multirow{2}{*}{7} & \multirow{2}{*}{7} & \multirow{2}{*}{0} & \multirow{2}{*}{0} \\ 
\cline{3-3}
 &  & 31714331800 &  &  &  &  &  &  &  &  &  &  &  &  &  &  \\ 
\hline\hline
\multirow{3}{*}{T5} & Honeybee & 11712303500 & 925 & 950 & 0 & 25 & 12 & 0 & 13 & 1 & 0.48 & 0.65 & 0 & 0 & 0 & 0 \\ 
\cline{2-17}
 & Vespidae & 11712383603 & 15368 & 15373 & 0 & 5 & 3 & 0 & 2 & 1 & 0.60 & 0.75 & 0 & 0 & 0 & 0 \\ 
\cline{2-17}
 & Honeybee & 11712394400 & 17396 & 18011 & 0 & 615 & 583 & 0 & 32 & 1 & 0.95 & 0.97 & 2 & 2 & 0 & 0 \\ 
\hline
\end{tabular}
}
\end{table}

\begin{table}
\centering
\caption*{\textbf{Supplementary Table: Results of the Insect Trajectory and Flower Visit Detection Evaluation for the Test Video Dataset.} ``Trajectory Code'' shows the identification code of the extracted track. ``Manual Observation'' shows the insect track data obtained through manual human observation of test videos. ``Entry frame'' is the frame the insect first entered the video frame. ``Exit frame'' is the last frame insect visible inside the frame. If an insect departed the frame and later reappeared, or if it flew under the foliage and later reappeared, it was considered a ``new'' insect. ``Occluded Frames'' shows the number of frames insect was partially occluded. ``Visible Frames'' shows the number of frames the insect is fully visible. ``Confusion Metrics'' compares trajectories extracted by our algorithm against manual observations. TP = True Positive, FP = False Positive, and FN = False Negative. A detection was considered a True Positive if the algorithm recorded the position of an insect in an area that was in fact covered by the body of the insect.  ``Detection Evaluation Metrics'' calculates the Precision, Recall and F-Score metrics for each track based on the confusion metrics. Metrics were only calculated for extracted trajectories. ``Flower Visit Detection'' compares the insect flower visits detected by the system against manual human observations.}
\resizebox{\linewidth}{!}{
\renewcommand{\arraystretch}{1.2}
\centering
\begin{tabular}{|c||c||c||c|c|c|c||c|c|c||c|c|c||c|c|c|c|} 
\hline
\multirow{2}{*}{\begin{tabular}[c]{@{}c@{}}\textbf{Test}\\\textbf{Video}\end{tabular}} & \multirow{2}{*}{\begin{tabular}[c]{@{}c@{}}\textbf{Insect}\\\textbf{Type}\end{tabular}} & \multirow{2}{*}{\begin{tabular}[c]{@{}c@{}}\textbf{Trajectory}\\\textbf{Code}\end{tabular}} & \multicolumn{4}{c||}{\textbf{Manual Observations}} & \multicolumn{3}{c||}{\textbf{Confusion Metrics}} & \multicolumn{3}{c||}{\textbf{Detection Evaluation Metrics}} & \multicolumn{4}{c|}{\textbf{Flower Visit Detection}} \\ 
\cline{4-17}
 &  &  & \begin{tabular}[c]{@{}c@{}}\textbf{Entry}\\\textbf{Frame}\end{tabular} & \begin{tabular}[c]{@{}c@{}}\textbf{Exit}\\\textbf{Frame~}\end{tabular} & \begin{tabular}[c]{@{}c@{}}\textbf{Occluded}\\\textbf{Frames}\end{tabular} & \begin{tabular}[c]{@{}c@{}}\textbf{Visible}\\\textbf{Frames}\end{tabular} & \textbf{TP} & \textbf{FP} & \textbf{FN} & \textbf{Precision} & \textbf{Recall} & \textbf{F-Score} & \textbf{Observed} & \textbf{TP} & \textbf{FP} & \textbf{FN} \\ 
\hline\hline

\multirow{5}{*}{T6} & Vespidae & 20811265803 & 5296 & 5420 & 6 & 118 & 114 & 1 & 3 & 0.99 & 0.97 & 0.98 & 0 & 0 & 0 & 0 \\ 
\cline{2-17}
 & Vespidae & 20811270403 & 5474 & 5563 & 0 & 89 & 79 & 0 & 10 & 1.00 & 0.89 & 0.94 & 0 & 0 & 0 & 0 \\ 
\cline{2-17}
 & Vespidae & 20811271003 & 5648 & 5670 & 0 & 22 & 17 & 0 & 5 & 1.00 & 0.77 & 0.87 & 0 & 0 & 0 & 0 \\ 
\cline{2-17}
 & Vespidae & 20811271203 & 5686 & 5768 & 0 & 82 & 59 & 0 & 23 & 1.00 & 0.72 & 0.84 & 0 & 0 & 0 & 0 \\ 
\cline{2-17}
 & Vespidae & 20811273303 & 6244 & 6545 & 80 & 221 & 163 & 0 & 58 & 1.00 & 0.74 & 0.85 & 0 & 0 & 0 & 0 \\ 
\hline\hline
\multirow{6}{*}{T7} & \multirow{3}{*}{Lepidoptera} & 31212363302 & \multirow{3}{*}{8113} & \multirow{3}{*}{11037} & \multirow{3}{*}{0} & \multirow{3}{*}{2924} & \multirow{3}{*}{1395} & \multirow{3}{*}{0} & \multirow{3}{*}{1529} & \multirow{3}{*}{1} & \multirow{3}{*}{0.48} & \multirow{3}{*}{0.65} & \multirow{3}{*}{5} & \multirow{3}{*}{5} & \multirow{3}{*}{0} & \multirow{3}{*}{0} \\ 
\cline{3-3}
 &  & 31212375502 &  &  &  &  &  &  &  &  &  &  &  &  &  &  \\ 
\cline{3-3}
 &  & 31212381002 &  &  &  &  &  &  &  &  &  &  &  &  &  &  \\ 
\cline{2-17}
 & Honeybee & 31212390300 & 12608 & 14210 & 0 & 1602 & 1588 & 1 & 13 & 1.00 & 0.99 & 1.00 & 11 & 11 & 0 & 0 \\ 
\cline{2-17}
 & Honeybee & 31212402700 & 15129 & 15587 & 0 & 458 & 458 & 0 & 0 & 1.00 & 1.00 & 1.00 & 2 & 2 & 0 & 0 \\ 
\cline{2-17}
 & Honeybee & 31212413000 & 17013 & 17348 & 0 & 335 & 335 & 0 & 0 & 1.00 & 1.00 & 1.00 & 3 & 3 & 0 & 0 \\ 
\hline\hline
\multirow{8}{*}{T8} & Honeybee & 11513300100 & 3517 & 4116 & 0 & 599 & 599 & 0 & 0 & 1.00 & 1.00 & 1.00 & 2 & 2 & 0 & 0 \\ 
\cline{2-17}
 & Honeybee & 11513303900 & 4649 & 5213 & 0 & 564 & 562 & 2 &  & 1.00 & 1.00 & 1.00 & 2 & 2 & 0 & 0 \\ 
\cline{2-17}
 & Honeybee & 11513305900 & 5258 & 5851 & 0 & 593 & 588 & 1 & 4 & 1.00 & 0.99 & 1.00 & 1 & 1 & 0 & 0 \\ 
\cline{2-17}
 & Honeybee & 11513313000 & 6200 & 6238 & 0 & 38 & 37 & 1 & 0 & 0.97 & 1.00 & 0.99 & 1 & 1 & 0 & 0 \\ 
\cline{2-17}
 & Vespidae & 11513325003 & 8590 & 8596 & 0 & 6 & 2 & 0 & 4 & 1.00 & 0.33 & 0.50 & 0 & 0 &  &  \\ 
\cline{2-17}
 & Lepidoptera & 11513335102 & 10434 & 10605 & 0 & 171 & 163 & 4 & 4 & 0.98 & 0.98 & 0.98 & 1 & 1 & 1 & 0 \\ 
\cline{2-17}
 & Honeybee & 11513365400 & 15919 & 16282 & 0 & 363 & 356 & 0 & 7 & 1.00 & 0.98 & 0.99 & 2 & 2 & 0 & 0 \\ 
\cline{2-17}
 & Vespidae & 11513375903 & 17877 & 17884 & 0 & 7 & 7 & 0 & 0 & 1.00 & 1.00 & 1.00 & 0 & 0 & 0 & 0 \\ 
\hline\hline
\multirow{2}{*}{T9} & Lepidoptera & 41711511502 & 5795 & 5846 & 0 & 48 & 33 & 0 & 15 & 1.00 & 0.69 & 0.81 & 0 & 0 & 0 & 0 \\ 
\cline{2-17}
 & Syrphidae & Not Tracked & 17374 & 17949 & 0 & 575 & \multicolumn{6}{c||}{Tracking Data Not Available} & 0 & 0 & 0 & 0 \\ 
\hline\hline
\multirow{4}{*}{T10} & \multirow{2}{*}{Syrphidae} & 41712110101 & \multirow{2}{*}{5350} & \multirow{2}{*}{7257} & \multirow{2}{*}{0} & \multirow{2}{*}{1907} & \multirow{2}{*}{584} & \multirow{2}{*}{0} & \multirow{2}{*}{1323} & \multirow{2}{*}{1.00} & \multirow{2}{*}{0.31} & \multirow{2}{*}{0.47} & \multirow{2}{*}{2} & \multirow{2}{*}{1} & \multirow{2}{*}{0} & \multirow{2}{*}{1} \\ 
\cline{3-3}
 &  & 41712111901 &  &  &  &  &  &  &  &  &  &  &  &  &  &  \\ 
\cline{2-17}
 & Syrphidae & 41712164401 & 15593 & 15864 & 0 & 271 & 176 & 2 & 93 & 0.99 & 0.65 & 0.79 & 1 & 1 & 0 & 0 \\ 
\cline{2-17}
 & Syrphidae & 41712171901 & 16716 & 16950 & 0 & 234 & 230 & 0 & 4 & 1.00 & 0.98 & 0.99 & 1 & 1 & 0 & 0 \\
\hline
\end{tabular}
}
\end{table}
\clearpage
\end{landscape}

\section*{Supplementary Figures: Insect Trajectories Extracted from Test Dataset Videos}
\addcontentsline{toc}{section}{Supplementary Figure: Insect Trajectories Extracted from Test Dataset Videos}

\begin{figure}[H]
\begin{center}
   \includegraphics[width=\linewidth]{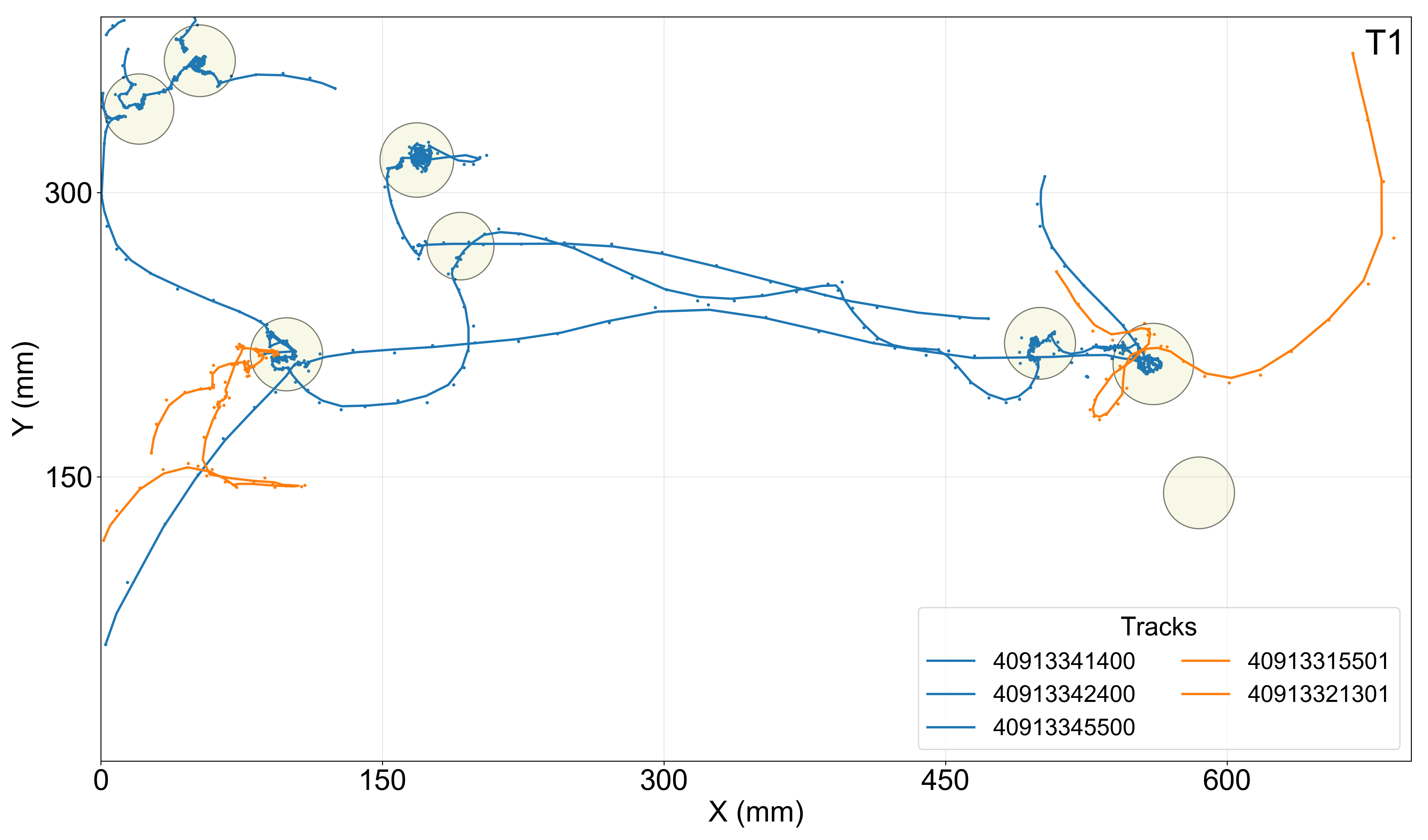}
   \includegraphics[width=\linewidth]{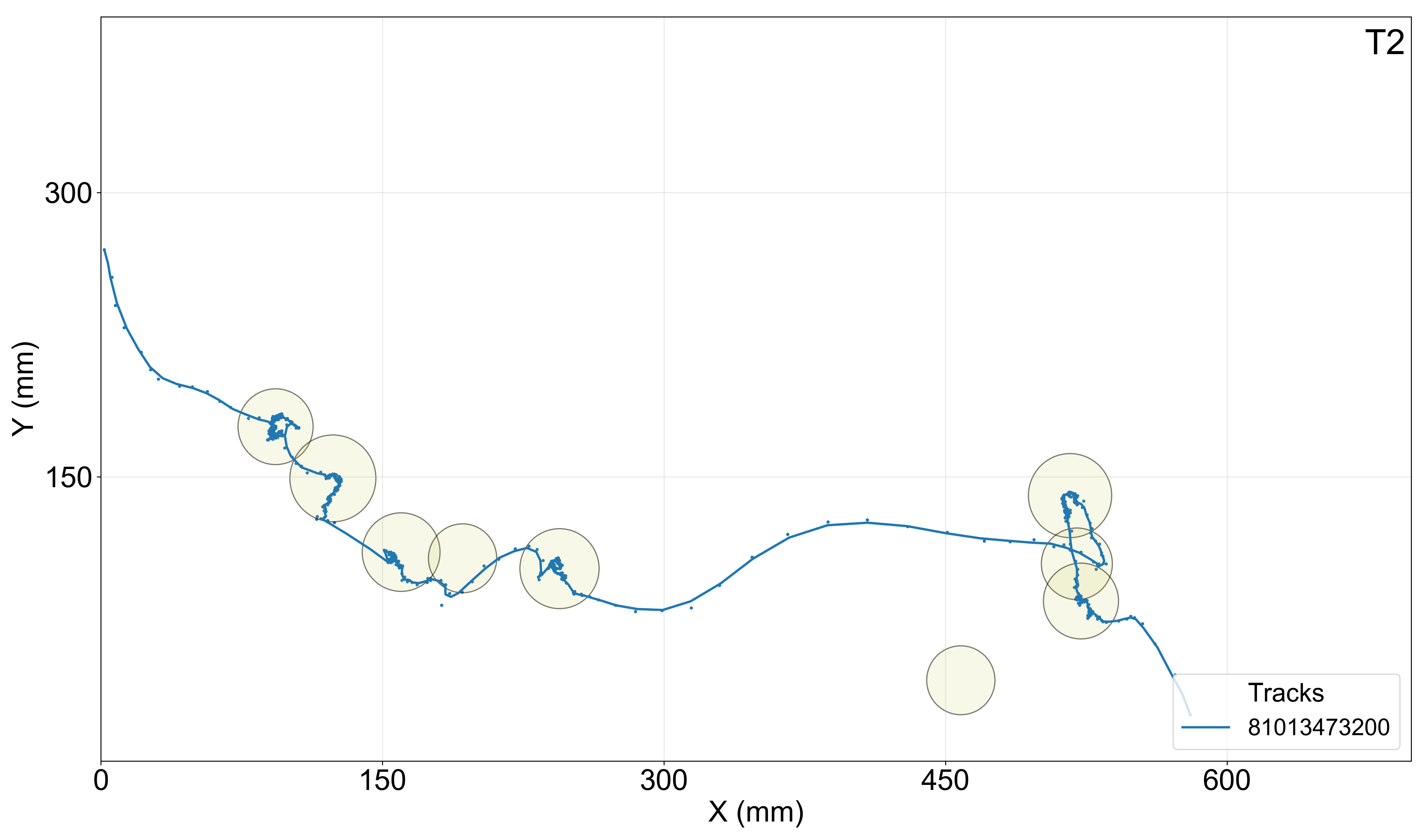}
   \includegraphics[width=0.95\linewidth]{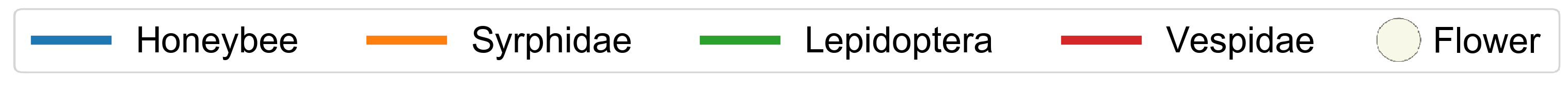}
\end{center}
\end{figure}

\begin{figure}[H]
\begin{center}
   \includegraphics[width=\linewidth]{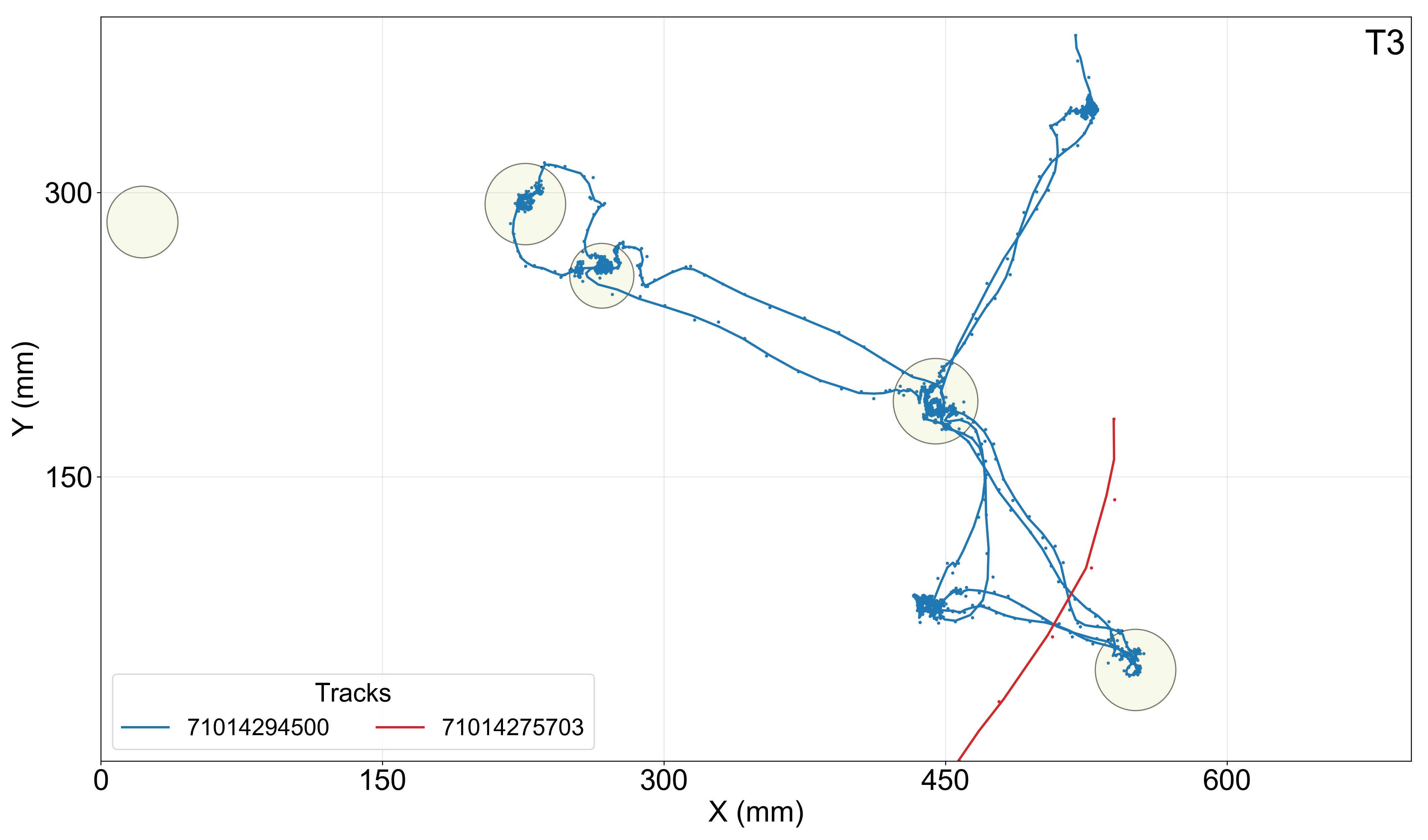}
   \includegraphics[width=\linewidth]{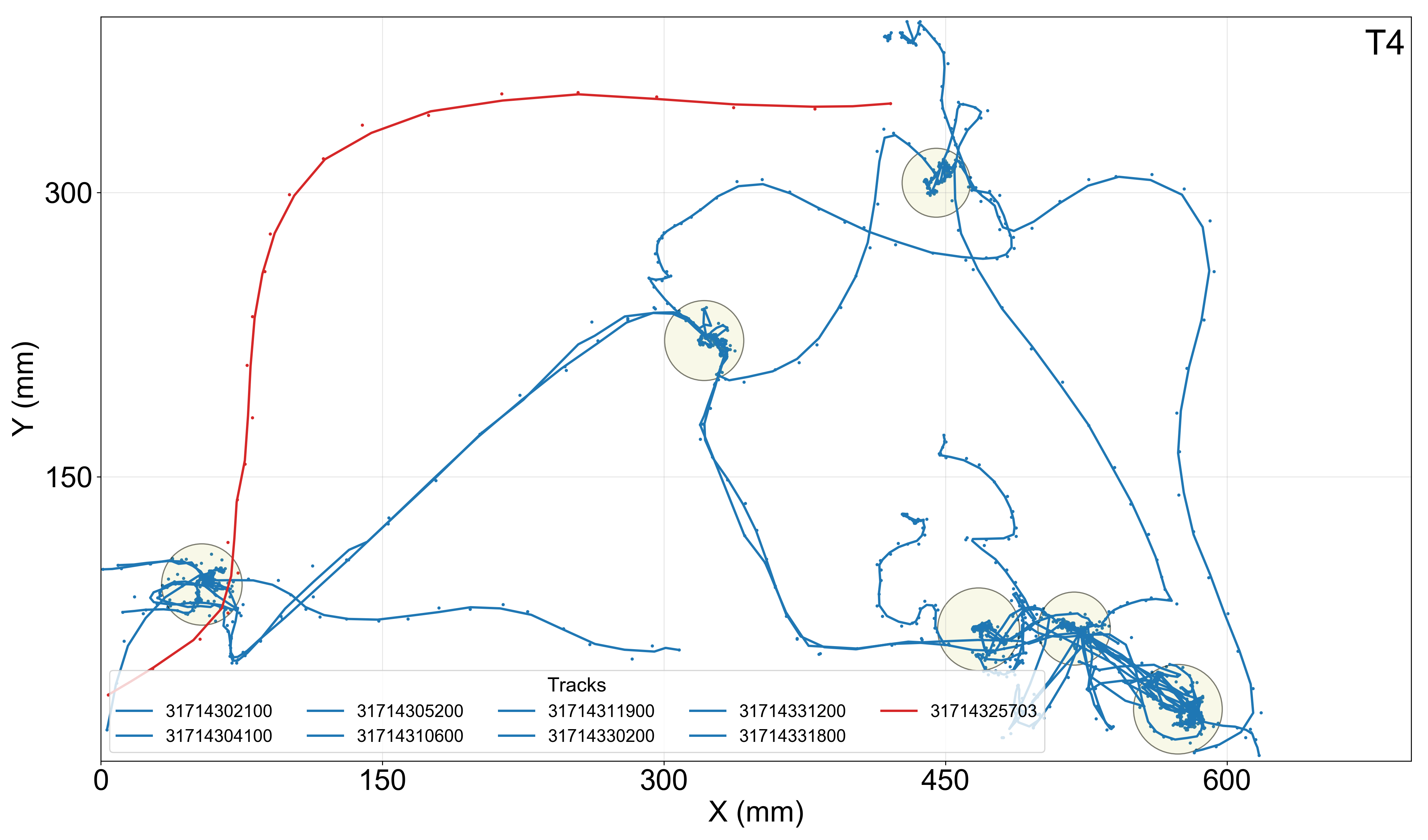}
   \includegraphics[width=0.95\linewidth]{test_tracks/legend_1.png}
\end{center}
\end{figure}

\begin{figure}[H]
\begin{center}
   \includegraphics[width=\linewidth]{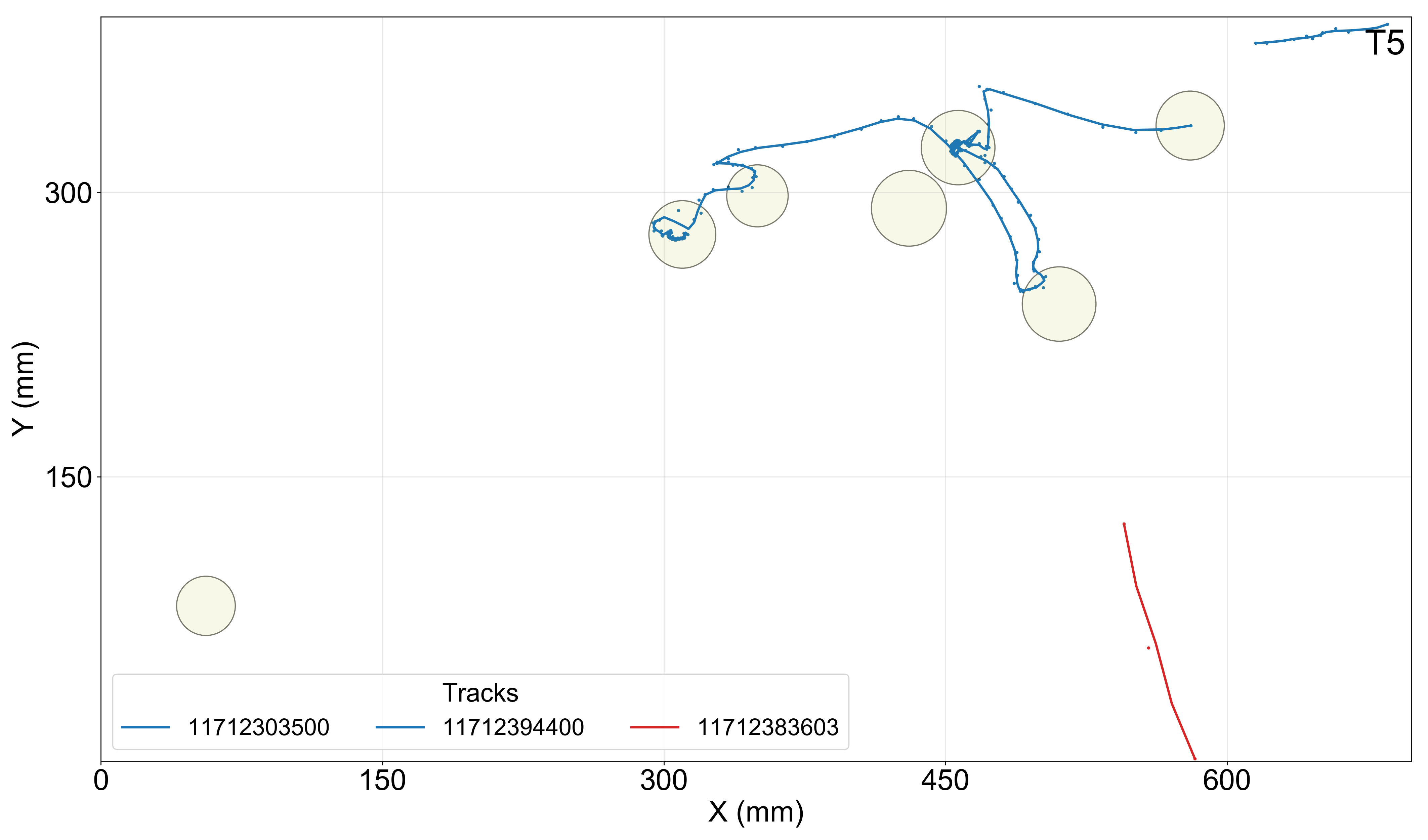}
   \includegraphics[width=\linewidth]{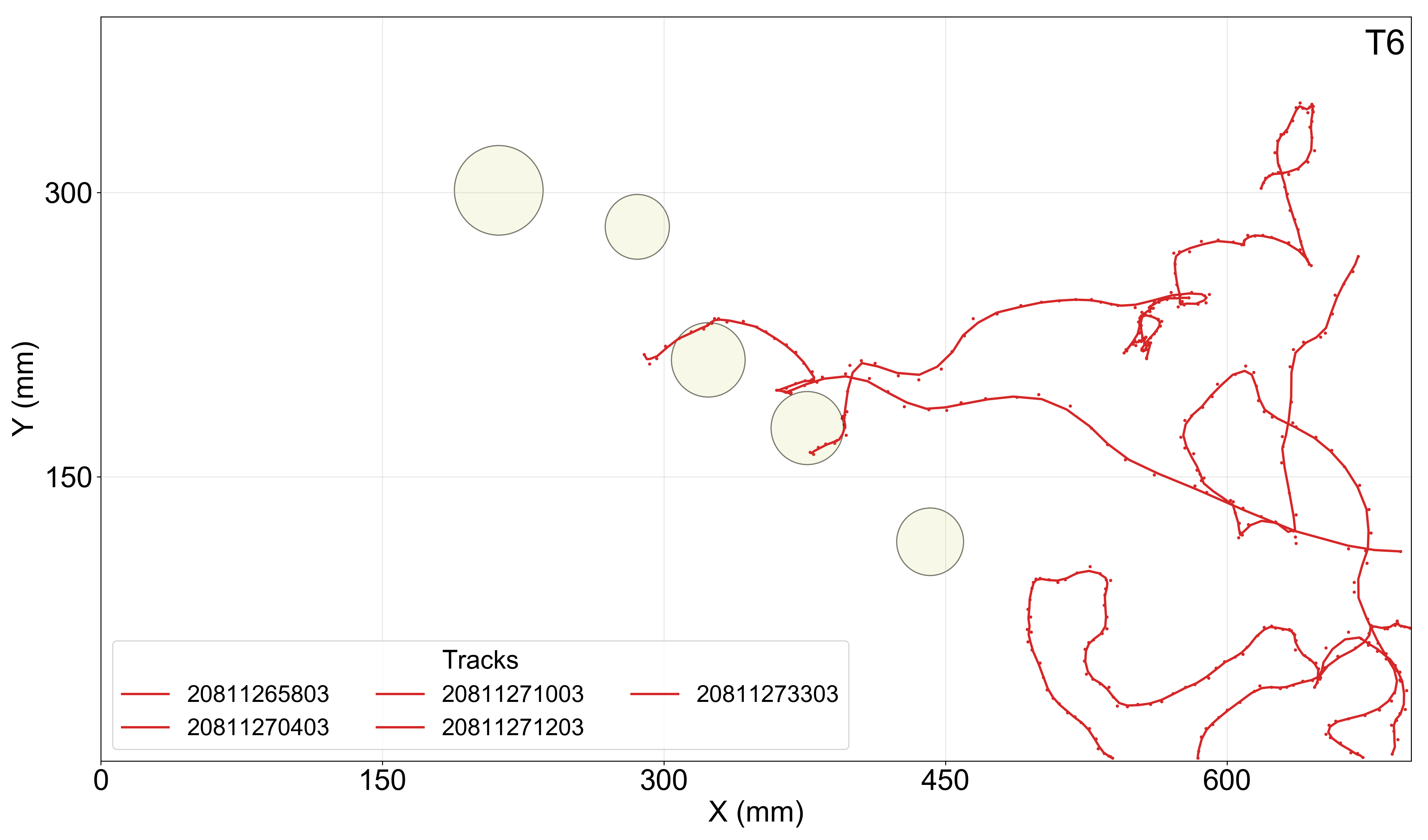}
   \includegraphics[width=0.95\linewidth]{test_tracks/legend_1.png}
\end{center}
\end{figure}

\begin{figure}[H]
\begin{center}
   \includegraphics[width=\linewidth]{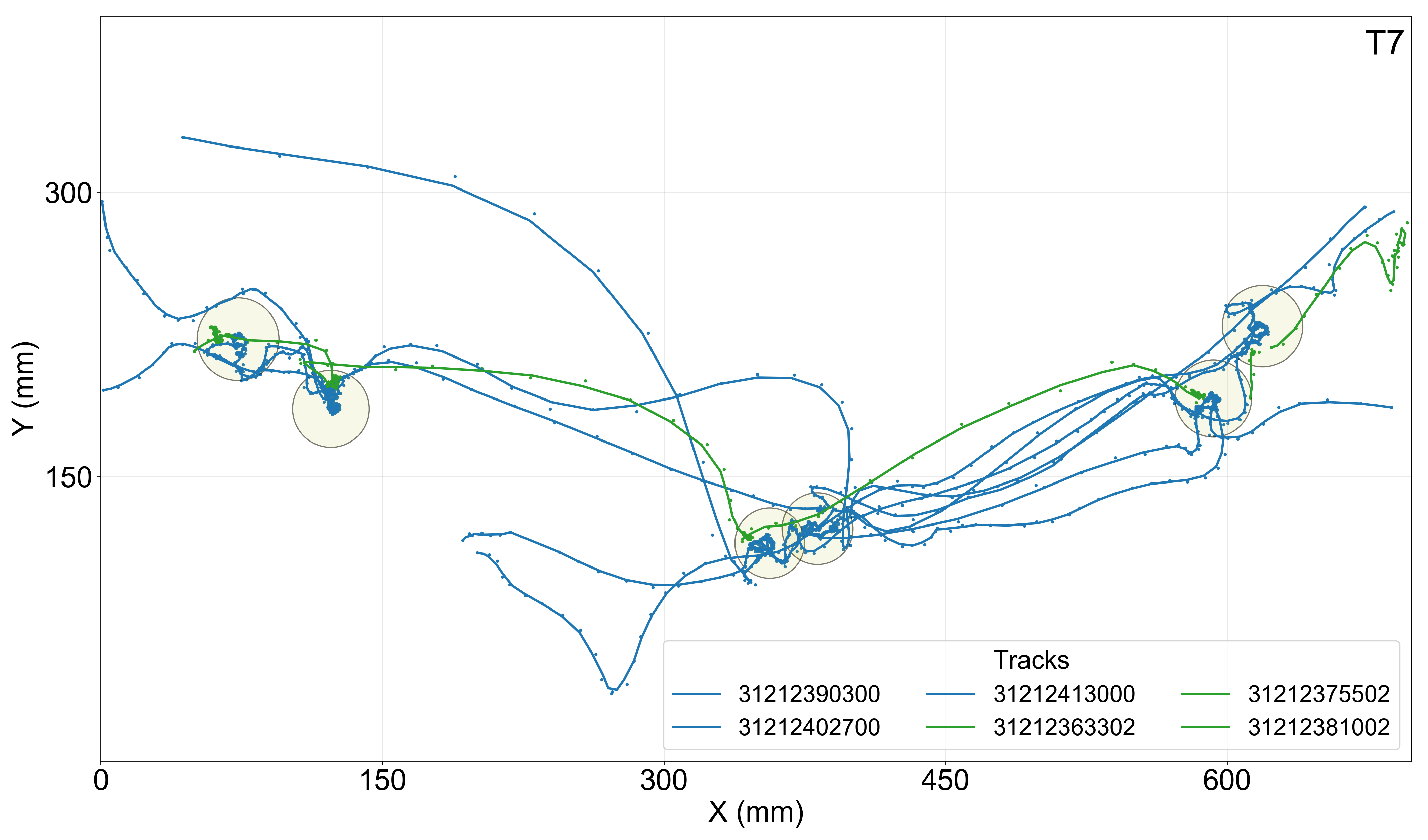}
   \includegraphics[width=\linewidth]{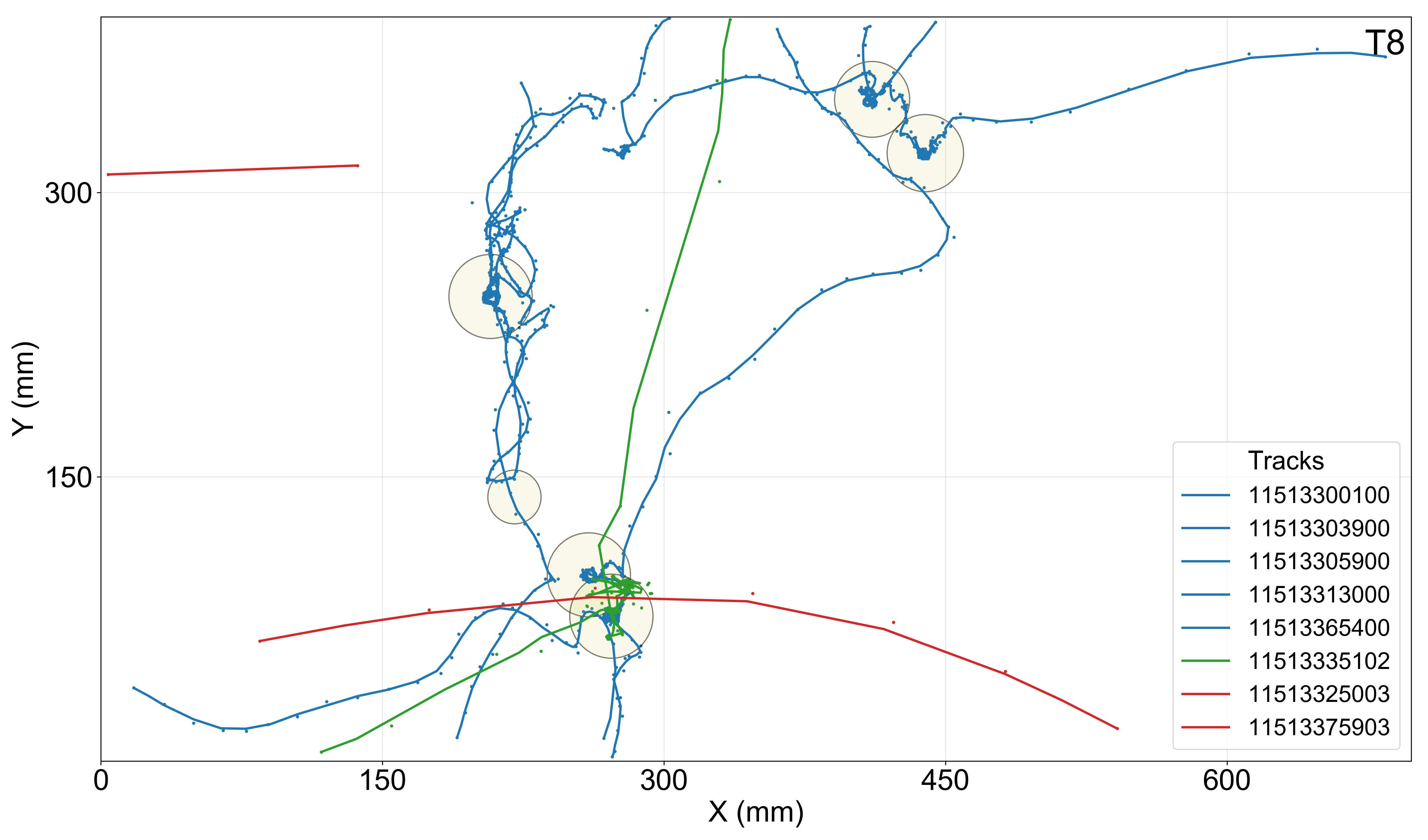}
   \includegraphics[width=0.95\linewidth]{test_tracks/legend_1.png}
\end{center}
\end{figure}

\begin{figure}[H]
\begin{center}
   \includegraphics[width=\linewidth]{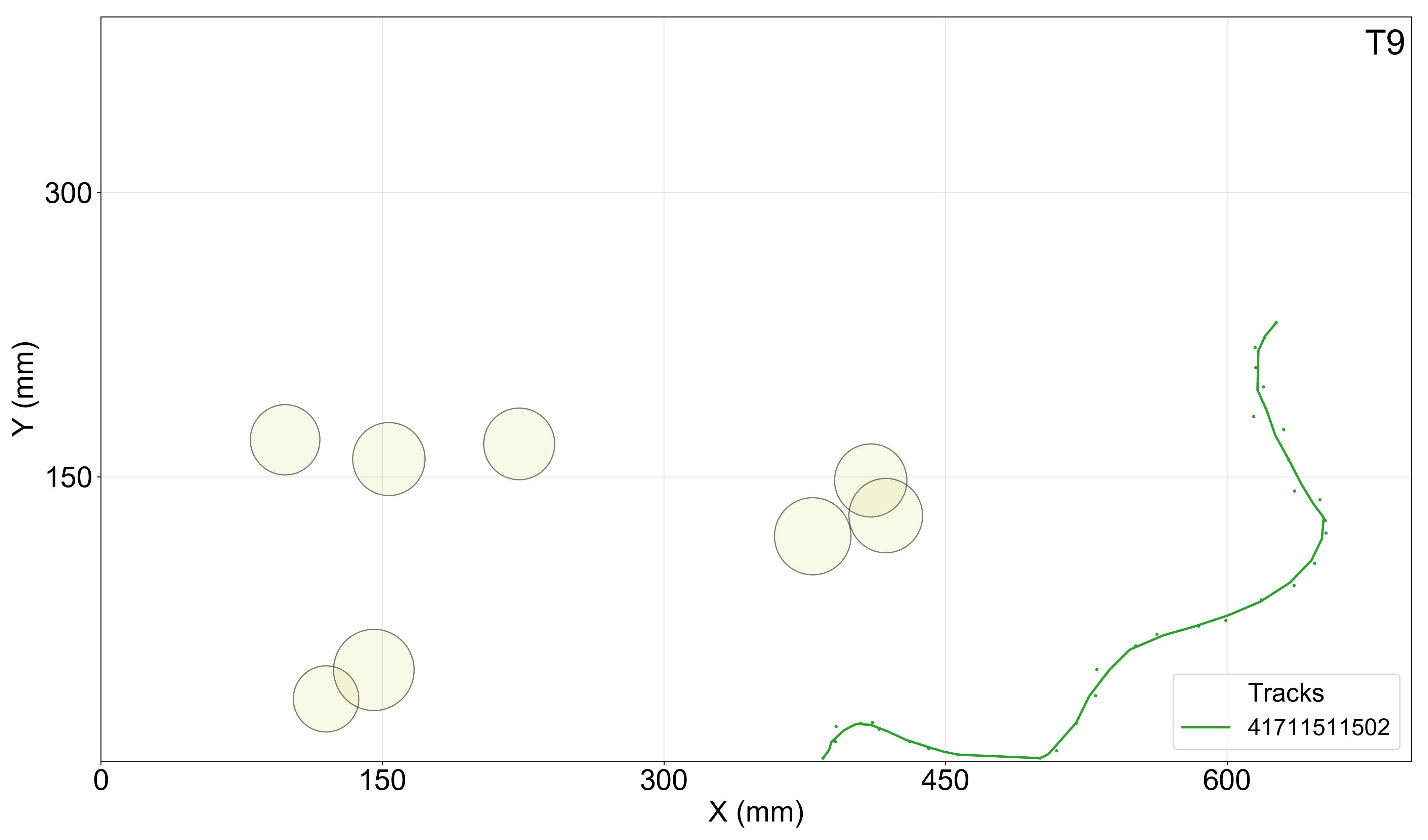}
   \includegraphics[width=\linewidth]{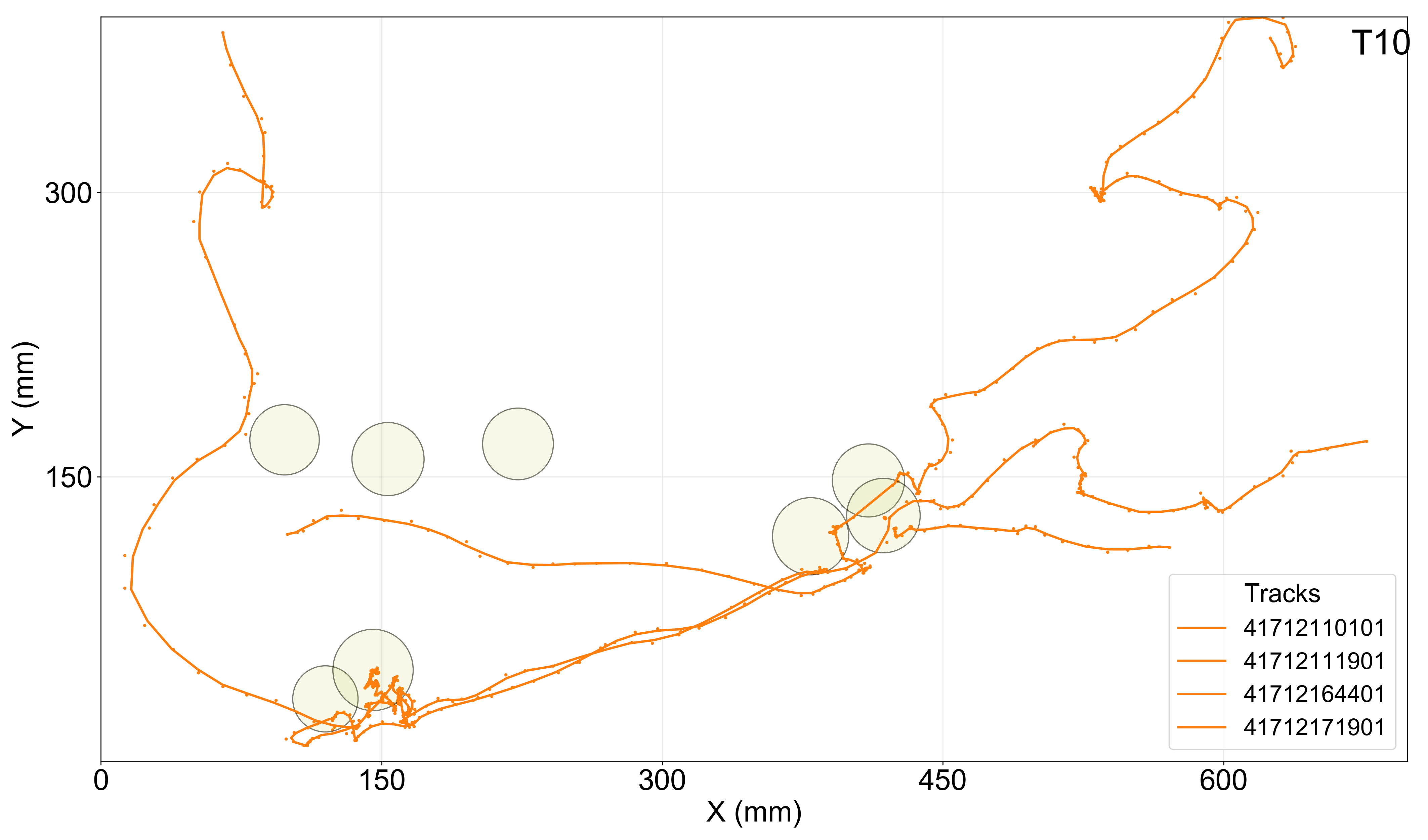}
   \includegraphics[width=0.95\linewidth]{test_tracks/legend_1.png}
\end{center}
\end{figure}

\section*{Supplementary Table: Results of the Flower Detection Evaluation}
\addcontentsline{toc}{section}{Supplementary Table: Flower Detection Evaluation}

\begin{table}[H]
\centering
\caption*{\textbf{Supplementary Table: Results of the Flower Detection Evaluation for the Test Video Dataset.} ``No. of Flowers'' compares the number of manual human observed fully open and fully visible flowers in the video against that detected by the flower tracking component of the algorithm. ``Visible Frames'' shows the number of frames the flower is fully visible. Flower positions were updated at an interval of 100 seconds (3000 frames). ``Confusion Metrics'' compares flower trajectories extracted by our algorithm against manual observations. TP = True Positive, FP = False Positive, and FN = False Negative. A detection was considered a True Positive if the flower bounding circle generated by the algorithm covered the area of the flower. ``Detection Evaluation Metrics'' presents the Precision, Recall and F-Score metrics for each track based on the confusion metrics. Metrics were only calculated for extracted trajectories. }

\resizebox{\linewidth}{!}{
\renewcommand{\arraystretch}{1.2}
\begin{tabular}{|c||c|c||c||c||c|c|c||c|c|c|} 
\hline
\multirow{2}{*}{\begin{tabular}[c]{@{}c@{}}\textbf{Test}\\\textbf{Video}\end{tabular}} & \multicolumn{2}{c||}{\textbf{No. of Flowers}} & \multirow{2}{*}{\begin{tabular}[c]{@{}c@{}}\textbf{Flower}\\\textbf{Code}\end{tabular}} & \multirow{2}{*}{\begin{tabular}[c]{@{}c@{}}\textbf{Visible}\\\textbf{Frames }\end{tabular}} & \multicolumn{3}{c||}{\textbf{Confusion Metrics}} & \multicolumn{3}{c|}{\textbf{Detection Evaluation Metrics}} \\ 
\cline{2-3}\cline{6-11}
 & \textbf{Observed} & \textbf{Recorded} &  &  & \textbf{TP} & \textbf{FP} & \textbf{FN} & \textbf{Precision} & \textbf{Recall} & \textbf{F-Score} \\ 
\hline\hline
\multirow{8}{*}{T1} & \multirow{8}{*}{8} & \multirow{8}{*}{8} & F0 & 17942 & 6 & 0 & 0 & 1.00 & 1.00 & 1.00 \\ 
\cline{4-11}
     &  &  & F1 & 17942 & 6 & 0 & 0 & 1.00 & 1.00 & 1.00 \\ 
\cline{4-11}
 &  &  & F2 & 17942 & 6 & 0 & 0 & 1.00 & 1.00 & 1.00 \\ 
\cline{4-11}
 &  &  & F3 & 17942 & 6 & 0 & 0 & 1.00 & 1.00 & 1.00 \\ 
\cline{4-11}
 &  &  & F4 & 17942 & 6 & 0 & 0 & 1.00 & 1.00 & 1.00 \\ 
\cline{4-11}
 &  &  & F5 & 17942 & 6 & 0 & 0 & 1.00 & 1.00 & 1.00 \\ 
\cline{4-11}
 &  &  & F6 & 17942 & 6 & 0 & 0 & 1.00 & 1.00 & 1.00 \\ 
\cline{4-11}
 &  &  & F7 & 17942 & 6 & 0 & 0 & 1.00 & 1.00 & 1.00 \\ 
\hline\hline
\multirow{9}{*}{T2} & \multirow{9}{*}{9} & \multirow{9}{*}{9} & F0 & 17943 & 6 & 0 & 0 & 1.00 & 1.00 & 1.00 \\ 
\cline{4-11}
 &  &  & F1 & 17943 & 6 & 0 & 0 & 1.00 & 1.00 & 1.00 \\ 
\cline{4-11}
 &  &  & F2 & 17943 & 6 & 0 & 0 & 1.00 & 1.00 & 1.00 \\ 
\cline{4-11}
 &  &  & F3 & 17943 & 6 & 0 & 0 & 1.00 & 1.00 & 1.00 \\ 
\cline{4-11}
 &  &  & F4 & 17943 & 6 & 0 & 0 & 1.00 & 1.00 & 1.00 \\ 
\cline{4-11}
 &  &  & F5 & 17943 & 6 & 0 & 0 & 1.00 & 1.00 & 1.00 \\ 
\cline{4-11}
 &  &  & F6 & 17943 & 6 & 0 & 0 & 1.00 & 1.00 & 1.00 \\ 
\cline{4-11}
 &  &  & F7 & 17943 & 6 & 0 & 0 & 1.00 & 1.00 & 1.00 \\ 
\cline{4-11}
 &  &  & F8 & 17943 & 6 & 0 & 0 & 1.00 & 1.00 & 1.00 \\ 
\hline\hline
\multirow{6}{*}{T3} & \multirow{6}{*}{6} & \multirow{6}{*}{5} & F0 & 17944 & 6 & 0 & 0 & 1.00 & 1.00 & 1.00 \\ 
\cline{4-11}
 &  &  & F1 & 17944 & 6 & 0 & 0 & 1.00 & 1.00 & 1.00 \\ 
\cline{4-11}
 &  &  & F2 & 17944 & 6 & 0 & 0 & 1.00 & 1.00 & 1.00 \\ 
\cline{4-11}
 &  &  & F3 & 17944 & 6 & 0 & 0 & 1.00 & 1.00 & 1.00 \\ 
\cline{4-11}
 &  &  & F4 & 17944 & 6 & 0 & 0 & 1.00 & 1.00 & 1.00 \\ 
\cline{4-11}
 &  &  & N/A & 17944 & \multicolumn{6}{c|}{Trajectory Not Available} \\ 
\hline\hline
\multirow{7}{*}{T4} & \multirow{7}{*}{7} & \multirow{7}{*}{6} & F0 & 17899 & 6 & 0 & 0 & 1.00 & 1.00 & 1.00 \\ 
\cline{4-11}
 &  &  & F1 & 17899 & 6 & 0 & 0 & 1.00 & 1.00 & 1.00 \\ 
\cline{4-11}
 &  &  & F2 & 17899 & 6 & 0 & 0 & 1.00 & 1.00 & 1.00 \\ 
\cline{4-11}
 &  &  & F3 & 17899 & 6 & 0 & 0 & 1.00 & 1.00 & 1.00 \\ 
\cline{4-11}
 &  &  & F4 & 17899 & 6 & 0 & 0 & 1.00 & 1.00 & 1.00 \\ 
\cline{4-11}
 &  &  & F5 & 17899 & 6 & 0 & 0 & 1.00 & 1.00 & 1.00 \\ 
\cline{4-11}
 &  &  & N/A & 17899 & \multicolumn{6}{c|}{Trajectory Not Available} \\ 
\hline
\end{tabular}}
\end{table}

\begin{table}
\centering
\resizebox{\linewidth}{!}{
\renewcommand{\arraystretch}{1.2}
\begin{tabular}{|c||c|c||c||c||c|c|c||c|c|c|} 
\hline
\multirow{2}{*}{\begin{tabular}[c]{@{}c@{}}\textbf{Test}\\\textbf{Video}\end{tabular}} & \multicolumn{2}{c||}{\textbf{No. of Flowers}} & \multirow{2}{*}{\begin{tabular}[c]{@{}c@{}}\textbf{Flower}\\\textbf{Code}\end{tabular}} & \multirow{2}{*}{\begin{tabular}[c]{@{}c@{}}\textbf{Visible}\\\textbf{Frames }\end{tabular}} & \multicolumn{3}{c||}{\textbf{Confusion Metrics}} & \multicolumn{3}{c|}{\textbf{Detection Evaluation Metrics}} \\ 
\cline{2-3}\cline{6-11}
 & \textbf{Observed} & \textbf{Recorded} &  &  & \textbf{TP} & \textbf{FP} & \textbf{FN} & \textbf{Precision} & \textbf{Recall} & \textbf{F-Score} \\ 
\hline\hline
\multirow{7}{*}{T5} & \multirow{7}{*}{7} & \multirow{7}{*}{7} & F0 & 17952 & 6 & 0 & 0 & 1.00 & 1.00 & 1.00 \\ 
\cline{4-11}
 &  &  & F1 & 17952 & 6 & 0 & 0 & 1.00 & 1.00 & 1.00 \\ 
\cline{4-11}
 &  &  & F2 & 17952 & 6 & 0 & 0 & 1.00 & 1.00 & 1.00 \\ 
\cline{4-11}
 &  &  & F3 & 17952 & 6 & 0 & 0 & 1.00 & 1.00 & 1.00 \\ 
\cline{4-11}
 &  &  & F4 & 17952 & 6 & 0 & 0 & 1.00 & 1.00 & 1.00 \\ 
\cline{4-11}
 &  &  & F5 & 17952 & 6 & 0 & 0 & 1.00 & 1.00 & 1.00 \\ 
\cline{4-11}
 &  &  & F6 & 17952 & 6 & 0 & 0 & 1.00 & 1.00 & 1.00 \\ 
\hline\hline
\multirow{5}{*}{T6} & \multirow{5}{*}{5} & \multirow{5}{*}{5} & F0 & 17944 & 6 & 0 & 0 & 1.00 & 1.00 & 1.00 \\ 
\cline{4-11}
 &  &  & F1 & 17944 & 6 & 0 & 0 & 1.00 & 1.00 & 1.00 \\ 
\cline{4-11}
 &  &  & F2 & 17944 & 6 & 0 & 0 & 1.00 & 1.00 & 1.00 \\ 
\cline{4-11}
 &  &  & F3 & 17944 & 6 & 0 & 0 & 1.00 & 1.00 & 1.00 \\ 
\cline{4-11}
 &  &  & F4 & 17944 & 6 & 0 & 0 & 1.00 & 1.00 & 1.00 \\ 
\hline\hline
\multirow{7}{*}{T7} & \multirow{7}{*}{7} & \multirow{7}{*}{6} & F0 & 17940 & 6 & 0 & 0 & 1.00 & 1.00 & 1.00 \\ 
\cline{4-11}
 &  &  & F1 & 17940 & 6 & 0 & 0 & 1.00 & 1.00 & 1.00 \\ 
\cline{4-11}
 &  &  & F2 & 17940 & 6 & 0 & 0 & 1.00 & 1.00 & 1.00 \\ 
\cline{4-11}
 &  &  & F3 & 17940 & 6 & 0 & 0 & 1.00 & 1.00 & 1.00 \\ 
\cline{4-11}
 &  &  & F4 & 17940 & 6 & 0 & 0 & 1.00 & 1.00 & 1.00 \\ 
\cline{4-11}
 &  &  & F5 & 17940 & 6 & 0 & 0 & 1.00 & 1.00 & 1.00 \\ 
\cline{4-11}
 &  &  & N/A & 17940 & \multicolumn{6}{c|}{Trajectory Not Available} \\ 
\hline\hline
\multirow{6}{*}{T8} & \multirow{6}{*}{6} & \multirow{6}{*}{6} & F0 & 17907 & 6 & 0 & 0 & 1.00 & 1.00 & 1.00 \\ 
\cline{4-11}
 &  &  & F1 & 17907 & 6 & 0 & 0 & 1.00 & 1.00 & 1.00 \\ 
\cline{4-11}
 &  &  & F2 & 17907 & 6 & 0 & 0 & 1.00 & 1.00 & 1.00 \\ 
\cline{4-11}
 &  &  & F3 & 17907 & 6 & 0 & 0 & 1.00 & 1.00 & 1.00 \\ 
\cline{4-11}
 &  &  & F4 & 17907 & 6 & 0 & 0 & 1.00 & 1.00 & 1.00 \\ 
\cline{4-11}
 &  &  & F5 & 17907 & 6 & 0 & 0 & 1.00 & 1.00 & 1.00 \\ 
\hline\hline
\multirow{9}{*}{T9} & \multirow{9}{*}{9} & \multirow{9}{*}{8} & F0 & 17944 & 6 & 0 & 0 & 1.00 & 1.00 & 1.00 \\ 
\cline{4-11}
 &  &  & F1 & 17944 & 6 & 0 & 0 & 1.00 & 1.00 & 1.00 \\ 
\cline{4-11}
 &  &  & F2 & 17944 & 6 & 0 & 0 & 1.00 & 1.00 & 1.00 \\ 
\cline{4-11}
 &  &  & F3 & 17944 & 6 & 0 & 0 & 1.00 & 1.00 & 1.00 \\ 
\cline{4-11}
 &  &  & F4 & 17944 & 6 & 0 & 0 & 1.00 & 1.00 & 1.00 \\ 
\cline{4-11}
 &  &  & F5 & 17944 & 6 & 0 & 0 & 1.00 & 1.00 & 1.00 \\ 
\cline{4-11}
 &  &  & F6 & 17944 & 6 & 0 & 0 & 1.00 & 1.00 & 1.00 \\ 
\cline{4-11}
 &  &  & F7 & 17944 & 6 & 0 & 0 & 1.00 & 1.00 & 1.00 \\ 
\cline{4-11}
 &  &  & N/A & 17944 & \multicolumn{6}{c|}{Trajectory Not Available} \\ 
\hline\hline
\multirow{8}{*}{T10} & \multirow{8}{*}{8} & \multirow{8}{*}{8} & F0 & 17891 & 6 & 0 & 0 & 1.00 & 1.00 & 1.00 \\ 
\cline{4-11}
 &  &  & F1 & 17891 & 6 & 0 & 0 & 1.00 & 1.00 & 1.00 \\ 
\cline{4-11}
 &  &  & F2 & 17891 & 6 & 0 & 0 & 1.00 & 1.00 & 1.00 \\ 
\cline{4-11}
 &  &  & F3 & 17891 & 6 & 0 & 0 & 1.00 & 1.00 & 1.00 \\ 
\cline{4-11}
 &  &  & F4 & 17891 & 6 & 0 & 0 & 1.00 & 1.00 & 1.00 \\ 
\cline{4-11}
 &  &  & F5 & 17891 & 6 & 0 & 0 & 1.00 & 1.00 & 1.00 \\ 
\cline{4-11}
 &  &  & F6 & 17891 & 6 & 0 & 0 & 1.00 & 1.00 & 1.00 \\ 
\cline{4-11}
 &  &  & F7 & 17891 & 6 & 0 & 0 & 1.00 & 1.00 & 1.00 \\
\hline
\end{tabular}}
\end{table}

\clearpage

\section*{Supplementary Table: Test Dataset Processing Time}
\addcontentsline{toc}{section}{Supplementary Table: Test Dataset Processing Time}

\begin{table}[H]
\centering
\caption*{\textbf{Supplementary Table: Test Dataset Processing Time.} Processing Times (seconds) and speeds (frames per second - fps) for the test dataset with and without the low-resolution processing mode (adopted from Polytrack \citep{Ratnayake_2021_CVPR}). Processing time denotes the time taken by the algorithm to process a video and the processing speed shows the average number of frames processed in a second.}
\renewcommand{\arraystretch}{1.2}
\begin{tabular}{|c|c|c|c|c|} 
\hline
\multirow{2}{*}{\begin{tabular}[c]{@{}c@{}}\textbf{Test }\\\textbf{Video}\end{tabular}} & \multicolumn{2}{c|}{\textbf{Processing Time (sec)}} & \multicolumn{2}{c|}{\textbf{Processing Speed (fps)}} \\ 
\cline{2-5}
 & \begin{tabular}[c]{@{}c@{}}\textbf{With }\\\textbf{ Low-Res Mode}\end{tabular} & \begin{tabular}[c]{@{}c@{}}\textbf{Without }\\\textbf{ Low-Res Mode}\end{tabular} & \begin{tabular}[c]{@{}c@{}}\textbf{With }\\\textbf{ Low-Res Mode}\end{tabular} & \begin{tabular}[c]{@{}c@{}}\textbf{Without }\\\textbf{ Low-Res Mode}\end{tabular} \\ 
\hline
T1 & \textbf{939.9} & 2414.3 & \textbf{19.2} & 7.5 \\
T2 & \textbf{993.5} & 2631.9 & \textbf{18.1} & 6.8 \\
T3 & \textbf{1440.0} & 2937.4 & \textbf{12.5} & 6.1 \\
T4 & \textbf{921.8} & 2314.2 & \textbf{19.5} & 7.8 \\
T5 & \textbf{737.6} & 2316.7 & \textbf{24.4} & 7.8 \\
T6 & \textbf{587.2} & 1959.0 & \textbf{30.7} & 9.2 \\
T7 & \textbf{1167.7} & 2332.5 & \textbf{15.4} & 7.7 \\
T8 & \textbf{853.9} & 2153.4 & \textbf{21.0} & 8.3 \\
T9 & \textbf{500.3} & 2042.8 & \textbf{36.0} & 8.8 \\
T10 & \textbf{656.0} & 2258.9 & \textbf{27.4} & 7.9 \\ 
\hline
Overall & \textbf{8797.9} & 23361.1 & \textbf{20.4} & 7.7 \\
\hline
\end{tabular}
\end{table}

\clearpage

\bibliography{sn-bibliography}